\documentclass{article} 
\usepackage{iclr2026_conference,times}

\usepackage{amsmath,amsfonts,bm}

\def\eqref#1{equation~\ref{#1}}

\def\1{\bm{1}}

\DeclareMathAlphabet{\mathsfit}{\encodingdefault}{\sfdefault}{m}{sl}
\SetMathAlphabet{\mathsfit}{bold}{\encodingdefault}{\sfdefault}{bx}{n}

\usepackage[utf8]{inputenc} 
\usepackage[T1]{fontenc}    
\usepackage{hyperref}       
\usepackage{url}            
\usepackage{booktabs}       
\usepackage{amsfonts}       
\usepackage{nicefrac}       
\usepackage{microtype}      
\usepackage[table]{xcolor}  
\usepackage{comment}  
\usepackage{subcaption}  
\usepackage{graphicx}  
\usepackage{amsmath}  
\usepackage{makecell}  
\usepackage{multirow}  
\usepackage{tcolorbox}  
\usepackage{algorithm}      
\usepackage{algpseudocode}  
\usepackage{colortbl}  
\usepackage{placeins}
\usepackage{array}
\usepackage{siunitx}

\title{Taming the Fragility of KV Cache Eviction \\ in LLM Inference}

\author{
	Yuan Feng\textsuperscript{1,3,$\dagger$}, 
	Haoyu Guo\textsuperscript{2,3,$\dagger$}, 
	Junlin Lv\textsuperscript{1,3}, 
	S. Kevin Zhou\textsuperscript{2,3},
	Xike Xie\textsuperscript{2,3,$*$},  \\
	\\
	\textsuperscript{1}School of Computer Science, University of Science and Technology of China \\
	\textsuperscript{2}School of Biomedical Engineering, USTC \\
	\textsuperscript{3}Data Darkness Lab, MIRACLE Center, Suzhou Institute for Advanced Research
}

\iclrfinalcopy 
\begin{document}
\pagestyle{plain} 
	
\def\thefootnote{}\footnotetext{
	\textsuperscript{$\dagger$}Equal Contribution:  \texttt{ \{yfung,haoyuguo\}@mail.ustc.edu.cn}}
\def\thefootnote{}\footnotetext{
 \textsuperscript{*}Corresponding Author:  \texttt{ xkxie@ustc.edu.cn}}
\def\thefootnote{\arabic{footnote}}

\maketitle

\begin{abstract}\label{sec:abstract}
	Large language models have revolutionized natural language processing, yet their deployment remains hampered by the substantial memory and runtime overhead of the transformer’s Key-Value cache. To mitigate this, recent methods employ a scoring-aggregation framework to evict unimportant cache entries, based on the "stability assumption"—that a fixed subset of entries remains consistently important during generation. However, prior work has largely focused on refining importance indicators for scoring, while defaulting to mean aggregation due to a faithful trust in the stability assumption. 
	In this work, we argue that this underlying assumption is inherently fragile, making mean aggregation highly vulnerable in extreme cases.
	To counter this, we propose a simple yet elegant defensive aggregation strategy: a two-step, linear-time approach that controls worst-case risk, thereby defending against extreme cases with negligible computational overhead.
	Embodying this strategy, we propose a novel cache eviction method, DefensiveKV and its extension, Layer-DefensiveKV, which incorporates layer-wise budget allocation. Across seven task domains (18 datasets), our methods reduce generation quality loss by 2.3× and 4.3× respectively, versus the strongest baseline under a 20\% cache size. These results set new performance benchmarks and pioneer a promising direction for optimizing cache eviction against underlying fragility through worst-case risk management. Our code is available at \url{https://github.com/FFY0/DefensiveKV}.	
\end{abstract}

\section{Introduction}\label{sec:introduction}
Transformer-based Large Language Models (LLMs) have enabled a wide range of applications~\citep{yi2024survey, gu2023llm}. Due to their autoregressive nature, LLMs maintain a Key-Value (KV) cache to store intermediate representations of previously tokens, which supports efficient computation of future  generation. However, as the input sequence length increases, the KV cache grows linearly, leading to substantial  overhead. For example, a 70B-parameter model with a batch size of 8 and a sequence length of 128k may require up to 330GB of memory just for caching. This poses significant challenges regarding storage expenses and I/O bottlenecks for LLM deployment~\citep{kvpress}.

Early solutions like StreamingLLM~\citep{streamingllm} reduce cache size by keeping only recent cache entries, but this sacrifices long-range context.
More recent solutions~\citep{h2o,liu2024scissorhands, SnapKV,feng2025identifycriticalkvcache} on selective cache eviction operate under a key underlying assumption that {\it a fixed subset of cache entries remain consistently important and contributes to future generation}.
Thus, by retaining the selected subset, the full KV cache can be approximated with a much smaller memory footprint. Building on this, existing methods typically follow a two-step {scoring-aggregation} framework: In the scoring step, different historical token queries are used to observe multiple importance scores for each past KV cache entry.
In the aggregation step, these multiple observed scores for  each cache entry  are aggregated—typically by averaging—to estimate its expected significance and guide the eviction strategy.

Following this two-step framework, previous research has primarily focused on improving the scoring step by exploring various importance indicators.
Early studies often relied solely on naive attention weights~\citep{h2o,liu2024scissorhands,oren2024transformers}. SnapKV~\citep{SnapKV} improved by introducing a pooling mechanism, while more recent work like CriticalKV~\citep{feng2025identifycriticalkvcache} employed the norm of projected value states to offer a more principled measure of importance.
However, the second step—aggregating these importance scores—remains largely underexplored. Most existing methods default to a simple averaging strategy.
While this may seem reasonable, averaging is only effective if the underlying assumption holds that importance are stable-when it does, averaging helps reduce observation noise and capture the consistent significance of cache entries.

\begin{figure}[t]
	\vspace{-0.4cm}
	\centering
	\includegraphics[width=0.95\textwidth]{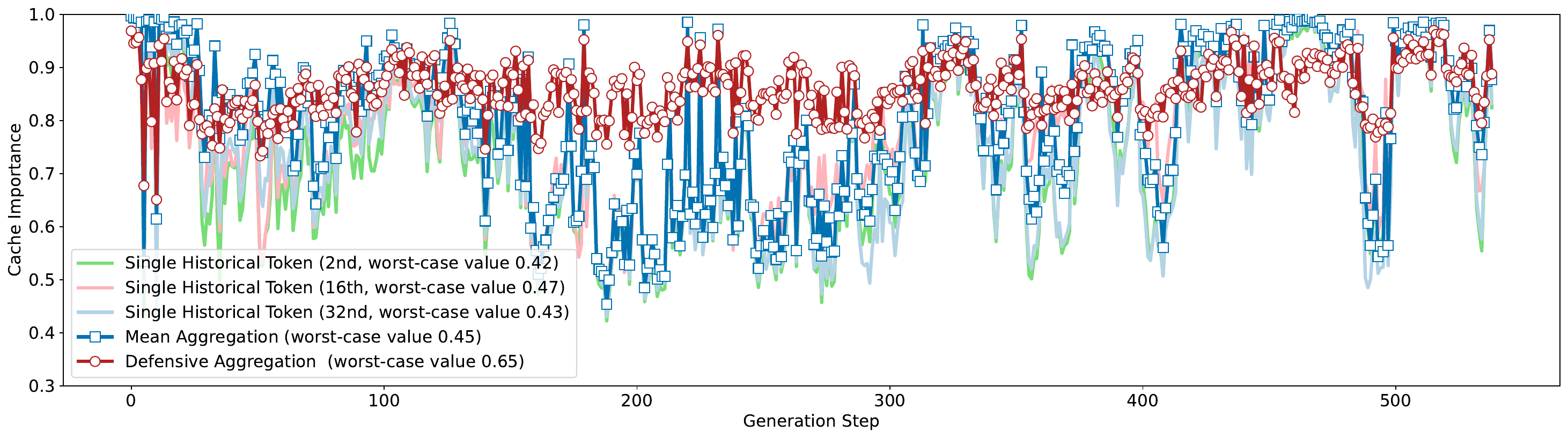}
	\vspace{-0.5cm}
	\caption{Defensive aggregation demonstrates robustness against  fragile stability assumption  (Llama-3.1-8B, 50\% cache size, layer 14, summary task). Appendix~\ref{apdx:more_main} provides additional visualizations.} 
	\label{fig:main}
\end{figure}

\begin{figure}[t]
	\centering
	\vspace{-0.3cm}
	\begin{subfigure}{0.32\textwidth}
		\centering
		\includegraphics[width=\linewidth]{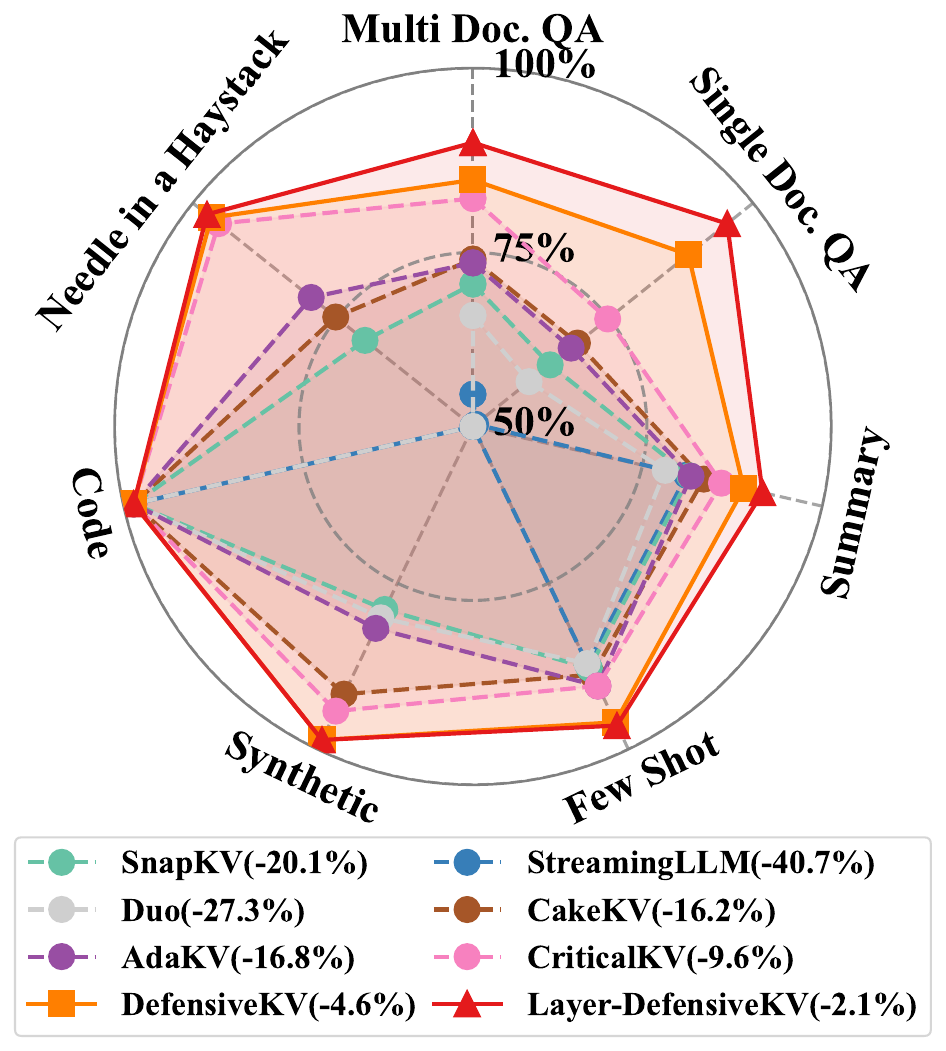}
		\vspace{-0.6cm}
		\caption{Llama-3.1-8B, 20\% Cache}
	\end{subfigure}
	\begin{subfigure}{0.32\textwidth}
		\centering
		\includegraphics[width=\linewidth]{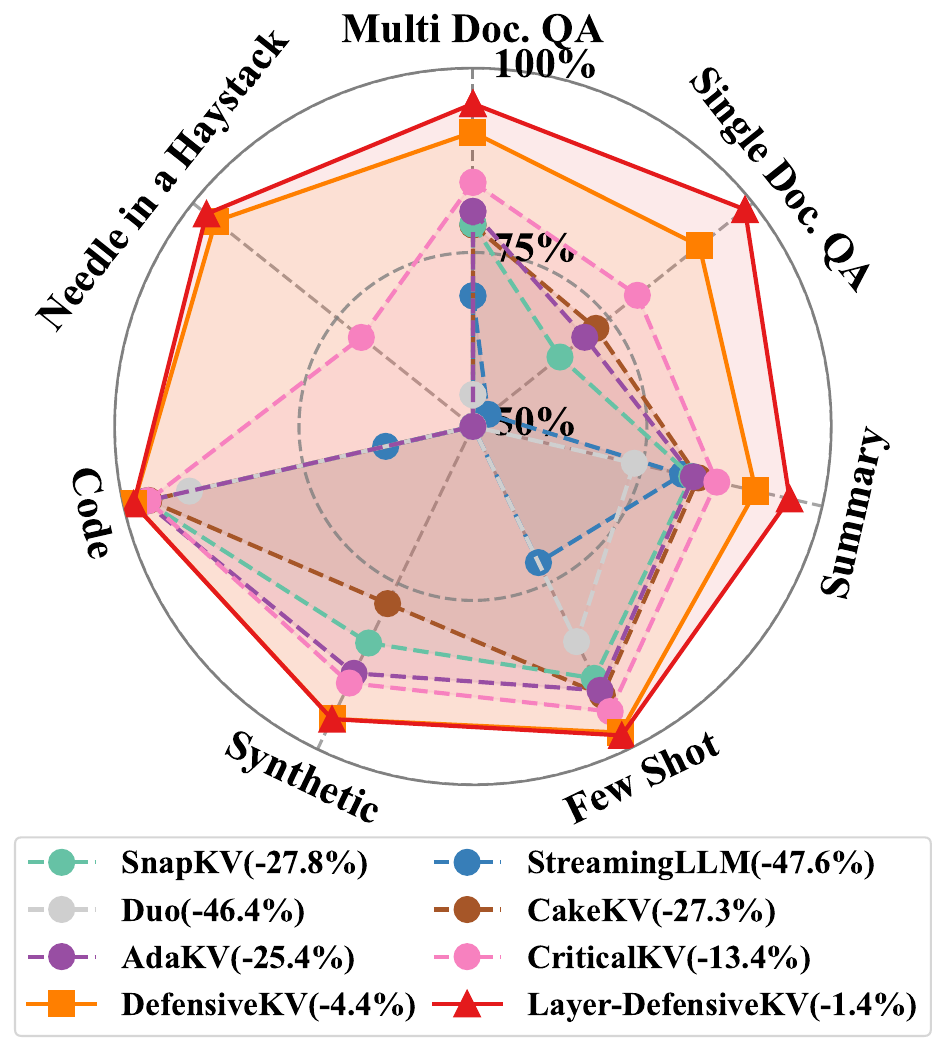}
		\vspace{-0.6cm}
		\caption{Mistral-7B, 20\% Cache}
	\end{subfigure}
	\begin{subfigure}{0.32\textwidth}
		\centering
		\includegraphics[width=\linewidth]{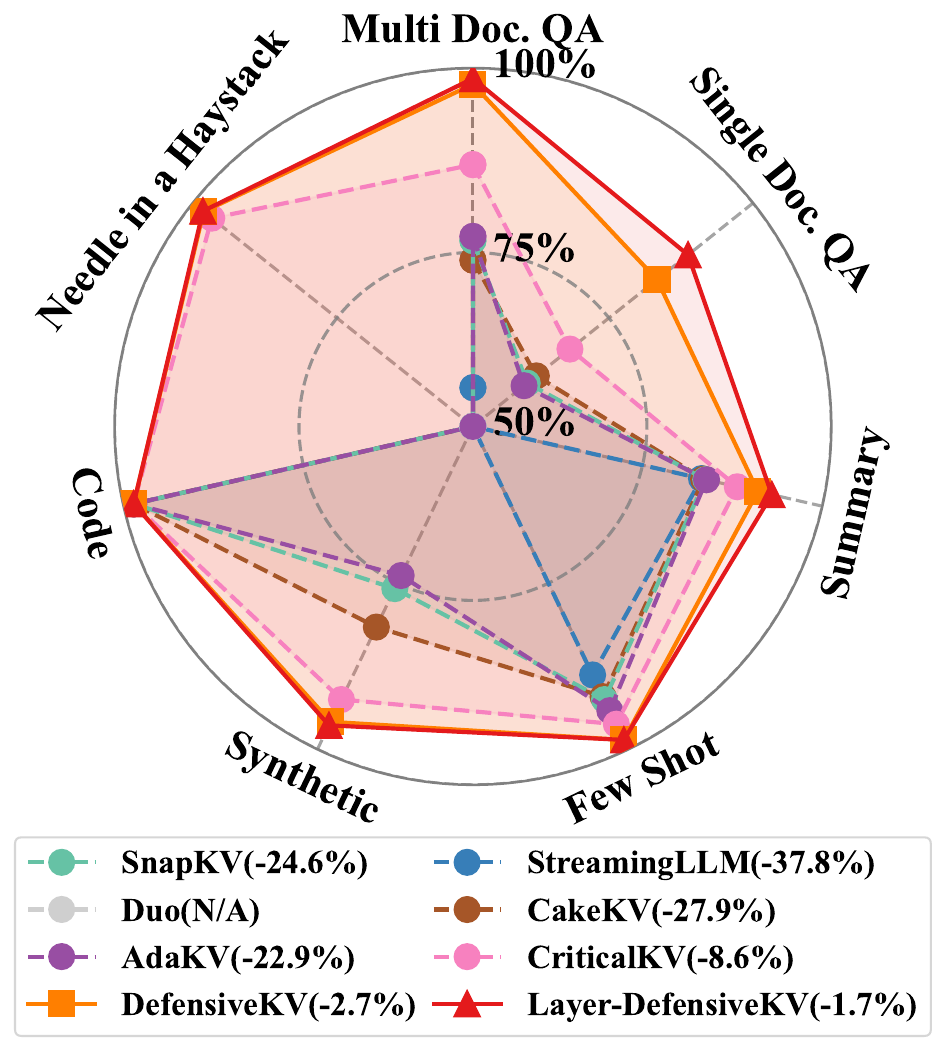}
		\vspace{-0.6cm}
		\caption{Qwen-32B, 20\% Cache}
	\end{subfigure}
	\vspace{-0.2cm}
	\caption{ DefensiveKV and Layer-DefensiveKV achieve significantly lower losses of generation quality compared to all baselines across various domains and models. }
	\vspace{-0.4cm}
	\label{fig:main_performance}
\end{figure}

\begin{tcolorbox}[colback=gray!10, colframe=gray!50, boxrule=0.1pt, sharp corners=south,left=2pt,
	right=2pt,
	top=0pt,
	bottom=0pt,
	boxsep=1pt ]
	\emph{There raises a critical question: if the stability assumption proves unreliable,
is averaging still the best aggregation strategy, or might better alternatives exist?}
\end{tcolorbox}

In this work, we show that even when the assumption generally holds, it remains inherently fragile, as importance scores can shift abruptly during generation. As demonstrated in Figure \ref{fig:main},
performing cache compression based on the observed importance score of a single historical token often yields promising results, with most steps retaining over 0.8 correlation with full-cache importance. However, in certain intervals (e.g., steps 150–230), the stability assumption breaks down—consequently, results based on single historical token fail, leading to sharp drops, with some outliers falling as low as 0.5. 
In these cases, the current standard practice of mean aggregation, simply averaging these single token predictions, inevitably results in similar outlier performance.

\begin{tcolorbox}[colback=gray!10, colframe=gray!50, boxrule=0.1pt, sharp corners=south,left=2pt,
	right=2pt,
	top=0pt,
	bottom=0pt,
	boxsep=1pt ]
	\emph{
	This reflects a classic pitfall, a flaw directly analogous to a foundational lesson from finance: strategies that optimize only for the average case (expected returns) are fundamentally flawed because they ignore the risk of rare but extreme negative cases (worst-case risks).
	}
\end{tcolorbox}

Inspired by this insight, we abandon average-case optimization in favor of a worst-case risk management framework for KV cache eviction, which we term defensive aggregation. Our strategy is actualized through an elegant two-step process: worst-case estimation and adaptive prior-risk correction. Remarkably, this approach requires only two linear-time operations, matching the computational efficiency of standard mean aggregation. As shown in Figure \ref{fig:main}, Defensive Aggregation demonstrates clear superiority, boosting the worst-case retained importance to 0.65—a substantial improvement over both mean aggregation (0.45) and single-token baselines (0.42, 0.47, 0.43).

Building on the defensive aggregation strategy, we introduce DefensiveKV, a general cache eviction method, which we further develop into Layer-DefensiveKV by leveraging a popular layer-wise budget allocation strategy. Figure~\ref{fig:main_performance} summarizes that these two methods significantly outperform prior approaches across seven task domains, evaluated on 18 datasets from the LongBench and Needle-in-a-Haystack benchmarks. With a 20\% cache budget, DefensiveKV and Layer-DefensiveKV incur generation quality losses of only 4.8\% and 2.6\%, respectively, representing 2.3× and 4.3× reductions versus the best baseline, CriticalKV (11.1\%).
\section{Related Works}\label{sec:retated}

KV cache eviction is crucial for efficient long-sequence inference in LLMs. Early methods like StreamingLLM~\citep{streamingllm} retained only recent cache entries, often losing valuable information. Subsequent H2O~\citep{h2o} and Scissorhands~\citep{liu2024scissorhands} introduced importance-based eviction, assuming "cache importance stability"—that a small set of entries remains consistently important. These methods typically observe importance multi times with several historical tokens and aggregate these, often by averaging, to decide on eviction~\cite{ren2024efficacyevictionpolicykeyvalue,oren2024transformers}. While research has advanced importance observation— SnapKV~\citep{SnapKV} with pooling, and CriticalKV~\citep{feng2025identifycriticalkvcache} with projected value norms—the foundational stability assumption has rarely been rigorously examined. This paper revisits and reveals the fragility of this assumption, further showing prevalent mean aggregation's vulnerability. Consequently, we are the first to underscore the necessity of risk-control defensive aggregation strategies to against fragile assumption. This pioneers a new research direction, entirely orthogonal to prior work focused on optimizing importance indicators.
For demonstration, we build our DefensiveKV method upon CriticalKV, the current SOTA importance indicator.

Additionally, our contributions is also orthogonal to various KV cache budget allocation strategies, including intra-layer (e.g., AdaKV~\citep{ada}), inter-layer (e.g., PyramidKV~\citep{pyramidkv}, LightTransfer~\citep{zhang2025lighttransfer}, CAKE~\citep{qin2025cake}), and also offline training-based allocation (e.g., HeadKV~\citep{fu2024not}, DuoAttention~\citep{xiao2025duoattention}). 
These strategies focus on optimizing budget allocation for cache eviction methods, and are thus inherently orthogonal to our investigation.  Direct comparison is not essential for validating our contributions. However, to demonstrate our principles' adaptability, we introduce Layer-DefensiveKV, a variant using layer-wise budget allocation for enhancement. Broader related methods like quantization, channel pruning, and sparse attention are discussed in Appendix~\ref{apdx:related}. {Furthermore, we provide a case study in Appendix~\ref{apdx:quan} on integrating our DefensiveKV with quantization, showing minimal loss even at 10\% cache footprint.}

\section{Methods}\label{sec:methods}
\subsection{Preliminary}

LLM generation consists of two stages, {\it prefilling} and {\it decoding}. During prefilling, the KV states for all input tokens are computed and cached as: $K = H W_K, \: V = H W_V$, where \( H \in \mathbb{R}^{n \times d} \) denotes the hidden states for  \( n \) tokens, and \( W_K, W_V \in \mathbb{R}^{d \times d_h} \) are learned  matrices.
In decoding, the LLM takes the most recent token, computes its query vector \( q_{j} = H_{j=-1,:} W_Q \), and retrieves information from the cached KV entries using attention to produce the output $o_j$ and predict the next token:
\vspace{-0.15cm}
\[
\small
o_{j} = A_{j} V W_O \: \text{where} \: A_{j} = \mathrm{softmax}\left( q_{j} K^\top/ \sqrt{d_h} \right)
\]
To reduce the memory overhead of maintaining the full KV cache, cache eviction methods have been developed. These methods largely operate under a stability assumption: {\it a fixed subset of KV cache entries, denoted as} $(\hat{K}, \hat{V})$, {\it retains consistent importance throughout generation}.
Based on this assumption, the objective of cache eviction is to identify this crucial subset ($\hat{K}, \hat{V}$) using historical queries (i.e., tokens from earlier generation process),
and use it to replace the full KV cache \((K, V)\) in subsequent steps.
This process typically follows a two-step {\it scoring-aggregation} framework, where the importance of each KV entry is first estimated (or scored) and then aggregated:

\begin{enumerate}
	\item \textbf{Scoring.} Given \( m \) historical tokens, represented as queries \( Q = [q_1, \ldots, q_m] \), each of the \( n \) KV cache entries \( K, V \in \mathbb{R}^{n\times d_h} \) is scored. This results in an importance matrix \( I \in \mathbb{R}^{m \times n} \), where each element \( I_{j,i} \) measures the relevance of the \( i \)-th KV cache entry \((k_i, v_i)\) for the \( j \)-th historical query \( q_j \). In practice, the attention weight \(A_{j,i}\) serves as a direct measure of importance \( I_{j,i} \), given the attention mechanism's inherent weighted-sum formulation.
	
	\item \textbf{Aggregation.} Subsequent aggregation step converts the observed importance matrix $I \in \mathbb{R}^{m \times n}$ into a vector $S \in \mathbb{R}^n$, where each  $S_i$ represents the importance of the $i$-th KV cache entry. Existing works  adopt mean aggregation, $S_i = \frac{1}{m}\sum_{j=1}^{m} I_{j,i}$, to highlight entries with consistently high importance, in line with the stability assumption.
	    
\end{enumerate}

While subsequent studies have refined the scoring step—SnapKV employs a pooling mechanism, and CriticalKV utilizes the norm of projected value states $v_i W_O$ for more principled scoring—the aggregation step has received little attention. This is largely because mean aggregation appears to align closely with underlying importance stability assumption. However, in this work, we show the fragility of that assumption and reveal the vulnerability of mean aggregation under this premise. This underscores the necessity of revisiting and improving the aggregation step.

\begin{figure}[t]
	\centering
	\vspace{-0.3cm}
	\begin{minipage}[c]{0.84\textwidth}
		\centering
			\centering
			\begin{subfigure}[b]{0.85\textwidth}
				\includegraphics[width=\textwidth]{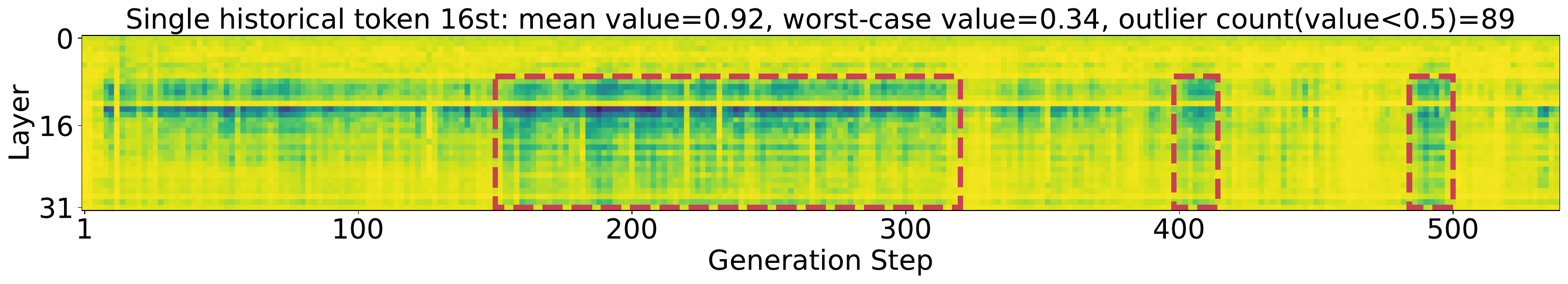}
				\vspace{-0.7cm}
				\caption{\small 16st historical token: Vulnerable; importance drops to 0.34 (worst case).}
				\label{fig:agg_assumption_single}
			\end{subfigure}
			\begin{subfigure}[b]{0.85\textwidth}
			\centering
				\includegraphics[width=\textwidth]{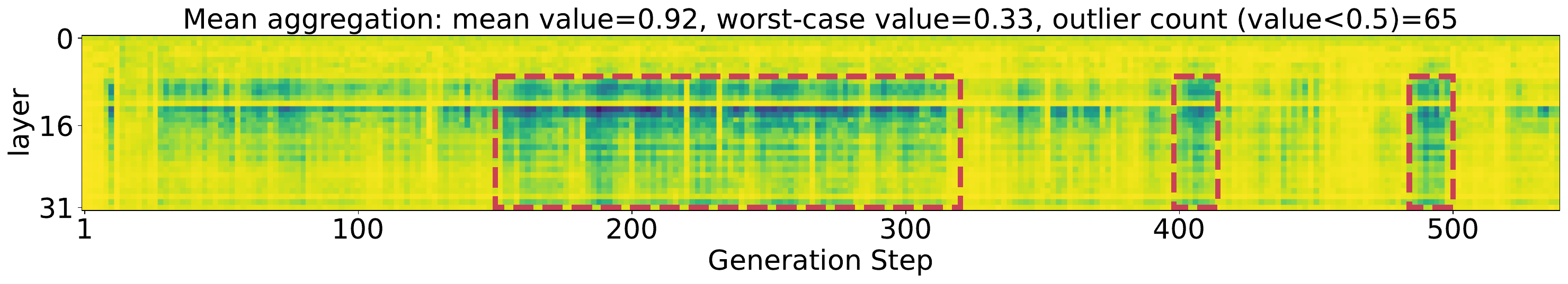}
				\vspace{-0.7cm}
				\caption{\small Mean aggregation: Vulnerable; importance drops to 0.33 (worst case).}
				\label{fig:mean_agg_assumption}
			\end{subfigure}
			\begin{subfigure}[b]{0.85\textwidth}
			\centering
				\includegraphics[width=\textwidth]{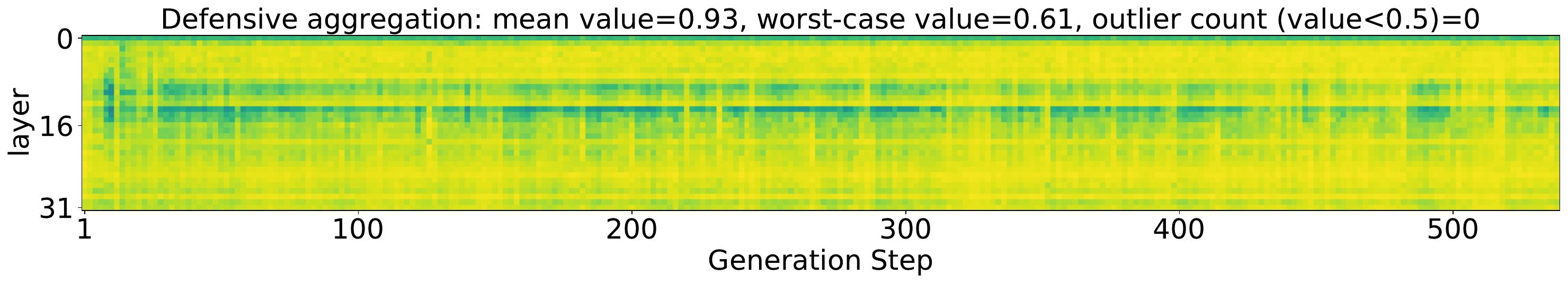}
				\vspace{-0.7cm}
				\caption{\small Defensive aggregation: Robust; maintains 0.61 importance (worst case).}
				\label{fig:defensive_agg_assumption}
			\end{subfigure}    	
		\end{minipage}
		\begin{minipage}[c]{0.06\textwidth}
			\includegraphics[width=\textwidth]{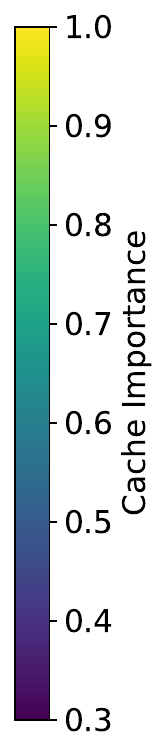}
		\end{minipage}
		\vspace{-0.2cm}
		\centering
		\caption{Mean vs. Defensive Aggregation: Against Fragility in Importance Stability}
		\label{fig:agg_assumption}
		\vspace{-0.5cm}
\end{figure}

\subsection{Fragility of Stability Assumption and  Vulnerability of Mean Aggregation}

\label{sc:assumption}
We examine this using Llama3.1-8B on the Government Report summarization task. Adopting the SOTA importance indicator, \( I_{j,i} =  A_{j,i} \times \text{norm}(v_i W_O) \), we observe each cache entry importance with 32 recent historical tokens \(q_j\). 
We then simulate a 50\% cache eviction using two different criteria. The first criterion uses importance scores from a single historical token observation, while the second uses scores averaged across all 32 historical tokens (mean aggregation). For each result, we track the proportion of total importance it retains during subsequent generation, relative to the full cache.

\textbf{Fragility of the Stability Assumption.} Figure~\ref{fig:agg_assumption_single} presents the results for the 16th historical token(see Appendix~\ref{apdx:more_assumption} for results from other tokens).  The results reveal a general, yet fragile, stability. On average, the 50\% retained cache subset accounts for 0.92 of the full cache's total importance during generation. However, this high average belies the underlying fragility: the retained importance can drop sharply, as seen in the interval between steps 150 and 320. In these moments of instability, the worst-case retained importance drops to as low as 0.34. Additionally, outliers—where the retained 50\% of the cache captures less than half of the total importance (value < 0.5)—are frequent, occurring in 89 instances in this trial alone.

\textbf{Vulnerability of Mean Aggregation.} Current eviction methods commonly employ mean aggregation over multiple importance observations. The rationale is to obtain an expected importance to guide eviction. However, by failing to account for worst cases, this strategy becomes vulnerable precisely under the fragile assumption, leading to outlier performance similar to that of using a single, unreliable token. As shown in Figure~\ref{fig:mean_agg_assumption}, significant drops persist at the same problematic steps observed in Figure~\ref{fig:agg_assumption_single}, reaching a worst-case importance value of 0.33 and resulting in 65 outlier instances.  This outcome is predictable. The observation score based on single historical token is inherently blind to the fragility of stability assumption; thus it cannot hedge against the worst-case risk. While simple averaging acts as a form of "reconciliation" among these individual observations to produce a moderate result, it does not incorporate any mechanism to control for this underlying risk. Consequently, it cannot aggregate a prediction to consistently outperform every single-token observation. When most single token-based observations fail, the mean-aggregated result is inevitably dragged down with them, thus offering no meaningful improvement and producing similarly damaging outliers.

This underscores a critical point: rather than focusing solely on designing more accurate importance indicators, it is equally—if not more—important to develop new aggregation methods explicitly designed for worst-case risk control, which can provide reliable estimates even when most single-token observations fail.

\begin{table}[t]
	\centering
	\caption{Defensive aggregation consistently improves the worst-case values across all task types.}
	\vspace{-0.3cm}
	\label{tab:fragility_tasks}
	\resizebox{\textwidth}{!}{%
		\begin{tabular}{lcccccc}
			\toprule
			Task type & Single-Doc. QA & Multi-Doc. QA & Summary & Few-shot & Synthetic & Code \\
			Dataset & NrtvQA & HotpotQA & GovReport & TREC & PCount & Lcc \\
			\midrule
			Mean aggregation & 0.44 & 0.39 & 0.28 & 0.47 & 0.47 & 0.30 \\
			Defensive aggregation & \textbf{0.62} & \textbf{0.60} & \textbf{0.52} & \textbf{0.61} & \textbf{0.61} & \textbf{0.50} \\
			\bottomrule
		\end{tabular}%
	}
	\vspace{-0.4cm}
\end{table}
\begin{algorithm}[t]
	\small
	\caption{Defensive Aggregation}
	\label{alg:defensiveagg}
	\begin{algorithmic}[1]
		\State \textbf{Input:} Importance scores $I \in \mathbb{R}^{m \times n}$, where $I_{j,i}$ is the importance of entry $i$ based on historical token $j$
		\State \textbf{Output:} Aggregated risk scores  $\tilde{R} \in \mathbb{R}^n$
		\State  $\tilde{R}_i = \max_{1 \leq j \leq m} I_{j,i} \:,\:  \forall i = 1, \ldots, n$  \Comment{ Worst-case Risk Estimation}
		\State   $
		R_i \;=\; \max\!\Big(\tilde{R}_i,\; \bar{R}\Big) \: \text{where} \: \bar{R} \;=\; \frac{1}{n}\sum_{i=1}^n{\tilde{R}_{i}}, \: \forall i = 1, \ldots, n$  \Comment{ Adaptive Prior-Risk Correction}
		\State \Return $R$
	\end{algorithmic}
\end{algorithm}
\vspace{-0.5cm}

{

\subsection{Defensive Aggregation via Worst-Case Risk Control}
\label{sc:defensive_agg}

Consider a cache entry may exhibits high importance in only a few single-token observations while remaining low in most others duo to fragile stability. Mean aggregation would not recognize this as important and would erroneously evict it. When this entry becomes crucial again in future generation, the prior eviction results in substantial importance loss. Thus, relying on mean aggregation fails to guard against these extreme cases. To address this, we introduce a novel \textit{defensive aggregation},  a novel strategy that eschews simple averaging in favor of a worst-case risk control perspective as shown in Algorithm~\ref{alg:defensiveagg}.

\textbf{Worst-case Risk Estimation.}  From a risk-control perspective, the penalty for evicting a KV cache entry is equivalent to the importance score it would have possessed at future moment. The "worst-case risk", $R_i^*$, is therefore the peak importance score an entry could attain over the entire future generation process. If we denote the future generated sequence as $L$, then $R_i^* = \max_{ t \in L} I_{t,i}$.  As this future maximum is unknowable at eviction time, we instead approximate it as the maximum importance score observed across all $j$ historical tokens, e.g. $\tilde{R_i} = \max_{1 \leq j \leq m} I_{j,i} \:,\:  \forall i = 1, \ldots, n$.~\footnote{For Grouped-Query Attention, a cache entry’s worst‑case risk estimate is the maximum importance score observed over historical tokens across all heads sharing its KV group.}. This \(O(n)\) procedure matches mean aggregation’s runtime yet yields significantly better empirical performance, as it better captures the potential worst-case risk if the entry were removed.

\textbf{Adaptive Prior-Risk Correction.}  Although the above estimator takes the maximum over observed history, it could still underestimate worst-case risk because eviction methods typically restrict the observation window (e.g., 32 tokens) to limit overhead. \footnote{Explicitly computing attention weights for all tokens is infeasible with FlashAttention optimization, and even storing all attention weights is prohibitively expensive (e.g., $\approx$64 GB for a 32k context in Llama-3.1-8B).} Such restricted observations could miss rare but critical risks. Inspired by Laplace smoothing in Bayesian estimation, we introduce an adaptive prior–risk correction. For each head, define a head-level prior risk $
\bar{R} \;=\; \frac{1}{n}\sum_{i=1}^n {\tilde{R}_{i}},$
i.e., the average observed worst-case risk across entries for that head. 
If the observed risk $R_i$  falls below  prior risk $\bar{R}$, we treat the shortfall as under-observation and substitute the prior: {$R_i \;=\; \max\!\Big(\tilde{R}_{i},\; \bar{R}\Big) \small$ }
. Thereby, heads with higher overall risk receive larger priors, reducing reliance on limited historical observations. The effectiveness of correction and its adaptive design is validated in Section~\ref{sc:abl}. 

By defending against risks of the fragile stability assumption, our defensive aggregation substantially improves worst-case performance compared to mean aggregation.
As shown in Figure \ref{fig:defensive_agg_assumption}, it boosts the worst-case retained importance from 0.33 to 0.61 and completely eliminates the 65 outlier instances produced by mean aggregation. Table \ref{tab:fragility_tasks} further confirms this advantage is consistent across six  datasets with different task types. Therefore, this simple two-operation method provides a crucial defense against the fragility of the importance stability assumption.

}

\subsection{Implementing DefensiveKV  Eviction Method with  Defensive Aggregation}

Building upon our proposed defensive aggregation strategy, we introduce two novel cache eviction methods: \emph{DefensiveKV} and \emph{Layer-DefensiveKV}.

\textbf{DefensiveKV} serves as the foundational variant. It directly integrates defensive aggregation into the traditional cache eviction workflow by replacing the conventional mean aggregation. Despite its simplicity, this modification alone leads to substantial performance improvements.

\textbf{Layer-DefensiveKV} further refines this by incorporating a layer-wise budget allocation, inspired by existing strategies~\citep{ada,pyramidkv}. It performs a joint selection of risky entries across layers, enabling more budget to be allocated to layers with more risky cache entries.

As shown in Algorithm~\ref{alg:defensivealg}, the overall process of DefensiveKV adheres to the established practices: preserving the KV cache entries of several recent historical tokens (Line 3) and then utilizing the query states of these tokens for importance measurement. The importance calculation begins with basic attention weights (Line 4) and incorporates further refinements—specifically pooling mechanisms from SnapKV (Line 5), and projected value norm scaling from CriticalKV (Line 7) \footnote{Although our method is based on current SOTA practice, defensive aggregation is widely applicable to other  eviction methods. Appendix~\ref{apdx:adakv_augment} includes a case study applying it to another baseline method for demonstration.}.

The key innovation in DefensiveKV is the strategic replacement of conventional mean aggregation  with our defensive aggregation (Line 6). This simple modification, requiring minimal changes, reduces over 2$\times$ in generation quality loss. The extension to Layer-DefensiveKV is also straightforward. 
It incorporates two additional refinements: first, projected value norms are normalized layer-wise to address their variance across layers (Line 11); second, risky entries are selected jointly across all layers (Line 12). This leads to an even more impressive  gain, with over 4× reduction in quality loss.
 
\begin{algorithm}[t]
	\small
	\caption{(Layer)-DefensiveKV}
	\label{alg:defensivealg}
	\begin{algorithmic}[1]
		\State \textbf{Input:} Cache Entries $K, V$, Parameter $W_O$,  queries of $m$ recent historical tokens  $Q = [q_{1},...,q_{m}]$
		\State \textbf{Output:} Retained KV Cache $\hat{K},\hat{V}$
		\State Append the KV cache of recent historical tokens $K[-m:,:],V[-m:,:]$ to $\hat{K},\hat{V}$.
		\State $A \leftarrow  \text{softmax}(QK^T/\sqrt{d})$
		\State $A \in \mathbb{R}^{m \times n} \leftarrow  \text{Pooling}(A, \text{dim}=-1)$  \Comment{Refined with pooling  by SnapKV~\citep{SnapKV}}
		\State $R \in \mathbb{R}^{n} \leftarrow  \text{Defensive Aggregation} \:  \text{Algorithm}~\ref{alg:defensiveagg} \:(A)$ \Comment{Our modification}
		\State $ R_i \leftarrow R_i \times \text{norm}(v_i W_O) \quad \forall v_i \in V, i = 1, \ldots, n$ \Comment{Refined with norm by CriticalKV~\citep{feng2025identifycriticalkvcache}} 
		\If{without layer-wise budget allocation} \Comment{Leading to DefensiveKV}  
		\State Select the cache entries with top worst-case risk $R$ independently in each layer  and append to   $\hat{K},\hat{V}$  
		\Else  \Comment{Leading to Layer-DefensiveKV}  
		\State $R_i \leftarrow R_i / \sum_i \text{norm}(v_i W_O) \quad \forall v_i \in V, i = 1, \ldots, n$  
		\State Select the cache entries with top worst-case risk $R$ jointly across all layers and append to   $\hat{K},\hat{V}$  
		\EndIf  
		\State \Return $\hat{K}, \hat{V}$
	\end{algorithmic}
\end{algorithm}

\section{Experiments}\label{sec:experiments}
\subsection{Experimental Settings}

\textbf{Models.} We evaluate our approach on three open-source LLMs: Llama-3.1-8B-Instruct \citep{llama3,llama2} and Qwen2.5-32B-Instruct~\citep{qwen2.5}, supporting context lengths of up to 128K, and Mistral-7B-Instruct-v0.3~\citep{jiang2023mistral}, supporting up to 32K.

 \textbf{Baselines.}
 We compare our method against six baselines. StreamingLLM~\citep{streamingllm} is an early sliding window approach. SnapKV \citep{SnapKV}, AdaKV \citep{ada}, and CAKE \citep{qin2025cake} use attention weight-coupled pooling for importance indicators; CAKE also employs a cascaded architecture for layer-wise budget allocation. The SOTA CriticalKV \citep{feng2025identifycriticalkvcache} introduces a more accurate importance indicator.  DuoAttention \citep{duo}, a training-based method, is included with official configurations for Llama-3.1-8B and Mistral-7B-v0.3, but marked N/A for Qwen2.5-32B due to unavailable configurations and high training costs. All methods use a historical window size of 32 and are accelerated with FlashAttention-2\citep{fa1,fa2}.

\textbf{Settings.} 
Following the settings in~\citep{kvpress, feng2025identifycriticalkvcache}, the context is compressed independently before  question is introduced. This better simulates practical  scenarios (e.g., multi-turn QA or prefixed contexts) where multiple questions often pertain to the same context, or the question is unavailable during context compression. Thus this setup is more challenging and better reflects the real-world performance of cache eviction methods.~\citep{feng2025identifycriticalkvcache}

\begin{figure}[t]
	\vspace{-0.2cm}
	\begin{minipage}{0.99\textwidth}
		\centering
		\includegraphics[width=0.6\textwidth]{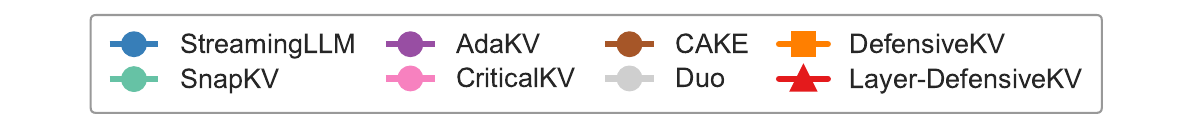}
	\end{minipage}
	\begin{subfigure}{0.33\textwidth}
		\centering
		\includegraphics[width=\linewidth]{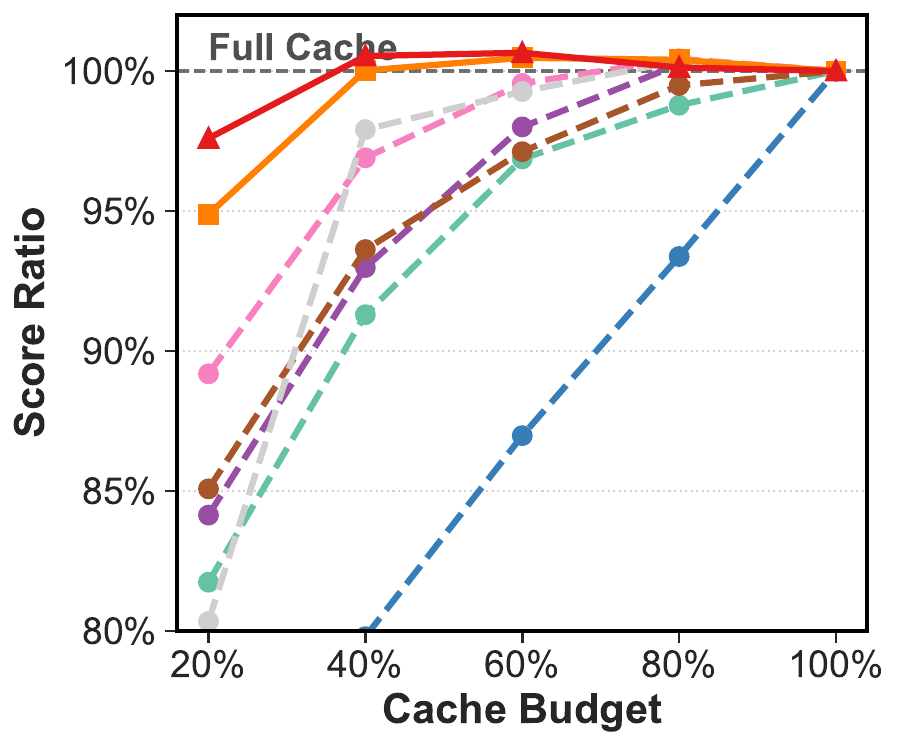}
		\vspace{-0.6cm}
		\caption{Meta-Llama-3.1-8B-Instruct}
		\vspace{-0.2cm}
		\label{fig:meta_llama_longbench}
	\end{subfigure}
	\begin{subfigure}{0.33\textwidth}
		\centering
		\includegraphics[width=\linewidth]{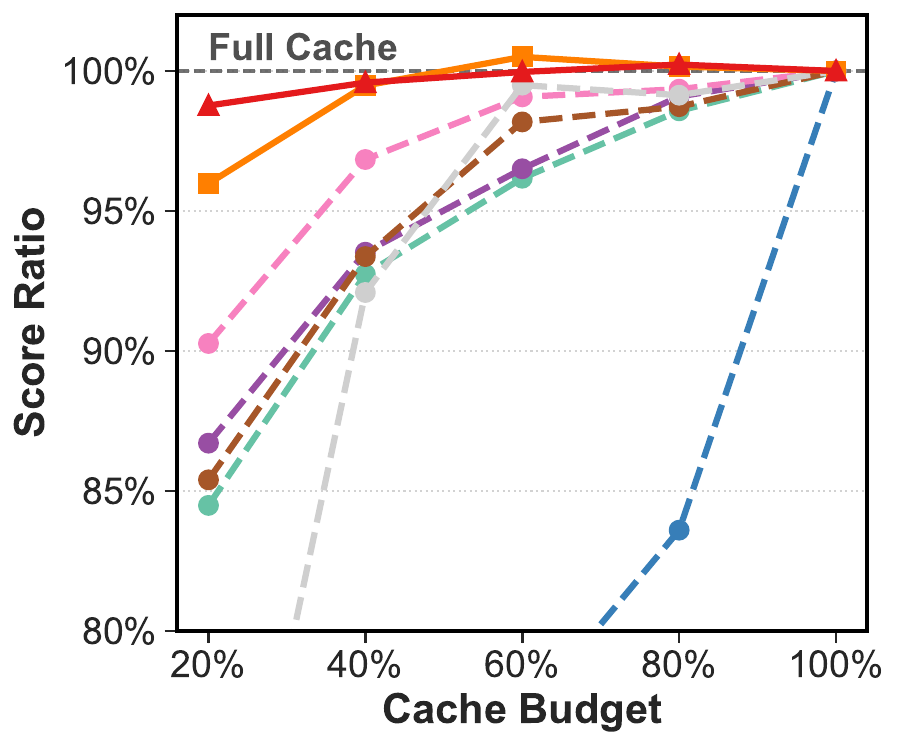}
		\vspace{-0.6cm}
		\caption{Mistral-7B-Instruct-v0.3}
		\vspace{-0.2cm}
		\label{fig:mistral_longbench}
	\end{subfigure}
	\begin{subfigure}{0.33\textwidth}
		\centering
		\includegraphics[width=\linewidth]{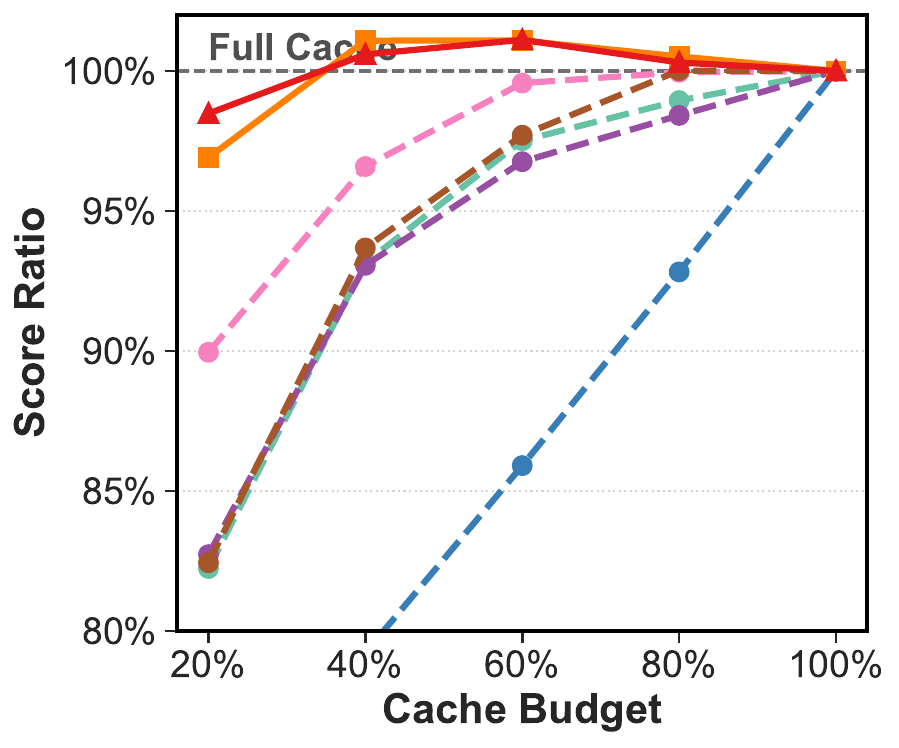}
			\vspace{-0.6cm}
		\caption{Qwen2.5-32B-Instruct}
		\vspace{-0.2cm}
		\label{fig:qwen_longbench}
	\end{subfigure}
	\vspace{-0.5cm}
	\caption{
		Overview of averaged generation quality across 16 datasets on LongBench.
	}
	\label{fig:longbench_comparison}
\end{figure}

\begin{figure}[t]
	\vspace{-0.4cm}
	\centering
	\begin{subfigure}{0.32\textwidth}
		\centering
		\includegraphics[width=\linewidth]{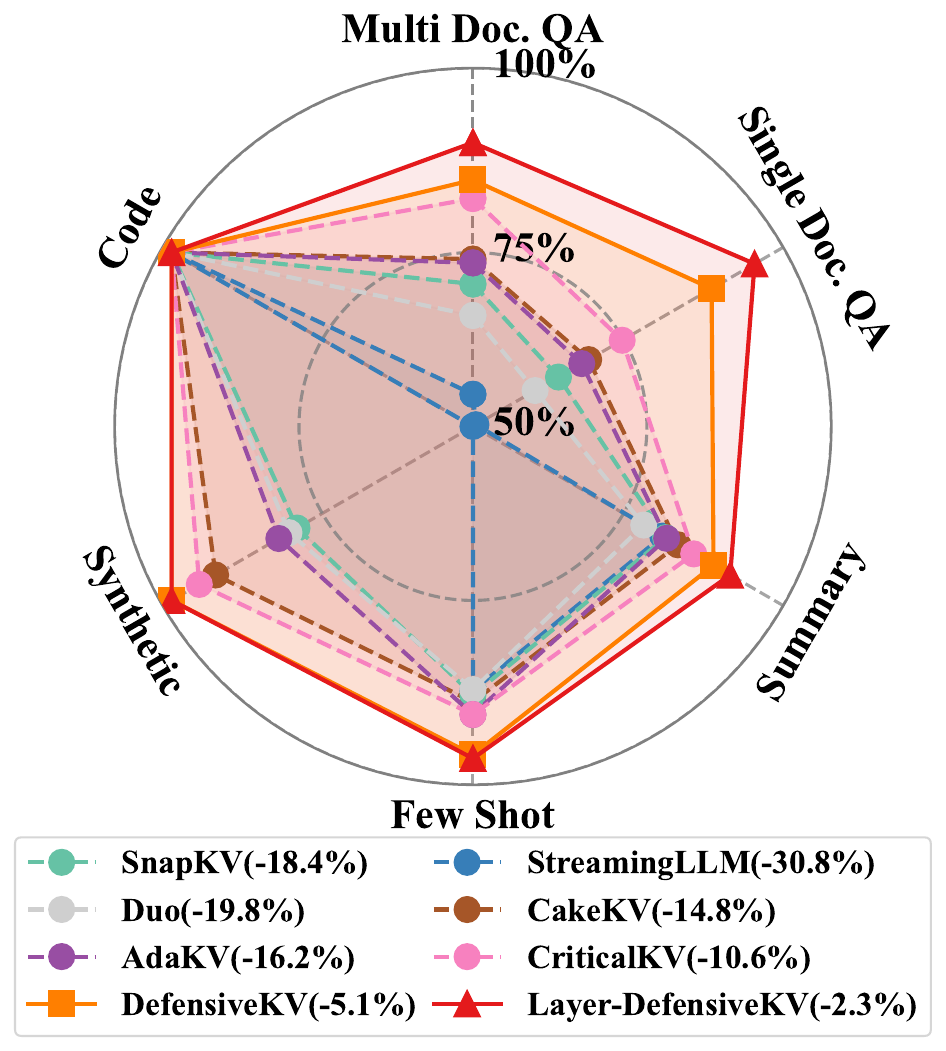}
		\vspace{-0.5cm}
		\caption{20\% Cache Size}
		\label{fig:llama_20}
	\end{subfigure}
	\begin{subfigure}{0.32\textwidth}
		\centering
		\includegraphics[width=\linewidth]{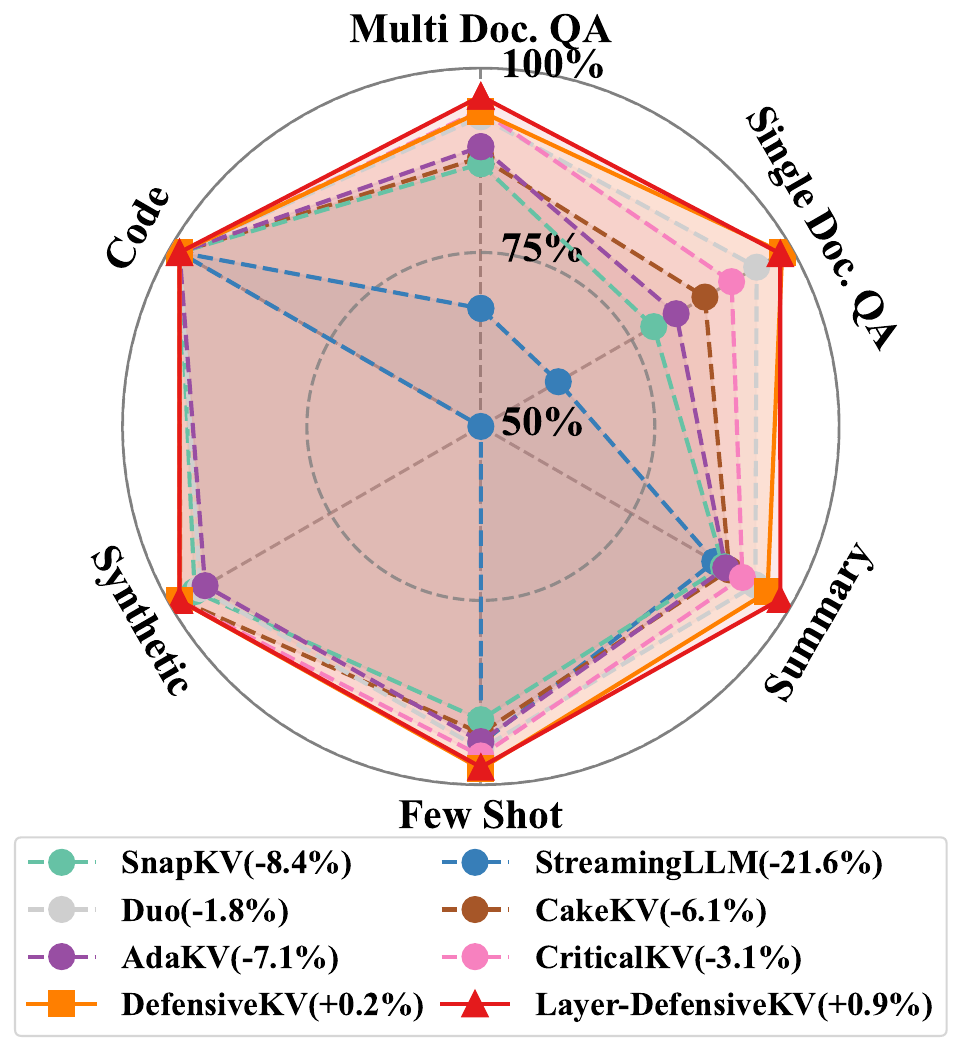}
		\vspace{-0.5cm}
		\caption{40\% Cache Size}
		\label{fig:llama_40}
	\end{subfigure}
	\begin{subfigure}{0.32\textwidth}
		\centering
		\includegraphics[width=\linewidth]{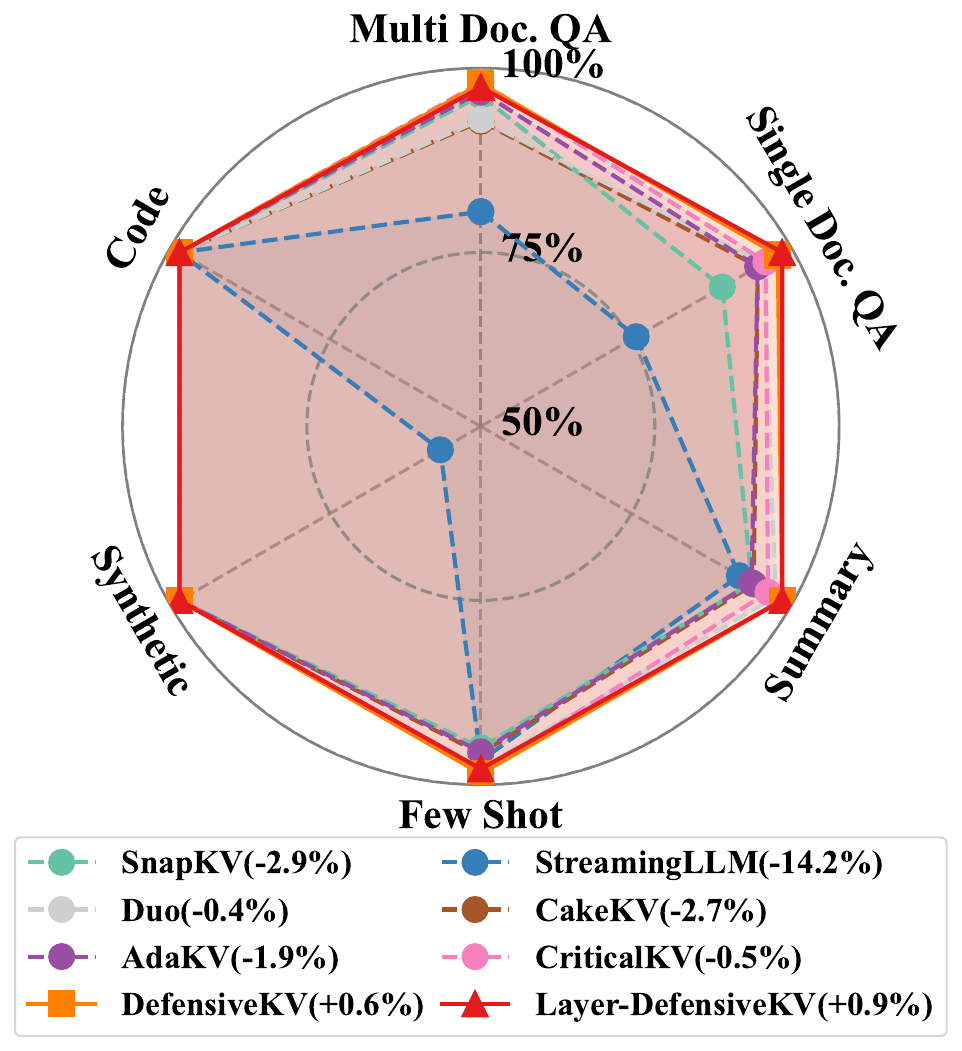}
		\vspace{-0.5cm}
		\caption{60\% Cache Size}
		\label{fig:llama_60}
	\end{subfigure}
	\vspace{-0.3cm}
	\caption{
		Analysis of the six task domains on LongBench for Meta-Llama-3.1-8B-Instruct.
	}
	\vspace{-0.5cm}
	\label{fig:domain_comparison}
\end{figure}

\subsection{LongBench Evaluation}
LongBench~\citep{longbench} serves as a comprehensive benchmark, featuring 16 datasets structured into six task domains: single-document QA, multi-document QA, summary, few-shot learning, synthetic tasks, and code completion. Detailed information for each dataset can be found in Appendix~\ref{apdx:details_datasets}

\textbf{{Overall} Analysis.}
Figure~\ref{fig:longbench_comparison} illustrates our methods' significant advantages in average quality loss across 16 datasets. As cache size drops from 100\% to 40\%, all baselines degrade noticeably, while our DefensiveKV and Layer-DefensiveKV remain nearly lossless. For instance, with a 40\% cache on Llama-3.1-8B (Figure~\ref{fig:meta_llama_longbench}), CriticalKV (best baseline) loses 3.1\% quality, whereas our {DefensiveKV} shows no degradation, surpassing even the training-based DuoAttention (2.2\% drop, despite offline training costs). At a smaller 20\% cache, CriticalKV's loss is 10.6\%, while {DefensiveKV} limits it to 5.1\%(over 2x reduction). Our {Layer-DefensiveKV} further cuts this loss to 2.3\%(over 4x reduction). Similar advantages hold for other models. For instance, on Mistral-7B with 20\% cache, {DefensiveKV} and {Layer-DefensiveKV} achieve 4.0\% and 1.3\% loss, respectively, far below CriticalKV's 9.7\%.

\begin{table*}[!t]
	\vspace{-0.2cm}
	\centering
	\caption{Detailed scores of 16 datasets on LongBench.}
	\vspace{-0.3cm}
	\label{longbenchdetail}
	\resizebox{\textwidth}{!}{%
		\setlength{\tabcolsep}{1.5pt}
		\begin{tabular}{@{}lllllllllllllllllllllllll@{}}
			\toprule
			
			\multirow{2}{*}[-20pt]{\makecell[l]{\hspace{10pt}\raisebox{0pt}{Method}}} & \multicolumn{3}{c}{Single-Document QA} & \multicolumn{3}{c}{Multi-Document QA} & \multicolumn{3}{c}{Summarization} & \multicolumn{3}{c}{Few-shot Learning} & \multicolumn{2}{c}{Synthetic} & \multicolumn{2}{c}{Code} & \multirow{2}{*}[-20pt]{Avg.} \\ \cmidrule(lr){2-4} \cmidrule(lr){5-7} \cmidrule(lr){8-10} \cmidrule(lr){11-13} \cmidrule(lr){14-15} \cmidrule(lr){16-17}
			
			&\rotatebox{30}{NrtvQA} & \rotatebox{30}{Qasper} & \rotatebox{30}{MF-en} & \rotatebox{30}{Hotpot}& \rotatebox{30}{2WikiQA} & \rotatebox{30}{Musique} & \rotatebox{30}{GovRep} & \rotatebox{30}{QMSum} & \rotatebox{30}{MultiNews} & \rotatebox{30}{TREC} & \rotatebox{30}{TriviaQA} & \rotatebox{30}{SAMSum} & \rotatebox{30}{PCount} & \rotatebox{30}{PR-en} & \rotatebox{30}{Lcc} & \rotatebox{30}{RB-P} & \\
			
			\midrule
			\multicolumn{18}{c}{Llama-3.1-8B-Instruct, $20\%$ Cache Size}\\
			\midrule
			Full Cache & 29.55 & 44.68 & 55.82 & 57.59 & 48.89 & 32.61 & 34.40 & 25.51 & 26.83 & 73.00 & 92.36 & 43.27 & 7.38 & 99.50 & 63.44 & 52.36 & 49.20 \\
			\arrayrulecolor{lightgray}
			\midrule
			\arrayrulecolor{black}
			DuoAttention & 23.28 & 21.22 & 34.03 & 42.89 & 28.14 & 20.57 & 25.32 & 19.48 & 23.12 & 56.00 & 86.54 & 40.67 & 7.50 & 78.50 & 65.94 & 59.19 & 39.52 \\
			StreamingLLM & 22.05 & 19.83 & 23.87 & 39.44 & 20.97 & 15.46 & 27.76 & 20.63 & 22.27 & 53.50 & 89.97 & 40.04 & 4.00 & 29.50 & 65.61 & \underline{\textbf{60.66}} & 34.72 \\
			SnapKV & 25.64 & 28.23 & 29.71 & 46.17 & 29.64 & 22.07 & 27.09 & 21.51 & 22.46 & 48.50 & \textbf{92.21} & \textbf{44.08} & 5.08 & 79.50 & \underline{67.17} & 54.36 & 40.21 \\
			CAKE & 26.29 & 30.54 & 33.28 & 46.03 & 32.08 & 24.73 & 27.77 & 22.16 & 22.91 & 51.50 & \underline{91.86} & 43.56 & 6.50 & 92.50 & 65.46 & 52.50 & 41.85 \\
			AdaKV & 27.07 & 28.69 & 32.85 & 49.64 & 30.89 & 21.57 & 26.70 & 21.85 & 22.67 & 55.50 & 91.30 & \underline{43.89} & 7.30 & 80.50 & 66.44 & 55.43 & 41.39 \\
			CriticalKV & 29.81 & 32.58 & 34.96 & \underline{52.34} & 36.24 & 26.37 & 28.35 & 23.52 & 23.24 & 56.50 & 90.80 & 43.37 & \textbf{8.89} & 93.00 & 67.05 & 54.99 & 43.88 \\
			DefensiveKV & \underline{29.97} & \underline{40.46} & \underline{46.23} & 52.20 & \underline{38.40} & \textbf{28.06} & \underline{29.96} & \underline{23.89} & \underline{24.11} & \underline{68.00} & 91.58 & 43.17 & 8.28 & \textbf{100.00} & 67.17 & 55.40 & \underline{46.68} \\
			Layer-DefensiveKV & \textbf{30.10} & \textbf{42.91} & \textbf{52.94} & \textbf{55.03} & \textbf{44.07} & \underline{27.00} & \textbf{30.99} & \textbf{24.95} & \textbf{24.42} & \textbf{69.00} & 91.30 & 43.54 & \underline{8.38} & \textbf{100.00} & \textbf{67.60} & 56.00 & \textbf{48.01} \\
			
			\midrule
			\multicolumn{18}{c}{Mistral-7B-Instruct-v0.3, $20\%$ Cache Size}\\
			\midrule
			Full Cache & 27.02 & 38.19 & 50.22 & 50.75 & 37.41 & 27.92 & 34.45 & 25.76 & 26.37 & 76.00 & 89.01 & 46.89 & 6.50 & 97.00 & 66.04 & 60.47 & 47.50 \\
			\arrayrulecolor{lightgray}
			\midrule
			\arrayrulecolor{black}
			DuoAttention & 11.91 & 13.58 & 29.88 & 31.73 & 22.43 & 9.18 & 23.96 & 17.25 & 22.67 & 49.50 & 86.08 & 43.08 & 2.67 & 18.00 & 59.89 & 56.23 & 31.13 \\
			StreamingLLM & 18.30 & 16.38 & 26.26 & 38.78 & 25.99 & 15.06 & 28.00 & 20.73 & 21.32 & 30.50 & 80.88 & 40.57 & 3.00 & 28.00 & 32.62 & 46.91 & 29.58 \\
			SnapKV & 21.91 & 23.69 & 30.59 & 43.71 & 28.28 & 19.81 & 27.91 & 21.15 & 22.15 & 55.00 & 89.41 & \underline{46.67} & \textbf{5.00} & 82.50 & 64.30 & 59.97 & 40.13 \\
			CAKE & 23.08 & 25.42 & 35.31 & 44.10 & 28.96 & 18.59 & 28.27 & 21.18 & 22.61 & 60.00 & \textbf{90.36} & 46.40 & 4.00 & 77.00 & 64.50 & 59.21 & 40.56 \\
			AdaKV & 24.00 & 26.29 & 31.15 & 45.26 & 27.99 & 20.65 & 27.37 & 21.67 & 22.38 & 59.00 & 89.87 & 46.27 & \underline{4.50} & 88.00 & 65.05 & 59.50 & 41.18 \\
			CriticalKV & \underline{24.14} & 29.56 & 38.91 & 45.42 & 32.08 & 21.26 & 28.59 & 22.71 & 23.11 & 65.50 & \underline{90.13} & 46.65 & 4.11 & 90.00 & 65.42 & 58.43 & 42.88 \\
			DefensiveKV & 21.05 & \underline{34.67} & \textbf{50.05} & \underline{48.76} & \underline{32.27} & \textbf{26.09} & \underline{31.95} & \underline{23.39} & \underline{24.05} & \underline{72.50} & 90.11 & \textbf{46.80} & 3.53 & \underline{96.50} & \underline{65.78} & \textbf{61.93} & \underline{45.59} \\
			Layer-DefensiveKV & \textbf{27.31} & \textbf{39.41} & \underline{49.70} & \textbf{49.89} & \textbf{37.82} & \underline{24.16} & \textbf{33.13} & \textbf{25.08} & \textbf{25.49} & \textbf{74.50} & 89.61 & 46.25 & 3.06 & \textbf{97.00} & \textbf{66.99} & \underline{61.21} & \textbf{46.91} \\
			
			\midrule
			\multicolumn{18}{c}{Qwen2.5-32B-Instruct, $20\%$  Cache Size}\\
			\midrule
			Full Cache & 30.88 & 46.13 & 52.87 & 63.59 & 59.75 & 38.78 & 32.59 & 24.35 & 24.95 & 72.00 & 88.26 & 47.05 & 12.50 & 100.00 & 49.64 & 34.24 & 48.60 \\
			\arrayrulecolor{lightgray}
			\midrule
			\arrayrulecolor{black}
			DuoAttention & N/A & N/A & N/A & N/A & N/A & N/A & N/A & N/A & N/A & N/A & N/A & N/A & N/A & N/A & N/A & N/A & N/A \\
			StreamingLLM & 20.74 & 17.68 & 25.04 & 39.53 & 33.19 & 17.38 & 27.83 & 19.04 & 21.67 & 61.00 & 82.01 & 42.80 & \underline{10.67} & 29.00 & \underline{\textbf{56.20}} & 30.14 & 33.37 \\
			SnapKV & 24.52 & 24.58 & 28.80 & 52.54 & 42.07 & 29.74 & 28.25 & 19.45 & 21.35 & 58.00 & 87.74 & \underline{48.21} & 9.00 & 76.35 & 53.19 & 35.58 & 39.96 \\
			CAKE & 22.25 & 26.49 & 31.33 & 49.20 & 42.48 & 28.11 & 27.86 & 18.92 & 22.02 & 58.00 & 87.77 & 47.22 & \textbf{11.00} & 81.25 & 51.82 & 35.25 & 40.06 \\
			AdaKV & 25.49 & 22.51 & 29.13 & 54.11 & 41.24 & 29.98 & 28.01 & 19.33 & 21.81 & 61.50 & 88.09 & 48.02 & 9.00 & 74.00 & 55.84 & 35.17 & 40.20 \\
			CriticalKV & 29.65 & 25.93 & 32.51 & 58.92 & 48.60 & 34.54 & 29.77 & 20.86 & 22.23 & 65.00 & 88.44 & \textbf{48.65} & 10.50 & 94.75 & 53.95 & 35.14 & 43.72 \\
			DefensiveKV & \underline{31.11} & \underline{32.11} & \underline{45.66} & \underline{62.55} & \underline{57.68} & \textbf{40.55} & \underline{30.30} & \underline{22.06} & \underline{22.98} & \underline{71.00} & \textbf{88.96} & 48.05 & 9.50 & \underline{99.75} & 54.29 & \textbf{36.98} & \underline{47.10} \\
			Layer-DefensiveKV & \textbf{31.38} & \textbf{35.65} & \textbf{49.25} & \textbf{64.24} & \textbf{58.86} & \underline{40.25} & \textbf{30.86} & \textbf{23.07} & \textbf{23.10} & \textbf{75.00} & \underline{88.92} & 47.47 & 10.00 & \textbf{99.88} & 52.21 & \underline{35.62} & \textbf{47.86} \\
			
			\bottomrule
		\end{tabular}%
	}
	\vspace{-0.4cm}
\end{table*}

\textbf{Task Analysis.}
Figure~\ref{fig:domain_comparison} displays Llama-3.1-8B average scores by task domain (Appendix~\ref{apdx:more_quality} for more models). DefensiveKV and Layer-DefensiveKV consistently excel. While simpler task domains (Code, Synthetic) show high performance for most methods, challenging ones (Doc QA, Summarization) reveal significant baseline degradation under reduced cache. Our methods maintain their advantages. For instance, in Single-doc QA (20\% cache), CriticalKV (strongest baseline) drops to 74.8\% of full-cache quality; DefensiveKV achieves 89.6\%, and Layer-DefensiveKV reaches 96.7\%. Table~\ref{longbenchdetail} further reports detailed 20\% cache scores (other results in Appendix~\ref{apdx:details_longbench}).
On Llama-3.1-8B (20\% cache), DefensiveKV beats CriticalKV on 13/16 datasets; Layer-DefensiveKV wins on 15/16. Such a  significant performance advantage, rarely observed between other baselines, highlights the effectiveness  of our  ``worst-case risk''  perspective to against underlying fragility across diverse tasks.

\subsection{Needle-in-a-Haystack Evaluation}
\label{sc:needle}
In the Needle-in-a-Haystack test, the key sentence is placed in a long context to evaluate retrieval ability. Following Ruler~\citep{hsieh2024ruler}, we test two representative  cases with a 32K context length: (1) Single-retrieval: one needle is randomly inserted for retrieval. (2) Multi-retrieval: four needles are randomly inserted and all need to be retrieved. Further details, along with evaluations on more  "needle-in-a-haystack-style" tasks from Ruler, are provided in Appendix~\ref{apdx:more_ruler}.

As shown in Figure~\ref{fig:needle_comparison}, our {DefensiveKV} and {Layer-DefensiveKV} achieve significantly higher scores across all settings. For instance, on long-context models like Llama-3.1-8B and Qwen2.5-32B (both supporting 128K context length), our methods maintain near-lossless, with scores 194 and 193 for Llama-3.1-8B at mere 10\% cache size. In contrast, even the strongest baseline, CriticalKV, drops to 140 under the same conditions, while others fall below 100—demonstrating a substantial gap. On weaker long-context ability model, i.e., Mistral-7B (maximum context length 32K), all baselines suffer severe performance degradation. At a 10\% cache size, most baselines score below 6, and CriticalKV only reaches 28. However, our {DefensiveKV} and {Layer-DefensiveKV} achieve scores of 139 and 161, over 5x and 5.8x improvements, respectively.

\begin{figure}[t]
	\centering
		\centering
		\vspace{-0.15cm}
		\includegraphics[width=1.0\linewidth]{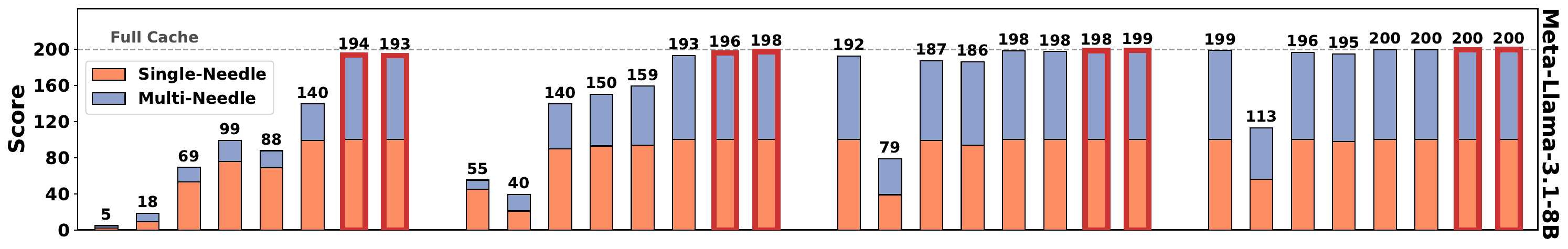}
	\vspace{-0.15cm}
		\centering
		\includegraphics[width=1.0\linewidth]{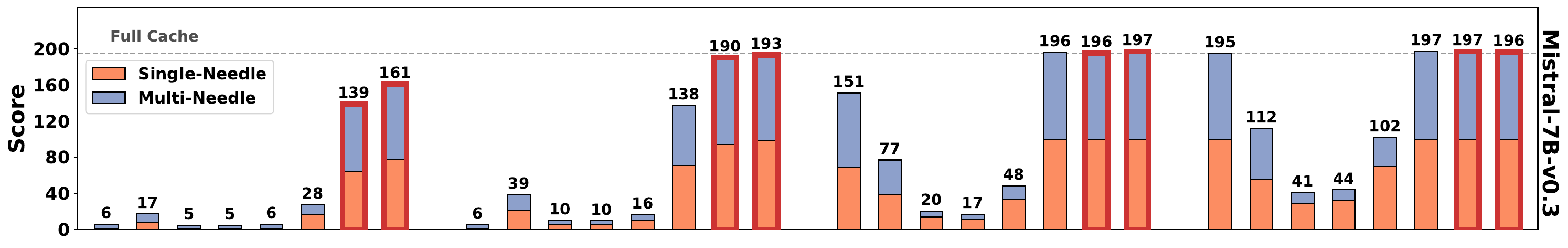}
		\vspace{-0.15cm}
		\centering
		\includegraphics[width=1.0\linewidth]{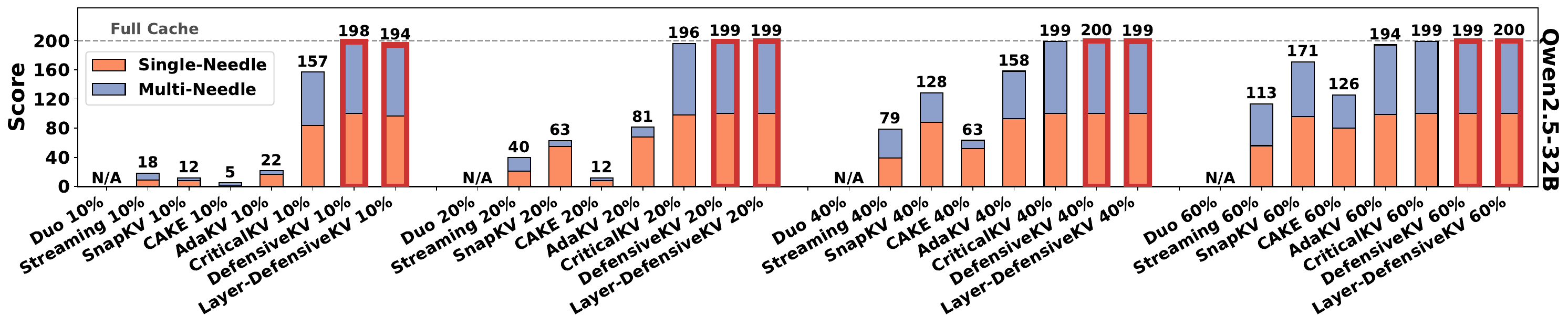}
		\vspace{-0.7cm}
	\caption{
		Evaluations on the needle-in-a-haystack tasks.
	}
	\vspace{-0.5cm}
	\label{fig:needle_comparison}
\end{figure}

\begin{figure}[t]
\begin{minipage}{0.33\linewidth}
	\centering
	\includegraphics[width=0.99\textwidth]{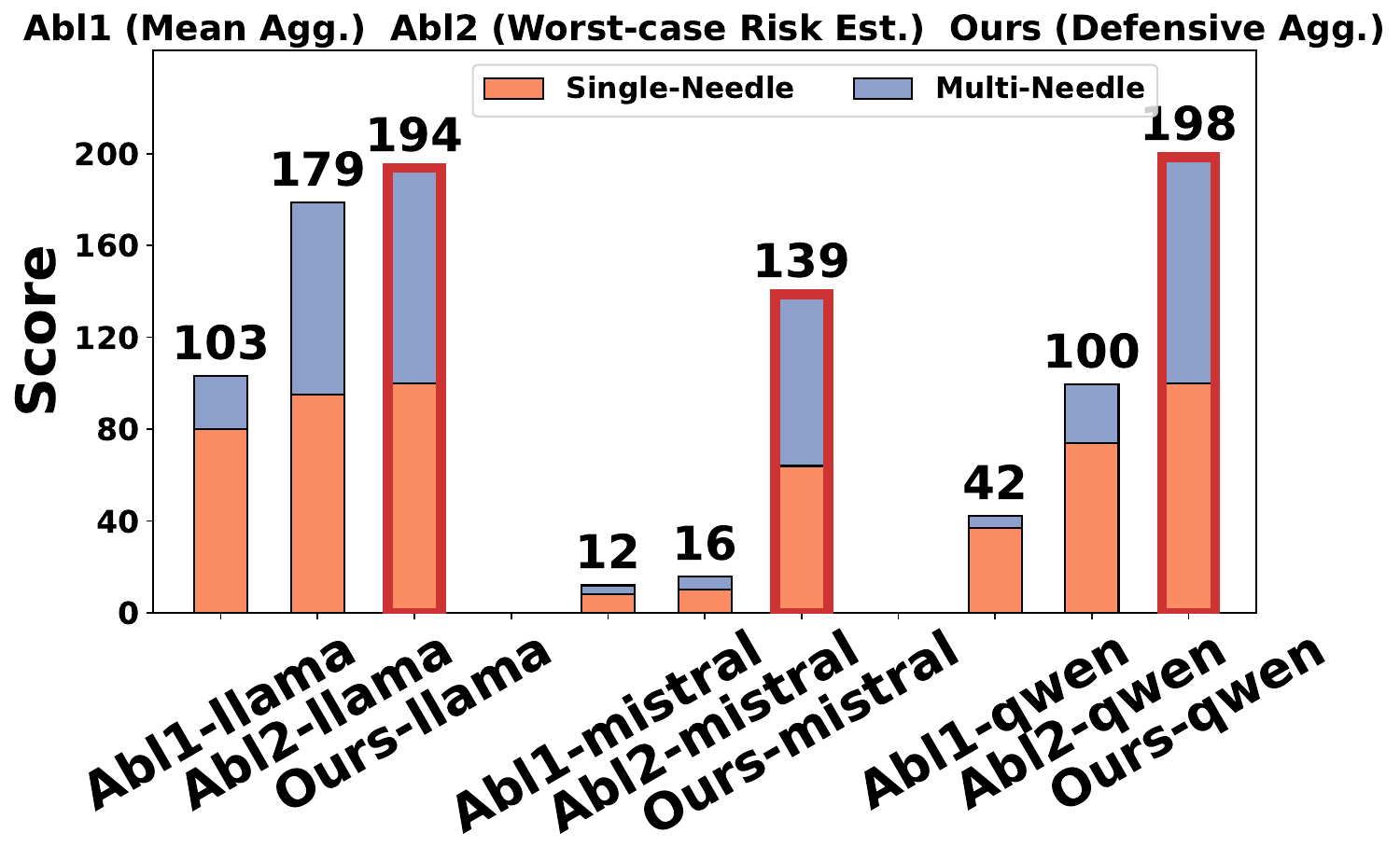}
	\vspace{-0.5cm}
	\caption{
		Ablation 10\% cache.
	}
	\vspace{-0.4cm}
	\label{fig:ablation}
\end{minipage}
\begin{minipage}{0.31\linewidth}
	\centering
	\includegraphics[width=0.99\textwidth]{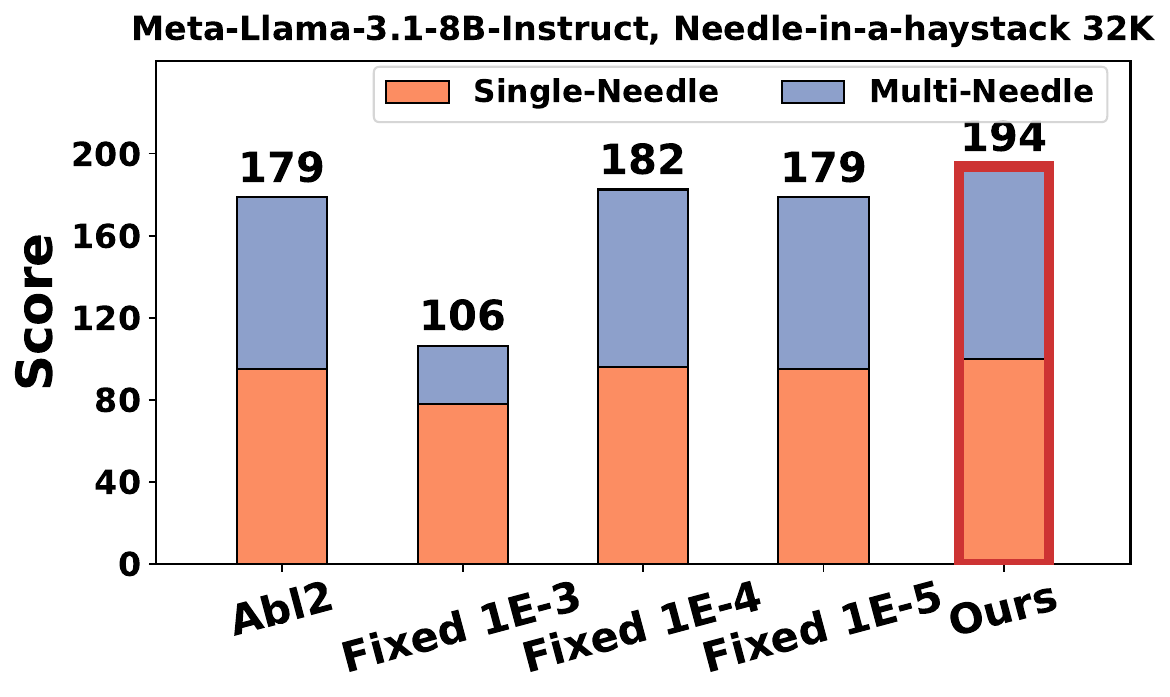}
		\vspace{-0.4cm}
	\caption{
		Adaptive Correction
	}
	\vspace{-0.4cm}
	\label{fig:adaptive_correction}
\end{minipage}
\begin{minipage}{0.36\linewidth}
	\centering
	\includegraphics[width=0.9\linewidth]{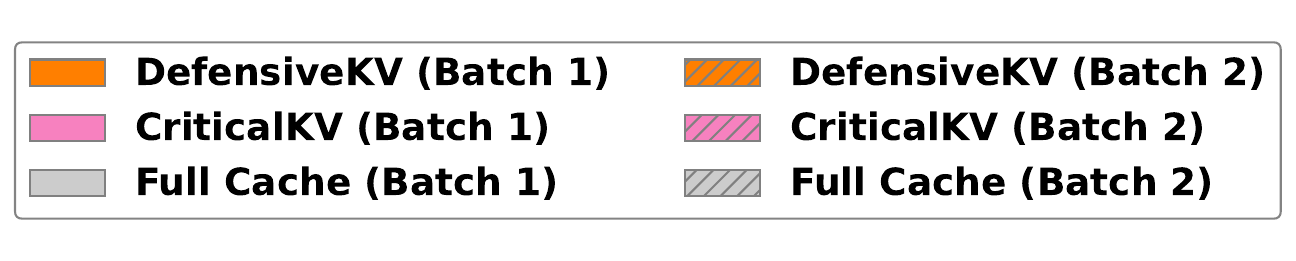}
	\begin{subfigure}{0.49\linewidth}
		\centering
		\vspace{-0.4cm}
		\includegraphics[width=\linewidth]{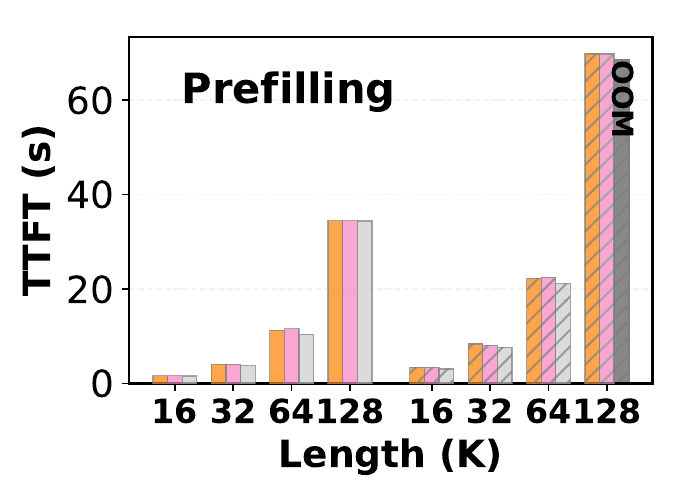}
	\end{subfigure}
	\hspace{-0.2cm}
	\begin{subfigure}{0.49\linewidth}
		\centering
			\vspace{-0.4cm}
		\includegraphics[width=\linewidth]{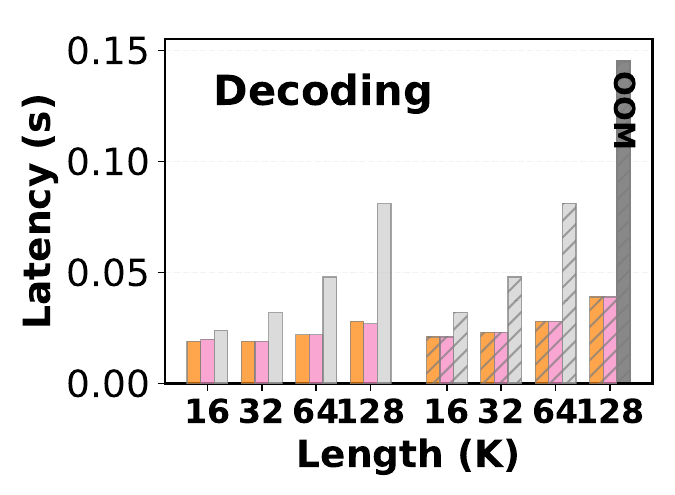}
	\end{subfigure}
	\vspace{0.2cm}
	\caption{\centering Efficiency(FlashAttn2)}
	\label{fig:efficiency}
\end{minipage}

\vspace{-0.2cm}

\end{figure}

\subsection{Ablation Studies}
\label{sc:abl}
To analyze the roles of the two key operations in our defensive aggregation—worst-case risk estimation and adaptive prior-risk correction—we conduct ablation studies based on our DefensiveKV method. First, we only remove adaptive prior-risk correction (denoted as Abl2). Then, we further ablate the worst-case risk estimation by replacing with common mean aggregation (denoted as Abl1). As shown in Figure~\ref{fig:ablation}, using only worst-case risk estimation (Abl2) already significantly outperforms that are with mean aggregation. For example, on Llama-3.1-8B, Abl2 improves the score from 103 (Abl1) to 179.
Adding our adaptive prior-risk correction provides further gains, with our full DefensiveKV method reaching a score of 194. These results confirm that both operations contribute meaningfully to the overall performance.

To validate the adaptive design of our prior-risk correction, we ablated it against fixed correction thresholds (1E-3, 1E-4, 1E-5). The results in Figure~\ref{fig:adaptive_correction} show that fixed thresholds are ineffective. Most fail to outperform the no-correction baseline score of 179 (Abl2), with the 1E-4 case providing only a marginal gain to 182. Our adaptive correction, however, reaches a score of 194, confirming that tailoring the correction to each head's risk profile is crucial.  Additionally, the hyperparameter-free nature of our adaptive design ensures consistently strong performance across two additional models (shown as Abl2 vs. ours in Figure~\ref{fig:ablation}).

\subsection{Efficiency Test}
\label{sc:effi}
We compare DefensiveKV and CriticalKV, which differ only in their aggregation mechanisms, to demonstrate that defensive aggregation introduces negligible computational overhead.
Our experiments, conducted on an 80GB A100 GPU with Llama-3.1-8B (20\% cache), show in Figure~\ref{fig:efficiency} that DefensiveKV and CriticalKV have nearly identical time-to-first-token (TTFT) and decoding latency. All KV cache eviction occurs during the prefilling stage and is included in TTFT, confirming that DefensiveKV adds negligible overhead. Additionally, cache eviction significantly reduces decoding latency versus Full Cache: e.g., for batch size 1 and a 128K context length, latency drops from 0.081s (Full Cache) to ~0.028s with eviction-based methods (a 2.9x speedup). Furthermore, cache eviction allows larger batch sizes; for example, eviction methods can handle batch size of 2, while Full Cache results in out-of-memory errors, leading to a 4.2× decoding throughput boost. See Appendix~\ref{apdx:memory} for memory usage details.

\section{Conclusion}

This work challenges the fragile stability assumption underlying existing KV cache eviction methods. 
We show that widely used mean aggregation strategies are highly vulnerable under the fragile stability, resulting in poor worst-case performance. To address this, we propose ``defensive aggregation'', a novel strategy explicitly designed from a ``worst-case risk'' perspective with negligible computational overhead.
Based on this, we investigate DefensiveKV and its layer-aware variant, Layer-DefensiveKV, both of which achieve significant improvements over state-of-the-art methods across comprehensive evaluations.
Our work pioneers a new research direction by emphasizing the ``worst-case risk''-aware aggregation to mitigate the often-overlooked 
fragility in cache eviction—a critical yet underexplored component of efficient LLM inference.  
We hope these contributions pave the way for more effective cache eviction methods, which are essential for advancing LLM inference.

\bibliographystyle{iclr2026_conference}
\bibliography{icml25}

\begin{thebibliography}{54}
\providecommand{\natexlab}[1]{#1}
\providecommand{\url}[1]{\texttt{#1}}
\expandafter\ifx\csname urlstyle\endcsname\relax
  \providecommand{\doi}[1]{doi: #1}\else
  \providecommand{\doi}{doi: \begingroup \urlstyle{rm}\Url}\fi

\bibitem[lla(2024)]{llama3}
Llama-3-8b-instruct-gradient-1048k, 2024.
\newblock URL
  \url{https://huggingface.co/gradientai/Llama-3-8B-Instruct-Gradient-1048k}.

\bibitem[Bai et~al.(2023)Bai, Lv, Zhang, Lyu, Tang, Huang, Du, Liu, Zeng, Hou,
  et~al.]{bai2023longbench}
Yushi Bai, Xin Lv, Jiajie Zhang, Hongchang Lyu, Jiankai Tang, Zhidian Huang,
  Zhengxiao Du, Xiao Liu, Aohan Zeng, Lei Hou, et~al.
\newblock Longbench: A bilingual, multitask benchmark for long context
  understanding.
\newblock \emph{arXiv preprint arXiv:2308.14508}, 2023.

\bibitem[Bai et~al.(2024)Bai, Lv, Zhang, Lyu, Tang, Huang, Du, Liu, Zeng, Hou,
  Dong, Tang, and Li]{longbench}
Yushi Bai, Xin Lv, Jiajie Zhang, Hongchang Lyu, Jiankai Tang, Zhidian Huang,
  Zhengxiao Du, Xiao Liu, Aohan Zeng, Lei Hou, Yuxiao Dong, Jie Tang, and
  Juanzi Li.
\newblock Longbench: A bilingual, multitask benchmark for long context
  understanding, 2024.
\newblock URL \url{https://arxiv.org/abs/2308.14508}.

\bibitem[Dao(2023)]{fa2}
Tri Dao.
\newblock Flashattention-2: Faster attention with better parallelism and work
  partitioning.
\newblock \emph{arXiv preprint arXiv:2307.08691}, 2023.

\bibitem[Dao et~al.(2022)Dao, Fu, Ermon, Rudra, and R{\'e}]{fa1}
Tri Dao, Dan Fu, Stefano Ermon, Atri Rudra, and Christopher R{\'e}.
\newblock Flashattention: Fast and memory-efficient exact attention with
  io-awareness.
\newblock \emph{Advances in Neural Information Processing Systems},
  35:\penalty0 16344--16359, 2022.

\bibitem[Dasigi et~al.(2021)Dasigi, Lo, Beltagy, Cohan, Smith, and
  Gardner]{dasigi2021dataset}
Pradeep Dasigi, Kyle Lo, Iz~Beltagy, Arman Cohan, Noah~A Smith, and Matt
  Gardner.
\newblock A dataset of information-seeking questions and answers anchored in
  research papers.
\newblock \emph{arXiv preprint arXiv:2105.03011}, 2021.

\bibitem[Fabbri et~al.(2019)Fabbri, Li, She, Li, and Radev]{fabbri2019multi}
Alexander~R Fabbri, Irene Li, Tianwei She, Suyi Li, and Dragomir~R Radev.
\newblock Multi-news: A large-scale multi-document summarization dataset and
  abstractive hierarchical model.
\newblock \emph{arXiv preprint arXiv:1906.01749}, 2019.

\bibitem[Face(2024{\natexlab{a}})]{huggingface_kv_cache}
Hugging Face.
\newblock Unlocking longer generation with key-value cache quantization.
\newblock Hugging Face Blog, 2024{\natexlab{a}}.
\newblock URL \url{https://huggingface.co/blog/kv-cache-quantization}.

\bibitem[Face(2024{\natexlab{b}})]{huggingface_quanto}
Hugging Face.
\newblock Quanto: a pytorch quantization backend for optimum.
\newblock Hugging Face Blog, 2024{\natexlab{b}}.
\newblock URL \url{https://huggingface.co/blog/quanto-introduction}.

\bibitem[Feng et~al.(2024)Feng, Lv, Cao, Xie, and Zhou]{ada}
Yuan Feng, Junlin Lv, Yukun Cao, Xike Xie, and S.~Kevin Zhou.
\newblock Ada-kv: Optimizing kv cache eviction by adaptive budget allocation
  for efficient llm inference, 2024.
\newblock URL \url{https://arxiv.org/abs/2407.11550}.

\bibitem[Feng et~al.(2025)Feng, Lv, Cao, Xie, and
  Zhou]{feng2025identifycriticalkvcache}
Yuan Feng, Junlin Lv, Yukun Cao, Xike Xie, and S~Kevin Zhou.
\newblock Identify critical kv cache in llm inference from an output
  perturbation perspective, 2025.
\newblock URL \url{https://arxiv.org/abs/2502.03805}.

\bibitem[Fu et~al.(2024)Fu, Cai, Asi, Xiong, Dong, and Xiao]{fu2024not}
Yu~Fu, Zefan Cai, Abedelkadir Asi, Wayne Xiong, Yue Dong, and Wen Xiao.
\newblock Not all heads matter: A head-level kv cache compression method with
  integrated retrieval and reasoning.
\newblock \emph{arXiv preprint arXiv:2410.19258}, 2024.

\bibitem[Gliwa et~al.(2019)Gliwa, Mochol, Biesek, and Wawer]{gliwa2019samsum}
Bogdan Gliwa, Iwona Mochol, Maciej Biesek, and Aleksander Wawer.
\newblock Samsum corpus: A human-annotated dialogue dataset for abstractive
  summarization.
\newblock \emph{arXiv preprint arXiv:1911.12237}, 2019.

\bibitem[Gu(2023)]{gu2023llm}
Qiuhan Gu.
\newblock Llm-based code generation method for golang compiler testing.
\newblock In \emph{Proceedings of the 31st ACM Joint European Software
  Engineering Conference and Symposium on the Foundations of Software
  Engineering}, pp.\  2201--2203, 2023.

\bibitem[Guo et~al.(2023)Guo, Xu, Duan, Yin, and
  McAuley]{guo2023longcoderlongrangepretrainedlanguage}
Daya Guo, Canwen Xu, Nan Duan, Jian Yin, and Julian McAuley.
\newblock Longcoder: A long-range pre-trained language model for code
  completion, 2023.
\newblock URL \url{https://arxiv.org/abs/2306.14893}.

\bibitem[Ho et~al.(2020)Ho, Duong~Nguyen, Sugawara, and
  Aizawa]{ho-etal-2020-constructing}
Xanh Ho, Anh-Khoa Duong~Nguyen, Saku Sugawara, and Akiko Aizawa.
\newblock Constructing a multi-hop {QA} dataset for comprehensive evaluation of
  reasoning steps.
\newblock In Donia Scott, Nuria Bel, and Chengqing Zong (eds.),
  \emph{Proceedings of the 28th International Conference on Computational
  Linguistics}, pp.\  6609--6625, Barcelona, Spain (Online), December 2020.
  International Committee on Computational Linguistics.
\newblock \doi{10.18653/v1/2020.coling-main.580}.
\newblock URL \url{https://aclanthology.org/2020.coling-main.580}.

\bibitem[Hooper et~al.(2024)Hooper, Kim, Mohammadzadeh, Mahoney, Shao, Keutzer,
  and Gholami]{hooper2024kvquant}
Coleman Hooper, Sehoon Kim, Hiva Mohammadzadeh, Michael~W Mahoney, Sophia Shao,
  Kurt Keutzer, and Amir Gholami.
\newblock Kvquant: Towards 10 million context length llm inference with kv
  cache quantization.
\newblock \emph{Advances in Neural Information Processing Systems},
  37:\penalty0 1270--1303, 2024.

\bibitem[Hsieh et~al.(2024)Hsieh, Sun, Kriman, Acharya, Rekesh, Jia, Zhang, and
  Ginsburg]{hsieh2024ruler}
Cheng-Ping Hsieh, Simeng Sun, Samuel Kriman, Shantanu Acharya, Dima Rekesh, Fei
  Jia, Yang Zhang, and Boris Ginsburg.
\newblock Ruler: What's the real context size of your long-context language
  models?
\newblock \emph{arXiv preprint arXiv:2404.06654}, 2024.

\bibitem[Huang et~al.(2021)Huang, Cao, Parulian, Ji, and
  Wang]{huang2021efficient}
Luyang Huang, Shuyang Cao, Nikolaus Parulian, Heng Ji, and Lu~Wang.
\newblock Efficient attentions for long document summarization.
\newblock \emph{arXiv preprint arXiv:2104.02112}, 2021.

\bibitem[Jiang et~al.(2023)Jiang, Sablayrolles, Mensch, Bamford, Chaplot,
  Casas, Bressand, Lengyel, Lample, Saulnier, et~al.]{jiang2023mistral}
Albert~Q Jiang, Alexandre Sablayrolles, Arthur Mensch, Chris Bamford,
  Devendra~Singh Chaplot, Diego de~las Casas, Florian Bressand, Gianna Lengyel,
  Guillaume Lample, Lucile Saulnier, et~al.
\newblock Mistral 7b.
\newblock \emph{arXiv preprint arXiv:2310.06825}, 2023.

\bibitem[Jiang et~al.(2024)Jiang, Li, Zhang, Wu, Luo, Ahn, Han, Abdi, Li, Lin,
  et~al.]{jiang2024minference}
Huiqiang Jiang, Yucheng Li, Chengruidong Zhang, Qianhui Wu, Xufang Luo, Surin
  Ahn, Zhenhua Han, Amir~H Abdi, Dongsheng Li, Chin-Yew Lin, et~al.
\newblock Minference 1.0: Accelerating pre-filling for long-context llms via
  dynamic sparse attention.
\newblock \emph{arXiv preprint arXiv:2407.02490}, 2024.

\bibitem[Joshi et~al.(2017)Joshi, Choi, Weld, and
  Zettlemoyer]{joshi2017triviaqalargescaledistantly}
Mandar Joshi, Eunsol Choi, Daniel~S. Weld, and Luke Zettlemoyer.
\newblock Triviaqa: A large scale distantly supervised challenge dataset for
  reading comprehension, 2017.
\newblock URL \url{https://arxiv.org/abs/1705.03551}.

\bibitem[Kamradt(2023)]{needle}
Gregory Kamradt.
\newblock {Needle In A Haystack} - pressure testing {LLM}s.
\newblock \emph{Github}, 2023.
\newblock URL
  \url{https://github.com/gkamradt/LLMTest_NeedleInAHaystack/tree/main}.

\bibitem[Ko{\v{c}}isk{\`y} et~al.(2018)Ko{\v{c}}isk{\`y}, Schwarz, Blunsom,
  Dyer, Hermann, Melis, and Grefenstette]{kovcisky2018narrativeqa}
Tom{\'a}{\v{s}} Ko{\v{c}}isk{\`y}, Jonathan Schwarz, Phil Blunsom, Chris Dyer,
  Karl~Moritz Hermann, G{\'a}bor Melis, and Edward Grefenstette.
\newblock The narrativeqa reading comprehension challenge.
\newblock \emph{Transactions of the Association for Computational Linguistics},
  6:\penalty0 317--328, 2018.

\bibitem[Li \& Roth(2002)Li and Roth]{li2002learning}
Xin Li and Dan Roth.
\newblock Learning question classifiers.
\newblock In \emph{COLING 2002: The 19th International Conference on
  Computational Linguistics}, 2002.

\bibitem[Li et~al.(2024)Li, Huang, Yang, Venkitesh, Locatelli, Ye, Cai, Lewis,
  and Chen]{SnapKV}
Yuhong Li, Yingbing Huang, Bowen Yang, Bharat Venkitesh, Acyr Locatelli,
  Hanchen Ye, Tianle Cai, Patrick Lewis, and Deming Chen.
\newblock Snapkv: Llm knows what you are looking for before generation.
\newblock \emph{arXiv preprint arXiv:2404.14469}, 2024.

\bibitem[Liu et~al.(2024{\natexlab{a}})Liu, Liu, Pan, He, Haffari, and
  Zhuang]{liu2024minicache}
Akide Liu, Jing Liu, Zizheng Pan, Yefei He, Reza Haffari, and Bohan Zhuang.
\newblock Minicache: Kv cache compression in depth dimension for large language
  models.
\newblock \emph{Advances in Neural Information Processing Systems},
  37:\penalty0 139997--140031, 2024{\natexlab{a}}.

\bibitem[Liu et~al.(2024{\natexlab{b}})Liu, Chen, Lu, Jiang, Han, Zhang, Chen,
  Zhang, Ding, Zhang, et~al.]{liu2024retrievalattention}
Di~Liu, Meng Chen, Baotong Lu, Huiqiang Jiang, Zhenhua Han, Qianxi Zhang,
  Qi~Chen, Chengruidong Zhang, Bailu Ding, Kai Zhang, et~al.
\newblock Retrievalattention: Accelerating long-context llm inference via
  vector retrieval.
\newblock \emph{arXiv preprint arXiv:2409.10516}, 2024{\natexlab{b}}.

\bibitem[Liu et~al.(2023)Liu, Xu, and
  McAuley]{liu2023repobenchbenchmarkingrepositorylevelcode}
Tianyang Liu, Canwen Xu, and Julian McAuley.
\newblock Repobench: Benchmarking repository-level code auto-completion
  systems, 2023.
\newblock URL \url{https://arxiv.org/abs/2306.03091}.

\bibitem[Liu et~al.(2024{\natexlab{c}})Liu, Desai, Liao, Wang, Xie, Xu,
  Kyrillidis, and Shrivastava]{liu2024scissorhands}
Zichang Liu, Aditya Desai, Fangshuo Liao, Weitao Wang, Victor Xie, Zhaozhuo Xu,
  Anastasios Kyrillidis, and Anshumali Shrivastava.
\newblock Scissorhands: Exploiting the persistence of importance hypothesis for
  llm kv cache compression at test time.
\newblock \emph{Advances in Neural Information Processing Systems}, 36,
  2024{\natexlab{c}}.

\bibitem[Liu et~al.(2024{\natexlab{d}})Liu, Yuan, Jin, Zhong, Xu, Braverman,
  Chen, and Hu]{liu2024kivi}
Zirui Liu, Jiayi Yuan, Hongye Jin, Shaochen Zhong, Zhaozhuo Xu, Vladimir
  Braverman, Beidi Chen, and Xia Hu.
\newblock Kivi: A tuning-free asymmetric 2bit quantization for kv cache.
\newblock \emph{arXiv preprint arXiv:2402.02750}, 2024{\natexlab{d}}.

\bibitem[Nawrot et~al.(2025)Nawrot, Li, Huang, Ruder, Marchisio, and
  Ponti]{nawrot2025sparsefrontiersparseattention}
Piotr Nawrot, Robert Li, Renjie Huang, Sebastian Ruder, Kelly Marchisio, and
  Edoardo~M. Ponti.
\newblock The sparse frontier: Sparse attention trade-offs in transformer llms,
  2025.
\newblock URL \url{https://arxiv.org/abs/2504.17768}.

\bibitem[NVIDIA(2024)]{kvpress}
NVIDIA.
\newblock Kvpress, 2024.
\newblock URL \url{https://github.com/NVIDIA/kvpress}.

\bibitem[Oren et~al.(2024)Oren, Hassid, Yarden, Adi, and
  Schwartz]{oren2024transformers}
Matanel Oren, Michael Hassid, Nir Yarden, Yossi Adi, and Roy Schwartz.
\newblock Transformers are multi-state rnns.
\newblock \emph{arXiv preprint arXiv:2401.06104}, 2024.

\bibitem[Qin et~al.(2025)Qin, Cao, Lin, Hu, Fan, Cheng, Lin, and
  Li]{qin2025cake}
Ziran Qin, Yuchen Cao, Mingbao Lin, Wen Hu, Shixuan Fan, Ke~Cheng, Weiyao Lin,
  and Jianguo Li.
\newblock {CAKE}: Cascading and adaptive {KV} cache eviction with layer
  preferences.
\newblock In \emph{The Thirteenth International Conference on Learning
  Representations}, 2025.
\newblock URL \url{https://openreview.net/forum?id=EQgEMAD4kv}.

\bibitem[Ren \& Zhu(2024)Ren and Zhu]{ren2024efficacyevictionpolicykeyvalue}
Siyu Ren and Kenny~Q. Zhu.
\newblock On the efficacy of eviction policy for key-value constrained
  generative language model inference, 2024.
\newblock URL \url{https://arxiv.org/abs/2402.06262}.

\bibitem[Sun et~al.(2024)Sun, Chen, Yang, Tian, and Chen]{sun2024triforce}
Hanshi Sun, Zhuoming Chen, Xinyu Yang, Yuandong Tian, and Beidi Chen.
\newblock Triforce: Lossless acceleration of long sequence generation with
  hierarchical speculative decoding.
\newblock \emph{arXiv preprint arXiv:2404.11912}, 2024.

\bibitem[Tang et~al.(2024)Tang, Zhao, Zhu, Xiao, Kasikci, and
  Han]{tang2024quest}
Jiaming Tang, Yilong Zhao, Kan Zhu, Guangxuan Xiao, Baris Kasikci, and Song
  Han.
\newblock Quest: Query-aware sparsity for efficient long-context llm inference.
\newblock \emph{arXiv preprint arXiv:2406.10774}, 2024.

\bibitem[Team(2024)]{qwen2.5}
Qwen Team.
\newblock Qwen2.5: A party of foundation models, September 2024.
\newblock URL \url{https://qwenlm.github.io/blog/qwen2.5/}.

\bibitem[Touvron et~al.(2023)Touvron, Martin, Stone, Albert, Almahairi, Babaei,
  Bashlykov, Batra, Bhargava, Bhosale, et~al.]{llama2}
Hugo Touvron, Louis Martin, Kevin Stone, Peter Albert, Amjad Almahairi, Yasmine
  Babaei, Nikolay Bashlykov, Soumya Batra, Prajjwal Bhargava, Shruti Bhosale,
  et~al.
\newblock Llama 2: Open foundation and fine-tuned chat models.
\newblock \emph{arXiv preprint arXiv:2307.09288}, 2023.

\bibitem[Trivedi et~al.(2022)Trivedi, Balasubramanian, Khot, and
  Sabharwal]{trivedi2022musique}
Harsh Trivedi, Niranjan Balasubramanian, Tushar Khot, and Ashish Sabharwal.
\newblock Musique: Multihop questions via single-hop question composition.
\newblock \emph{Transactions of the Association for Computational Linguistics},
  10:\penalty0 539--554, 2022.

\bibitem[Xiao et~al.(2024{\natexlab{a}})Xiao, Zhang, Han, Xiao, Lin, Zhang,
  Liu, and Sun]{xiao2024infllm}
Chaojun Xiao, Pengle Zhang, Xu~Han, Guangxuan Xiao, Yankai Lin, Zhengyan Zhang,
  Zhiyuan Liu, and Maosong Sun.
\newblock Infllm: Training-free long-context extrapolation for llms with an
  efficient context memory.
\newblock \emph{arXiv preprint arXiv:2402.04617}, 2024{\natexlab{a}}.

\bibitem[Xiao et~al.(2023)Xiao, Tian, Chen, Han, and Lewis]{streamingllm}
Guangxuan Xiao, Yuandong Tian, Beidi Chen, Song Han, and Mike Lewis.
\newblock Efficient streaming language models with attention sinks.
\newblock \emph{arXiv preprint arXiv:2309.17453}, 2023.

\bibitem[Xiao et~al.(2024{\natexlab{b}})Xiao, Tang, Zuo, Guo, Yang, Tang, Fu,
  and Han]{duo}
Guangxuan Xiao, Jiaming Tang, Jingwei Zuo, Junxian Guo, Shang Yang, Haotian
  Tang, Yao Fu, and Song Han.
\newblock Duoattention: Efficient long-context llm inference with retrieval and
  streaming heads, 2024{\natexlab{b}}.
\newblock URL \url{https://arxiv.org/abs/2410.10819}.

\bibitem[Xiao et~al.(2025)Xiao, Tang, Zuo, junxian guo, Yang, Tang, Fu, and
  Han]{xiao2025duoattention}
Guangxuan Xiao, Jiaming Tang, Jingwei Zuo, junxian guo, Shang Yang, Haotian
  Tang, Yao Fu, and Song Han.
\newblock Duoattention: Efficient long-context {LLM} inference with retrieval
  and streaming heads.
\newblock In \emph{The Thirteenth International Conference on Learning
  Representations}, 2025.
\newblock URL \url{https://openreview.net/forum?id=cFu7ze7xUm}.

\bibitem[Xu et~al.(2024)Xu, Jie, Dong, Wang, Lu, Zhou, Saha, Xiong, and
  Sahoo]{xu2024think}
Yuhui Xu, Zhanming Jie, Hanze Dong, Lei Wang, Xudong Lu, Aojun Zhou, Amrita
  Saha, Caiming Xiong, and Doyen Sahoo.
\newblock Think: Thinner key cache by query-driven pruning.
\newblock \emph{arXiv preprint arXiv:2407.21018}, 2024.

\bibitem[Yang et~al.(2025)Yang, Du, Zhang, Wang, Pang, Du, and
  An]{yang2025longspeclongcontextspeculativedecoding}
Penghui Yang, Cunxiao Du, Fengzhuo Zhang, Haonan Wang, Tianyu Pang, Chao Du,
  and Bo~An.
\newblock Longspec: Long-context speculative decoding with efficient drafting
  and verification, 2025.
\newblock URL \url{https://arxiv.org/abs/2502.17421}.

\bibitem[Yang et~al.(2024)Yang, Cao, Chen, Qin, Yang, Zhao, and
  Chen]{yang2024kvsharer}
Yifei Yang, Zouying Cao, Qiguang Chen, Libo Qin, Dongjie Yang, Hai Zhao, and
  Zhi Chen.
\newblock Kvsharer: Efficient inference via layer-wise dissimilar kv cache
  sharing.
\newblock \emph{arXiv preprint arXiv:2410.18517}, 2024.

\bibitem[Yang et~al.(2018)Yang, Qi, Zhang, Bengio, Cohen, Salakhutdinov, and
  Manning]{multi_hop1}
Zhilin Yang, Peng Qi, Saizheng Zhang, Yoshua Bengio, William~W Cohen, Ruslan
  Salakhutdinov, and Christopher~D Manning.
\newblock Hotpotqa: A dataset for diverse, explainable multi-hop question
  answering.
\newblock \emph{arXiv preprint arXiv:1809.09600}, 2018.

\bibitem[Yi et~al.(2024)Yi, Ouyang, Liu, Liao, Xu, and Shen]{yi2024survey}
Zihao Yi, Jiarui Ouyang, Yuwen Liu, Tianhao Liao, Zhe Xu, and Ying Shen.
\newblock A survey on recent advances in llm-based multi-turn dialogue systems.
\newblock \emph{arXiv preprint arXiv:2402.18013}, 2024.

\bibitem[Zhang et~al.(2025)Zhang, Zhang, Du, Du, Pang, Gao, and
  Lin]{zhang2025lighttransfer}
Xuan Zhang, Fengzhuo Zhang, Cunxiao Du, Chao Du, Tianyu Pang, Wei Gao, and Min
  Lin.
\newblock Lighttransfer: Your long-context llm is secretly a hybrid model with
  effortless adaptation.
\newblock In \emph{Workshop on Reasoning and Planning for Large Language
  Models}, 2025.

\bibitem[Zhang et~al.(2024{\natexlab{a}})Zhang, Gao, Liu, Lu, Xiong, Dong,
  Chang, Hu, Xiao, et~al.]{pyramidkv}
Yichi Zhang, Bofei Gao, Tianyu Liu, Keming Lu, Wayne Xiong, Yue Dong, Baobao
  Chang, Junjie Hu, Wen Xiao, et~al.
\newblock Pyramidkv: Dynamic kv cache compression based on pyramidal
  information funneling.
\newblock \emph{arXiv preprint arXiv:2406.02069}, 2024{\natexlab{a}}.

\bibitem[Zhang et~al.(2024{\natexlab{b}})Zhang, Sheng, Zhou, Chen, Zheng, Cai,
  Song, Tian, R{\'e}, Barrett, et~al.]{h2o}
Zhenyu Zhang, Ying Sheng, Tianyi Zhou, Tianlong Chen, Lianmin Zheng, Ruisi Cai,
  Zhao Song, Yuandong Tian, Christopher R{\'e}, Clark Barrett, et~al.
\newblock H2o: Heavy-hitter oracle for efficient generative inference of large
  language models.
\newblock \emph{Advances in Neural Information Processing Systems}, 36,
  2024{\natexlab{b}}.

\bibitem[Zhong et~al.(2021)Zhong, Yin, Yu, Zaidi, Mutuma, Jha, Awadallah,
  Celikyilmaz, Liu, Qiu, et~al.]{zhong2021qmsum}
Ming Zhong, Da~Yin, Tao Yu, Ahmad Zaidi, Mutethia Mutuma, Rahul Jha,
  Ahmed~Hassan Awadallah, Asli Celikyilmaz, Yang Liu, Xipeng Qiu, et~al.
\newblock Qmsum: A new benchmark for query-based multi-domain meeting
  summarization.
\newblock \emph{arXiv preprint arXiv:2104.05938}, 2021.

\end{thebibliography}

\newpage
\appendix
\begin{figure}[t]
	\centering  
	\vspace{-0.2cm}  
	\begin{subfigure}{0.325\textwidth}  
		\centering  
		\includegraphics[width=\linewidth]{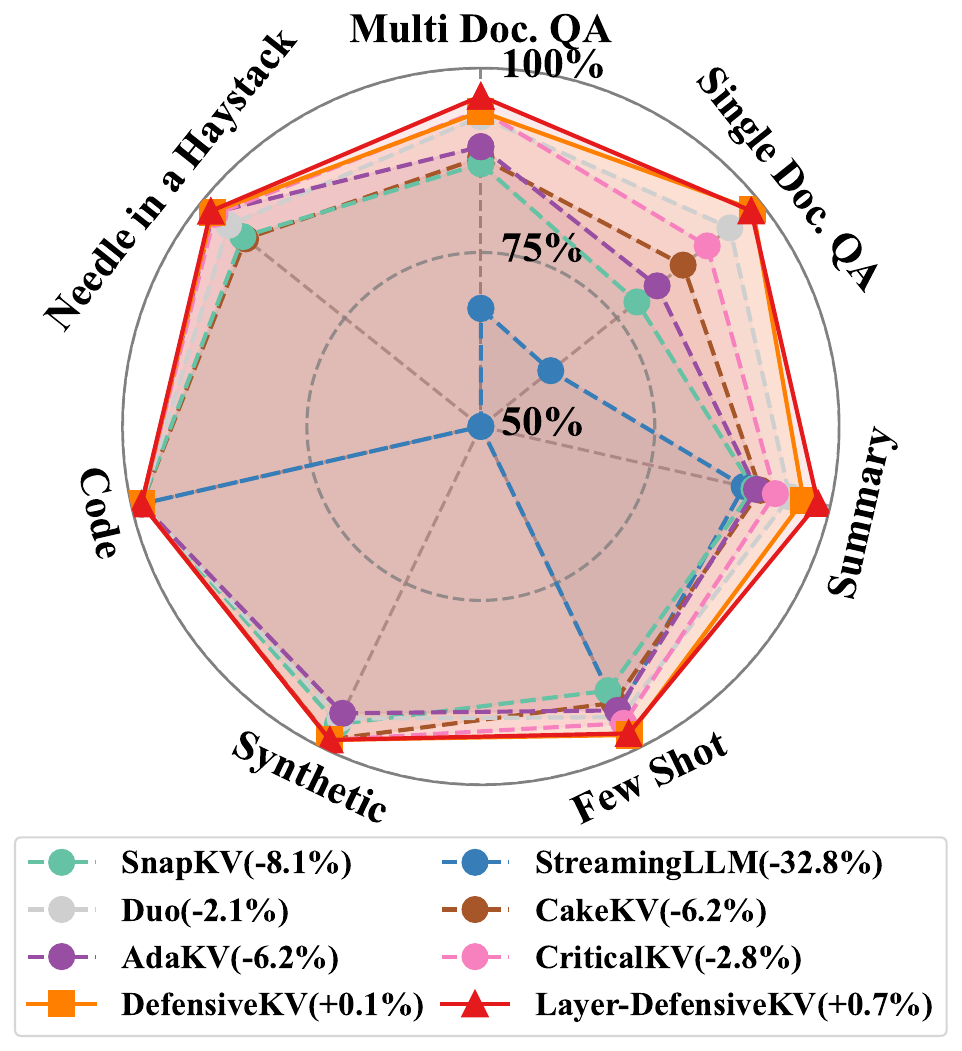}  
		\vspace{-0.5cm}  
		\caption{Llama-3.1-8B, 40\% Cache}  
	\end{subfigure}  
	\begin{subfigure}{0.325\textwidth}  
		\centering  
		\includegraphics[width=\linewidth]{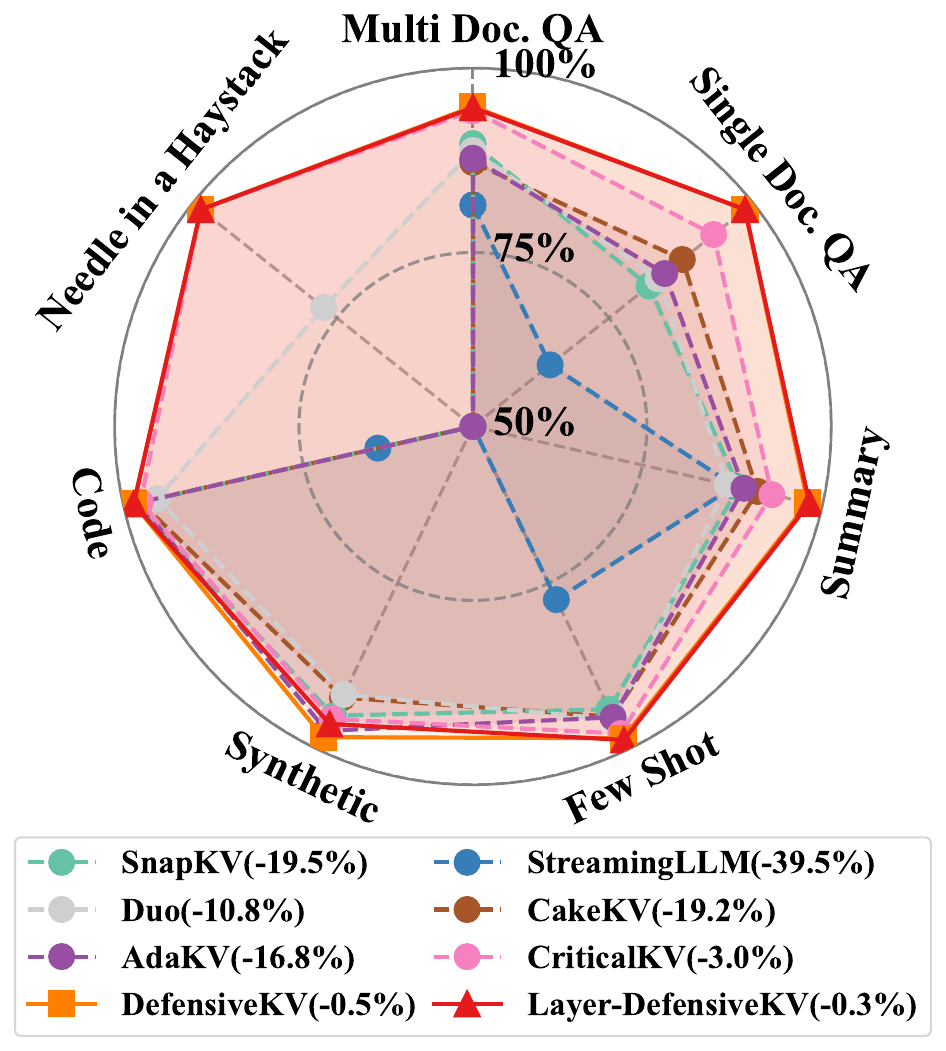}  
		\vspace{-0.5cm}  
		\caption{Mistral-7B, 40\% Cache}  
	\end{subfigure}  
	\begin{subfigure}{0.325\textwidth}  
		\centering  
		\includegraphics[width=\linewidth]{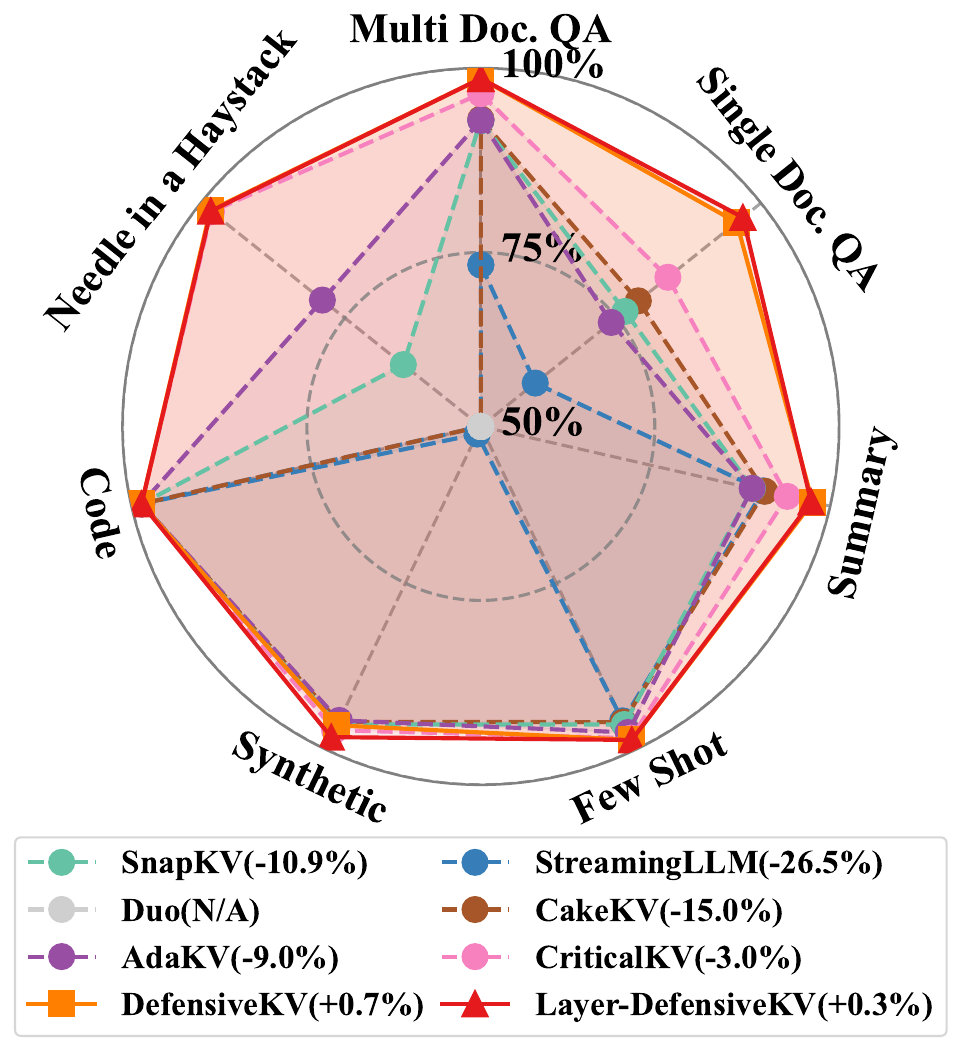}  
		\vspace{-0.5cm}  
		\caption{Qwen-32B, 40\% Cache}  
	\end{subfigure}  
	
	\begin{subfigure}{0.325\textwidth}  
		\centering  
		\includegraphics[width=\linewidth]{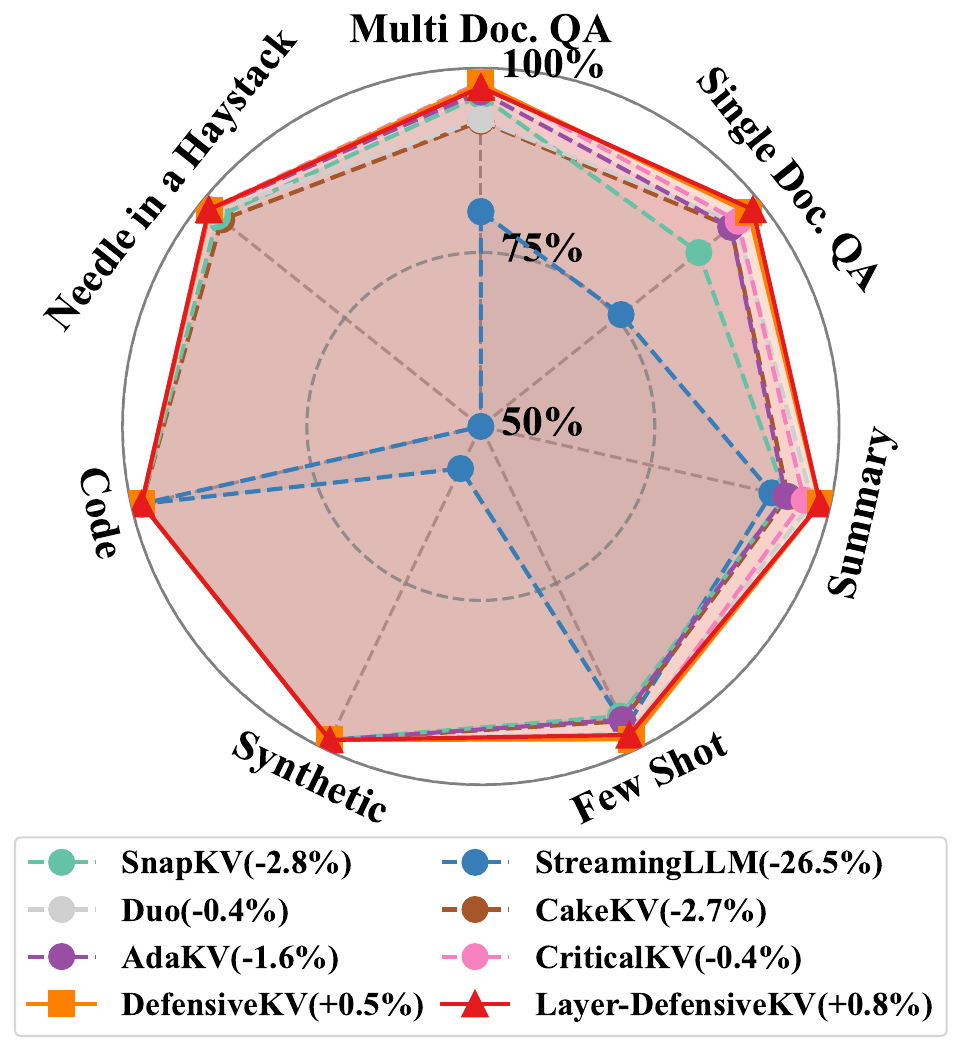}  
		\vspace{-0.5cm}  
		\caption{Llama-3.1-8B, 60\% Cache}  
	\end{subfigure}  
	\begin{subfigure}{0.325\textwidth}  
		\centering  
		\includegraphics[width=\linewidth]{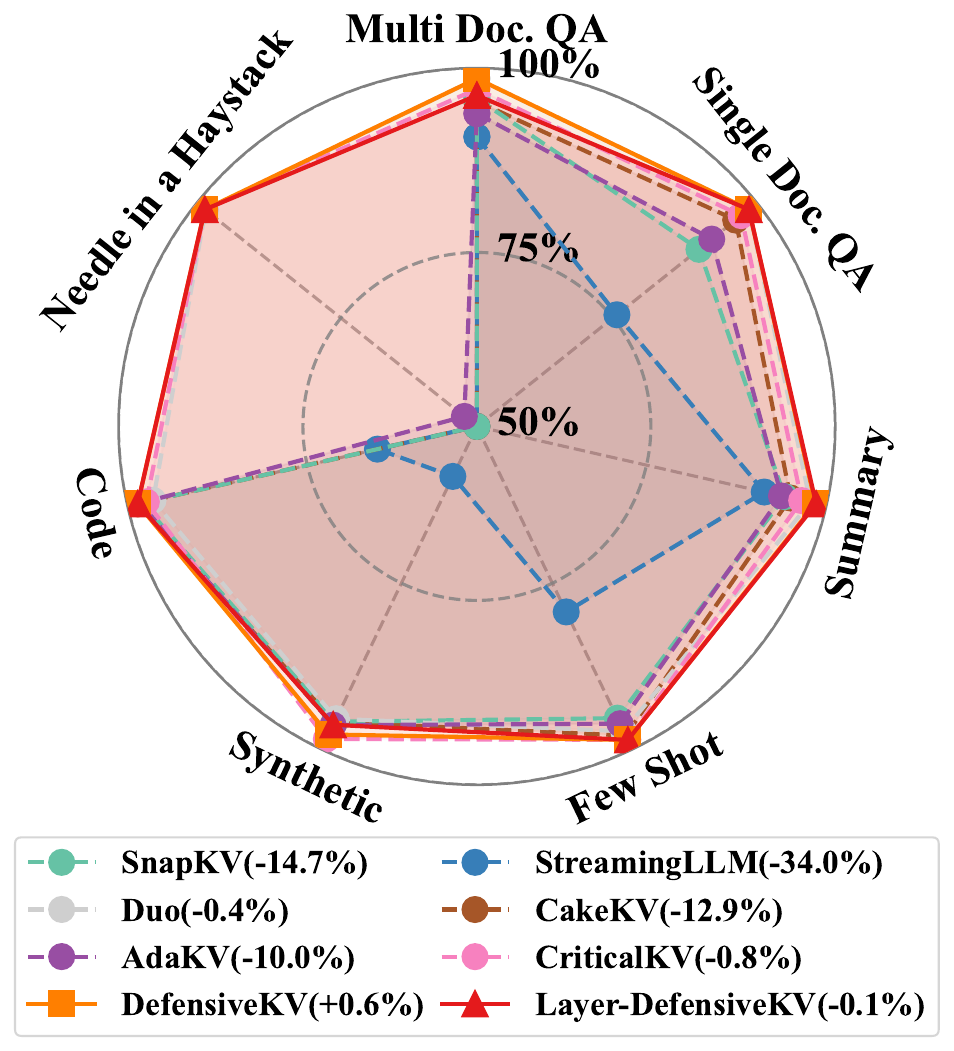}  
		\vspace{-0.5cm}  
		\caption{Mistral-7B, 60\% Cache}  
	\end{subfigure}  
	\begin{subfigure}{0.325\textwidth}  
		\centering  
		\includegraphics[width=\linewidth]{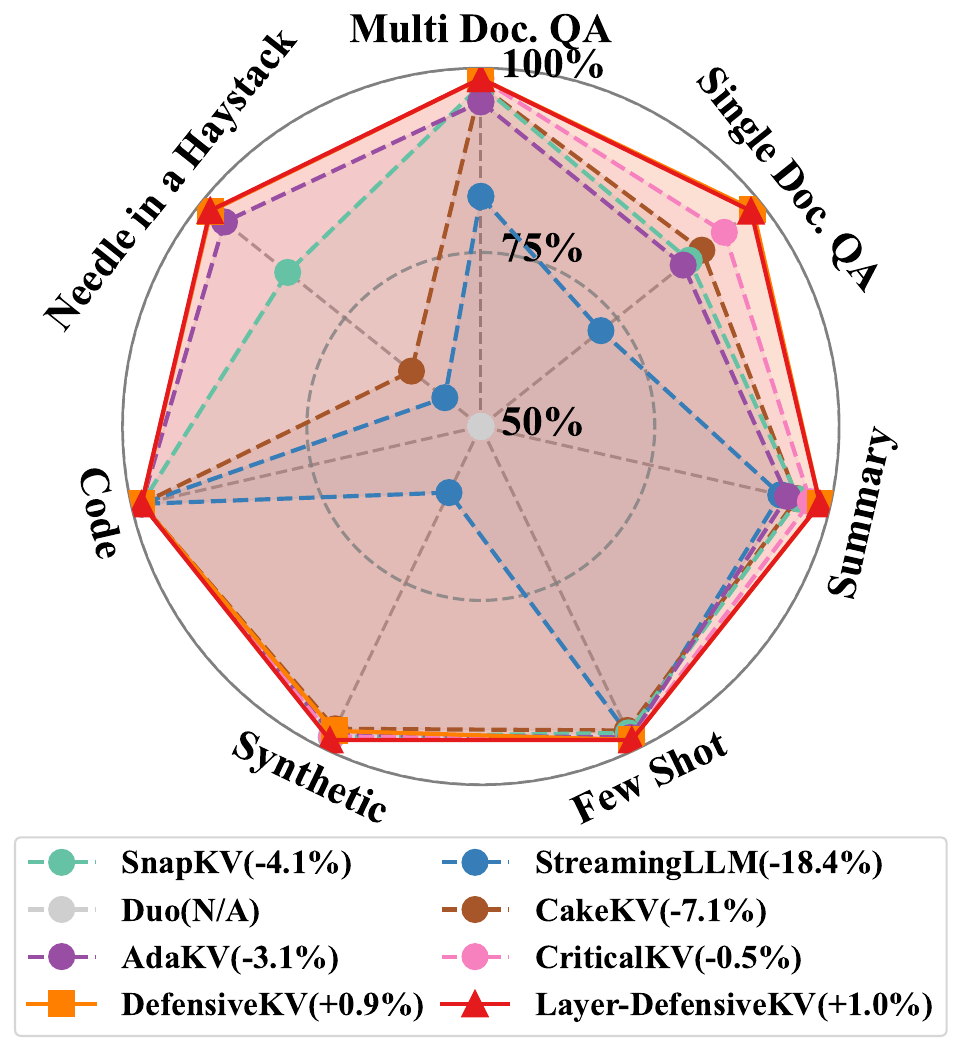}  
		\vspace{-0.5cm}  
		\caption{Qwen-32B, 60\% Cache}  
	\end{subfigure}

	\vspace{-0.3cm}
	\caption{Summarization of quality losses.}
	\vspace{-0.3cm}
	\label{fig:more_main_performance}
\end{figure}

\section{Quality Losses of Methods with 40\% and 60\% Cache Size}
\label{apdx:more_quality}
Figure~\ref{fig:more_main_performance} further summarizes the quality losses of different methods at 40\% and 60\% cache sizes. It can be observed that both DefensiveKV and Layer-DefensiveKV maintain nearly lossless performance, in some cases even surpassing the original uncompressed results. In contrast, all other methods exhibit notable declines in quality. These results demonstrate the effectiveness of our approach.

\section{Detailed  Settings}
\label{apdx:settings}
The fundamental settings for SnapKV, CAKE, AdaKV, CriticalKV and our methods were kept as originally defined, with an average-pooling kernel size of 5 and a historical token size of 32 for observation. For StreamingLLM~\citep{streamingllm}, we follow standard settings, using 4 sink tokens and retaining the most recent window’s cache. For DuoAttention~\citep{xiao2025duoattention}, we follow the publicly released training settings. 
Following standard practices in prior studies~\citep{SnapKV, pyramidkv, ada}, we perform cache eviction immediately after the prefilling phase of each layer. 

\section{Additional Related Works}
\label{apdx:related}
Beyond cache eviction methods, a broader range of related work can reduce KV cache footprint. For example, Think~\citep{xu2024think} compresses the KV cache by reducing the number of channels in the key states. Other approaches, such as MiniCache~\citep{liu2024minicache} and KVSharer~\citep{yang2024kvsharer}, exploit KV similarity between layers to achieve compression. These techniques are orthogonal to KV cache eviction methods and can be further combined with them. KV cache quantization~\citep{hooper2024kvquant, liu2024kivi}, which reduces the precision of individual cache entries (e.g., quantizing 16-bit entries to 4-bit or 2-bit), also offers footprint reduction. Because quantization methods typically retain all cache entries, they are fundamentally orthogonal to the cache eviction methods explored in this paper and can also be applied to further enhance them. Furthermore, recent speculative decoding methods explore using a reduced KV cache for draft generation in long-sequence generation~\citep{sun2024triforce,yang2025longspeclongcontextspeculativedecoding}. Refining cache eviction to enhance speculative decoding is also a promising research direction.

Sparse attention methods are conceptually related to KV cache eviction~\citep{xiao2024infllm,tang2024quest,jiang2024minference,liu2024retrievalattention}. The key difference is that KV cache eviction retains only a subset of the KV cache, while sparse attention methods keep all entries but selectively utilize only a critical subset during computation~\citep{nawrot2025sparsefrontiersparseattention}. Consequently, sparse attention methods do not reduce the memory footprint of the KV cache. The two technique lines are, in fact, orthogonal. Future research could explore firstly employing KV cache eviction to compress the cache to a certain proportion (e.g., 40\% cache size with minimal loss, as demonstrated in this paper) and then applying sparse attention for further acceleration. This represents a promising direction for future research.

\begin{figure}
	\includegraphics[width=0.5\linewidth]{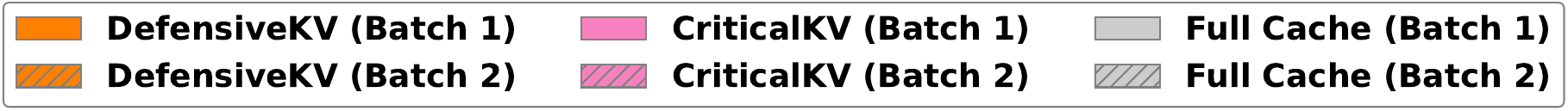}
	\centering
	\includegraphics[width=0.5\linewidth]{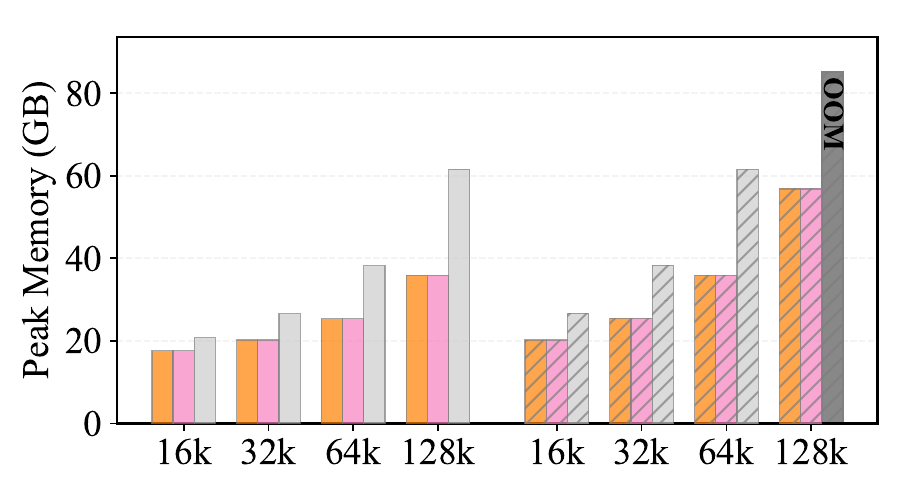}
	\caption{Peak Memory usage(All with FlashAttention-2).}
	\label{fig:memory}
	\vspace{-0.2cm}
	
\end{figure}

\section{Memory Usage during Generation}
\label{apdx:memory}
Following the efficiency evaluation in Section~\ref{sc:effi}, we also measured peak memory usage during inference. The memory savings from cache eviction are primarily determined by the compressed cache size. Our introduced defensive aggregation method does not differ in memory usage from standard mean aggregation. As shown in Figure~\ref{fig:memory}, DefensiveKV and CriticalKV exhibit significantly lower peak memory usage than Full Cache. For example, with a batch size of 1 and a 128K context length, DefensiveKV and CriticalKV use only 36GB, far less than Full Cache's 61.5 GB. This allows them to support larger batch sizes, such as batch size 2, further increasing decoding throughput, while Full Cache encounters out-of-memory (OOM) errors. This advantage enables DefensiveKV and CriticalKV to achieve up to a 4.2x speedup in 128K decoding throughput compared to Full Cache.

\section{Integration DefensiveKV with KV Cache Quantization}
\label{apdx:quan}
We combine DefensiveKV with another orthogonal technique, cache quantization. Specifically, we adopted the official HuggingFace-provided int4 quantization for the KV cache~\citep{huggingface_kv_cache}, with backend support from Quanto~\citep{huggingface_quanto}. As showed in Table~\ref{tab:quant}, we first compressed the cache entries to 40\% of their original num, and then further quantized them from bf16 to int4 (ultimately reducing the cache memory footprint to just 10\%). After full integration, the average score only dropped slightly from 49.21 to 48.55, demonstrating that reducing the cache memory to 10\% comes with less than one-point loss in performance. This highlights the great potential of combining these orthogonal techniques for practical applications.

\section{More Needle-in-A-Haystack-style Evaluations on Ruler Benchmark}
\label{apdx:more_ruler}

In the Needle-in-A-Haystack task, a keyword, referred to as the "needle", is embedded within a lengthy context known as the "haystack". The objective of this task is to extract the "needle" from the "haystack", which is composed of essays by Paul Graham \citep{needle}. In our main experiments, we adopt the respective prompt templates (see Table~\ref{tab:ruler_task_template1}) used in the Ruler Benchmark~\citep{hsieh2024ruler} (corresponding to NIAH-s2 and NIAH-MV in their formulation) to ensure consistency and reproducibility

The whole Ruler benchmark~\citep{hsieh2024ruler} comprises 13 synthetic, needle-in-a-haystack-style tasks designed to evaluate the long-context capabilities of models. A single evaluation on the full 32K RULER benchmark requires approximately 9 GPU hours. Consequently, a comprehensive assessment across all methods, compression rates, and models would demand an estimated 864 GPU hours, which is computationally prohibitive.

In this section, we further presents a more extensive analysis on the complete Ruler benchmark. We evaluated our proposed methods and the strongest baseline, CriticalKV, at 20\% and 40\% cache sizes using Llama-3.1-8B-Instruct, with the results detailed in Table~\ref{rulerdetail}. Both DefensiveKV and Layer-DefensiveKV demonstrated significant advantages; for instance, at a 20\% cache size, they achieved average scores of 84.91 and 86.39, respectively, substantially outperforming the CriticalKV baseline's score of 66.57. These findings underscore our method's ability to achieve strong compression performance with minimal loss in accuracy.
	
	\begin{table*}[!t]
		\vspace{0.2cm}
		\centering
		\caption{Detailed scores of 13 datasets on Ruler.}
		\vspace{-0.2cm}
		\label{rulerdetail}
		\resizebox{\textwidth}{!}{%
			\setlength{\tabcolsep}{1.5pt}
				\begin{tabular}{@{}llllllllllllllllllllll@{}}
					\toprule
					
					\makecell[l]{\scalebox{1.3}{Method}}
					
					& \rotatebox{40}{\textbf{cwe}} & \rotatebox{40}{\textbf{fwe}} & \rotatebox{40}{\textbf{niah\_mk1}} & \rotatebox{40}{\textbf{niah\_mk2}} & \rotatebox{40}{\textbf{niah\_mk3}} & \rotatebox{40}{\textbf{niah\_mq}} & \rotatebox{40}{\textbf{niah\_mv}} & \rotatebox{40}{\textbf{niah\_s1}} & \rotatebox{40}{\textbf{niah\_s2}} & \rotatebox{40}{\textbf{niah\_s3}} & \rotatebox{40}{\textbf{qa\_1}} & \rotatebox{40}{\textbf{qa\_2}} & \rotatebox{40}{\textbf{vt}} & \rotatebox{0}{\textbf{Avg. }} \\
					
					\midrule
					\multicolumn{15}{c}{Llama-3.1-8B-Instruct, 32K Ruler, $20\%$ Cache Size}\\
					\midrule
					FullCache & 45.22 & 94.13 & 99.60 & 99.60 & 99.40 & 98.75 & 99.10 & 100.00 & 100.00 & 100.00 & 79.80 & 54.80 & 99.24 & 89.97 \\
					\arrayrulecolor{lightgray}
					\midrule
					\arrayrulecolor{black}
					Strongest Baseline CriticalKV & \textbf{26.80} & 88.80 & 91.60 & 29.40 & 19.40 & 95.00 & 93.60 & \textbf{100.00} & 99.60 & 42.40 & 40.80 & 40.20 & 97.76 & 66.57 \\
					DefensiveKV & 22.94 & 90.00 & \textbf{99.80} & 86.80 & 97.00 & 98.65 & 97.90 & \textbf{100.00} & \textbf{100.00} & 97.40 & 68.80 & 45.80 & \textbf{98.76} & 84.91 \\
					Layer-DefensiveKV & 17.86 & \textbf{90.80} & 99.60 & \textbf{99.40} & \textbf{99.00} & \textbf{98.85} & \textbf{98.45} & \textbf{100.00} & \textbf{100.00} & \textbf{100.00} & \textbf{73.00} & \textbf{47.60} & 98.56 & \textbf{86.39} \\
					
					\midrule
					\multicolumn{15}{c}{Llama-3.1-8B-Instruct, 32K Ruler, $40\%$ Cache Size}\\
					\midrule
					FullCache & 45.22 & 94.13 & 99.60 & 99.60 & 99.40 & 98.75 & 99.10 & 100.00 & 100.00 & 100.00 & 79.80 & 54.80 & 99.24 & 89.97 \\
					\arrayrulecolor{lightgray}
					\midrule
					\arrayrulecolor{black}
					Strongest Baseline CriticalKV & 49.08 & 91.93 & \textbf{99.60} & 94.00 & 54.00 & \textbf{98.75} & 98.90 & \textbf{100.00} & \textbf{100.00} & 97.40 & 68.00 & 47.60 & \textbf{99.32} & 84.51 \\
					DefensiveKV & \textbf{51.12} & \textbf{92.87} & \textbf{99.60} & \textbf{99.80} & 98.60 & 98.65 & 98.90 & \textbf{100.00} & \textbf{100.00} & \textbf{100.00} & \textbf{78.20} & 51.80 & 99.24 & 89.91 \\
					Layer-DefensiveKV & 50.24 & 92.00 & \textbf{99.60} & \textbf{99.80} & \textbf{99.20} & 98.70 & \textbf{99.05} & \textbf{100.00} & \textbf{100.00} & \textbf{100.00} & 78.00 & \textbf{53.40} & 99.12 & \textbf{89.93} \\
				
					\bottomrule
				\end{tabular}%
			}
			\vspace{-0.4cm}
		\end{table*}

		\section{Case Study: Augmenting AdaKV via Defensive Aggregation}
		
		\label{apdx:adakv_augment}
		In Algorithm~\ref{alg:defensiveagg}, DefensiveKV is built on CriticalKV’s SOTA importance scoring. To demonstrate the generalizability of defensive aggregation, we further integrate it into another cache eviction method, AdaKV. As shown in the table~\ref{tab:adakv}, across all 16 LongBench datasets, defensive aggregation consistently improved AdaKV's performance, increasing the average score from 41.39 to 44.97. These results indicate that defensive aggregation can broadly enhance  existing  cache eviction methods.

		\begin{table*}[!t]
			\vspace{0.4cm}
			\centering
			\caption{Performance comparison of AdaKV with and without defensive aggregation on LongBench.}
			\vspace{-0.2cm}
			\label{tab:adakv}
			\resizebox{\textwidth}{!}{%
				\setlength{\tabcolsep}{1.5pt}
				\begin{tabular}{@{}lllllllllllllllllllllllll@{}}
					\toprule
					
					\multirow{2}{*}[-20pt]{\makecell[l]{\hspace{10pt}\raisebox{0pt}{Method}}} & \multicolumn{3}{c}{Single-Document QA} & \multicolumn{3}{c}{Multi-Document QA} & \multicolumn{3}{c}{Summarization} & \multicolumn{3}{c}{Few-shot Learning} & \multicolumn{2}{c}{Synthetic} & \multicolumn{2}{c}{Code} & \multirow{2}{*}[-20pt]{Avg.} \\ \cmidrule(lr){2-4} \cmidrule(lr){5-7} \cmidrule(lr){8-10} \cmidrule(lr){11-13} \cmidrule(lr){14-15} \cmidrule(lr){16-17}
					
					&\rotatebox{30}{NrtvQA} & \rotatebox{30}{Qasper} & \rotatebox{30}{MF-en} & \rotatebox{30}{Hotpot}& \rotatebox{30}{2WikiQA} & \rotatebox{30}{Musique} & \rotatebox{30}{GovRep} & \rotatebox{30}{QMSum} & \rotatebox{30}{MultiNews} & \rotatebox{30}{TREC} & \rotatebox{30}{TriviaQA} & \rotatebox{30}{SAMSum} & \rotatebox{30}{PCount} & \rotatebox{30}{PR-en} & \rotatebox{30}{Lcc} & \rotatebox{30}{RB-P} & \\
					
					\midrule
					\multicolumn{18}{c}{Llama-3.1-8B-Instruct, $20\%$ Cache Size}\\
					\midrule
					Full Cache & 29.55 & 44.68 & 55.82 & 57.59 & 48.89 & 32.61 & 34.40 & 25.51 & 26.83 & 73.00 & 92.36 & 43.27 & 7.38 & 99.50 & 63.44 & 52.36 & 49.20 \\
					\arrayrulecolor{lightgray}
					\midrule
					\arrayrulecolor{black}
					
					AdaKV & 27.07 & 28.69 & 32.85 & 49.64 & 30.89 & 21.57 & 26.70 & 21.85 & 22.67 & 55.50 & 91.30 & 43.89 & 7.30 & 80.50 & 66.44 & 55.43 & 41.39 \\
					AdaKV w/. defensive agg. & \textbf{28.60} & \textbf{37.62} & \textbf{41.08} & \textbf{51.74} & \textbf{36.87} & \textbf{22.83} & \textbf{27.83} & \textbf{23.18} & \textbf{23.51} & \textbf{66.00} & \textbf{91.64} & \textbf{44.35} & \textbf{8.10} & \textbf{92.50} & \textbf{67.97} & \textbf{55.71} & \textbf{44.97} \\
					
					\bottomrule
				\end{tabular}%
			}
			\vspace{-0.4cm}
		\end{table*}

\begin{table*}[!t]
	\vspace{0.4cm}
	\centering
	\caption{Performance of DefensiveKV combined with int4 cache quantization on LongBench}
	\vspace{-0.2cm}
	\label{tab:quant}
	\resizebox{\textwidth}{!}{%
		\setlength{\tabcolsep}{1.5pt}
		\begin{tabular}{@{}lllllllllllllllllllllllll@{}}
			\toprule
			
			\multirow{2}{*}[-20pt]{\makecell[l]{\hspace{10pt}\raisebox{0pt}{Method}}} & \multicolumn{3}{c}{Single-Document QA} & \multicolumn{3}{c}{Multi-Document QA} & \multicolumn{3}{c}{Summarization} & \multicolumn{3}{c}{Few-shot Learning} & \multicolumn{2}{c}{Synthetic} & \multicolumn{2}{c}{Code} & \multirow{2}{*}[-20pt]{Avg.} \\ \cmidrule(lr){2-4} \cmidrule(lr){5-7} \cmidrule(lr){8-10} \cmidrule(lr){11-13} \cmidrule(lr){14-15} \cmidrule(lr){16-17}
			
			&\rotatebox{30}{NrtvQA} & \rotatebox{30}{Qasper} & \rotatebox{30}{MF-en} & \rotatebox{30}{Hotpot}& \rotatebox{30}{2WikiQA} & \rotatebox{30}{Musique} & \rotatebox{30}{GovRep} & \rotatebox{30}{QMSum} & \rotatebox{30}{MultiNews} & \rotatebox{30}{TREC} & \rotatebox{30}{TriviaQA} & \rotatebox{30}{SAMSum} & \rotatebox{30}{PCount} & \rotatebox{30}{PR-en} & \rotatebox{30}{Lcc} & \rotatebox{30}{RB-P} & \\
			
			\midrule
			\multicolumn{18}{c}{Llama-3.1-8B-Instruct, $20\%$ Cache Size}\\
			\midrule
			Full Cache (100\% memory) & 29.55 & 44.68 & 55.82 & 57.59 & 48.89 & 32.61 & 34.4 & 25.51 & 26.83 & 73 & 92.36 & 43.27 & 7.38 & 99.5 & 63.44 & 52.36 & 49.2 \\
			\arrayrulecolor{lightgray}
			\midrule
			\arrayrulecolor{black}
			DefensiveKV-40\% Cache (40\% memory) & 30.07 & \textbf{46.37} & \textbf{54.9} & \textbf{57.5} & \textbf{45.97} & \textbf{28.85} & 33.7 & 24.69 & \textbf{26.2} & \textbf{71.5} & \textbf{91.78} & 43.69 & \textbf{9.88} & \textbf{100} & \textbf{66.25} & 55.97 & \textbf{49.21} \\
			DefensiveKV-40\% Cache-int4 (10\% memory) & \textbf{30.63} & 44.62 & 54.44 & 56.14 & 42.9 & 28.15 & \textbf{33.79} & \textbf{25.15} & 25.92 & 70.5 & 91.28 & \textbf{43.76} & 7.55 & \textbf{100} & 65.71 & \textbf{56.23} & 48.55 \\
			
			\bottomrule
		\end{tabular}%
	}
	\vspace{-0.4cm}
\end{table*}

\section{Detailed scores of Longbench}
\label{apdx:details_longbench}
We provide detailed scores on individual datasets for 40\%, 60\% and 80\% cache sizes in Tables \ref{longbench_detail_40}, \ref{longbench_detail_60} and, \ref{longbench_detail_80}. Our DefensiveKV and Layer-DefensiveKV methods maintain nearly lossless generation quality across these settings, while other baselines fail to achieve this level of performance. 

\begin{table*}[!t]
	\centering
	\caption{Detailed scores of 16 datasets on LongBench (40\% cache size).}
	\vspace{-0.2cm}
	\label{longbench_detail_40}
	\resizebox{\textwidth}{!}{%
		\setlength{\tabcolsep}{1.5pt}
		\begin{tabular}{@{}lllllllllllllllllllllllll@{}}
			\toprule
			
			\multirow{2}{*}[-20pt]{\makecell[l]{\hspace{10pt}\raisebox{0pt}{Method}}} & \multicolumn{3}{c}{Single-Document QA} & \multicolumn{3}{c}{Multi-Document QA} & \multicolumn{3}{c}{Summarization} & \multicolumn{3}{c}{Few-shot Learning} & \multicolumn{2}{c}{Synthetic} & \multicolumn{2}{c}{Code} & \multirow{2}{*}[-20pt]{Avg.} \\ \cmidrule(lr){2-4} \cmidrule(lr){5-7} \cmidrule(lr){8-10} \cmidrule(lr){11-13} \cmidrule(lr){14-15} \cmidrule(lr){16-17} 
			
			&\rotatebox{30}{NrtvQA} & \rotatebox{30}{Qasper} & \rotatebox{30}{MF-en} & \rotatebox{30}{Hotpot}& \rotatebox{30}{2WikiQA} & \rotatebox{30}{Musique} & \rotatebox{30}{GovRep} & \rotatebox{30}{QMSum} & \rotatebox{30}{MultiNews} & \rotatebox{30}{TREC} & \rotatebox{30}{TriviaQA} & \rotatebox{30}{SAMSum} & \rotatebox{30}{PCount} & \rotatebox{30}{PR-en} & \rotatebox{30}{Lcc} & \rotatebox{30}{RB-P} & \\
			
			\midrule
			\multicolumn{18}{c}{Llama-3.1-8B-Instruct, $40\%$ Cache Size}\\
			\midrule
			Full Cache & 29.55 & 44.68 & 55.82 & 57.59 & 48.89 & 32.61 & 34.40 & 25.51 & 26.83 & 73.00 & 92.36 & 43.27 & 7.38 & 99.50 & 63.44 & 52.36 & 49.20 \\
			\arrayrulecolor{lightgray} 
			\midrule      
			\arrayrulecolor{black} 
			DuoAttention & 28.83 & 42.51 & 53.35 & 55.76 & 45.37 & \underline{30.16} & 32.26 & \underline{25.07} & 25.51 & \underline{71.50} & 88.44 & 41.12 & 3.67 & 99.50 & \textbf{68.81} & \underline{58.85} & 48.17 \\
			StreamingLLM & 24.32 & 28.77 & 28.74 & 43.75 & 30.71 & 18.55 & 30.26 & 21.86 & 24.88 & 65.50 & 92.24 & 41.65 & 2.92 & 46.00 & 66.54 & \textbf{61.42} & 39.26 \\
			SnapKV & 27.72 & 36.42 & 38.31 & 54.92 & 40.02 & 26.95 & 30.60 & 23.33 & 24.31 & 56.00 & \underline{92.31} & \textbf{43.92} & 7.62 & 96.50 & 65.95 & 53.77 & 44.92 \\
			CAKE & \underline{30.43} & 37.57 & 45.51 & \underline{57.13} & 40.08 & 25.95 & 30.33 & 23.80 & 24.96 & 61.00 & 91.83 & 43.46 & 6.70 & \textbf{100.00} & 65.56 & 52.66 & 46.06 \\
			AdaKV & 28.36 & 37.58 & 41.35 & 54.80 & 41.47 & 29.02 & 30.18 & 23.72 & 24.68 & 63.50 & 91.73 & 43.57 & 7.27 & 95.00 & 64.93 & 54.75 & 45.74 \\
			CriticalKV & 30.10 & 40.14 & 49.03 & 55.95 & \underline{46.22} & \textbf{30.42} & 31.49 & 24.34 & 25.15 & 67.50 & \textbf{92.39} & 43.20 & 8.08 & 99.00 & 64.68 & 55.08 & 47.67 \\
			DefensiveKV & 30.07 & \textbf{46.37} & \underline{54.90} & \textbf{57.50} & 45.97 & 28.85 & \underline{33.70} & 24.69 & \underline{26.20} & \underline{71.50} & 91.78 & \underline{43.69} & \textbf{9.88} & \underline{100.00} & 66.25 & 55.97 & \underline{49.21} \\
			Layer-DefensiveKV & \textbf{30.94} & \underline{43.84} & \textbf{55.01} & 56.36 & \textbf{49.14} & 29.88 & \textbf{34.09} & \textbf{25.71} & \textbf{26.64} & \textbf{72.00} & 91.49 & 42.96 & \underline{8.56} & 99.50 &  \underline{67.30} & 57.98 & \textbf{49.46} \\

			\midrule
			\multicolumn{18}{c}{Mistral-7B-Instruct-v0.3, $40\%$ Cache Size}\\
			\midrule
			Full Cache & 27.02 & 38.19 & 50.22 & 50.75 & 37.41 & 27.92 & 34.45 & 25.76 & 26.37 & 76.00 & 89.01 & 46.89 & 6.50 & 97.00 & 66.04 & 60.47 & 47.50 \\
			\arrayrulecolor{lightgray} 
			\midrule      
			\arrayrulecolor{black} 
			DuoAttention & 20.37 & 26.96 & 49.69 & 48.92 & 34.96 & 20.16 & 29.14 & 21.74 & 24.86 & 73.50 & 87.39 & 44.06 & 3.00 & 93.00 & 63.95 & 58.16 & 43.74 \\
			StreamingLLM & 19.71 & 24.85 & 29.54 & 42.10 & 34.34 & 18.53 & 31.03 & 21.60 & 24.05 & 40.00 & 83.25 & 41.22 & 3.50 & 45.50 & 34.40 & 46.59 & 33.76 \\
			SnapKV & 25.32 & 30.09 & 39.68 & 49.16 & 33.66 & 22.38 & 30.60 & 22.39 & 24.22 & 65.00 & \textbf{89.37} & 47.17 & 5.00 & 94.50 & 65.60 & 60.66 & 44.05 \\
			CAKE & 25.02 & 31.82 & 45.30 & 48.42 & 31.94 & 21.73 & 31.70 & 23.15 & 24.77 & 68.50 & 89.22 & 46.34 & 4.00 & 92.50 & 64.99 & 60.21 & 44.35 \\
			AdaKV & 24.76 & 31.86 & 41.79 & \underline{49.59} & 32.95 & 20.20 & 30.51 & 22.94 & 24.42 & 68.00 & 88.96 & \underline{47.29} & \textbf{5.50} & 96.50 & 65.54 & 59.99 & 44.42 \\
			CriticalKV & \textbf{26.97} & 34.32 & 47.50 & 48.00 & \textbf{38.07} & 24.64 & 31.51 & 24.79 & 25.14 & 72.50 &  \textbf{89.37} & \textbf{47.86} & 4.53 & 95.50 & 65.59 & 59.68 & 46.00 \\
			DefensiveKV & 25.45 & \underline{39.24} & \textbf{51.42} & \textbf{50.13} & 34.89 & \textbf{26.36} & \textbf{34.43} & \textbf{25.42} & \underline{26.17} & \underline{75.50} & 89.21 & 46.59 & \underline{5.05} & \textbf{98.00} & \underline{66.54} & \underline{61.65} & \underline{47.25} \\
			Layer-DefensiveKV & \underline{26.29} & \textbf{40.49} & \underline{50.92} & 48.85 & \underline{36.34} & \underline{26.02} & \underline{34.33} & \underline{25.28} & \textbf{26.62} & \textbf{76.00} & \underline{89.36} & 46.71 & 3.86 & \underline{97.00} & \textbf{66.65} & \textbf{62.12} & \textbf{47.30} \\

			\midrule
			\multicolumn{18}{c}{Qwen2.5-32B-Instruct, $40\%$  Cache Size}\\
			\midrule
			Full Cache & 30.88 & 46.13 & 52.87 & 63.59 & 59.75 & 38.78 & 32.59 & 24.35 & 24.95 & 72.00 & 88.26 & 47.05 & 12.50 & 100.00 & 49.64 & 34.24 & 48.60 \\
			\arrayrulecolor{lightgray} 
			\midrule      
			\arrayrulecolor{black} 
			DuoAttention & N/A & N/A & N/A & N/A & N/A & N/A & N/A & N/A & N/A & N/A & N/A & N/A & N/A & N/A & N/A & N/A & N/A \\
			StreamingLLM & 23.95 & 25.49 & 28.48 & 49.89 & 44.66 & 24.15 & 30.32 & 20.60 & 24.12 & 68.00 & 87.08 & 45.90 & \underline{11.62} & 46.00 & \textbf{57.38} & 28.36 & 38.50 \\
			SnapKV & 27.18 & 33.99 & 38.19 & 61.11 & 53.18 & 38.01 & 29.95 & 20.84 & 22.95 & 66.00 & \textbf{89.16} & 47.02 & \textbf{12.00} & 97.75 & 52.19 & 34.75 & 45.27 \\
			CAKE & 28.13 & 35.18 & 39.22 & 60.17 & 56.22 & 36.25 & 30.33 & 21.03 & 23.88 & 66.00 & 88.74 & 46.68 & 10.50 & 98.75 & 52.20 & \underline{35.14} & 45.53 \\
			AdaKV & 27.16 & 31.83 & 37.07 & 60.81 & 53.92 & 37.66 & 29.84 & 20.79 & 23.07 & 68.50 & \underline{88.93} & \underline{47.40} & \textbf{12.00} & 97.00 & 53.05 & 34.56 & 45.22 \\
			CriticalKV & \textbf{31.65} & 35.23 & 42.70 & 59.24 & 59.26 & 40.10 & 31.25 & 22.67 & 24.02 & 71.00 & 88.80 & 46.91 & 11.00 & 99.75 & 51.76 & \textbf{35.70} & 46.94 \\
			DefensiveKV & 30.97 & \underline{43.88} & \underline{51.15} & \textbf{65.16} & \textbf{63.66} & \textbf{42.20} & \textbf{32.82} & \underline{23.50} & \textbf{24.67} & \underline{74.00} & 88.88 & \textbf{47.50} & 10.00 & \underline{99.88} & \underline{53.50} & 34.24 & \textbf{49.13} \\
			Layer-DefensiveKV & \underline{31.61} & \textbf{44.02} & \textbf{51.86} & \underline{64.35} & \underline{61.10} & \underline{40.88} & \underline{31.94} & \textbf{24.26} & \underline{24.62} & \textbf{74.50} & 88.75 & 46.76 & \textbf{12.00} & \textbf{100.00} & 50.88 & 34.64 & \underline{48.89} \\
			
			\bottomrule
		\end{tabular}%
	}
	
\end{table*}

\begin{table*}[t!]
	
	\centering
	\caption{Detailed scores of 16 datasets on LongBench (60\% cache size).}
	\vspace{-0.2cm}
	\label{longbench_detail_60}
	\resizebox{\textwidth}{!}{%
		\setlength{\tabcolsep}{1.5pt}
		\begin{tabular}{@{}lllllllllllllllllllllllll@{}}
			\toprule
			
			\multirow{2}{*}[-20pt]{\makecell[l]{\hspace{10pt}\raisebox{0pt}{Method}}} & \multicolumn{3}{c}{Single-Document QA} & \multicolumn{3}{c}{Multi-Document QA} & \multicolumn{3}{c}{Summarization} & \multicolumn{3}{c}{Few-shot Learning} & \multicolumn{2}{c}{Synthetic} & \multicolumn{2}{c}{Code} & \multirow{2}{*}[-20pt]{Avg.} \\ \cmidrule(lr){2-4} \cmidrule(lr){5-7} \cmidrule(lr){8-10} \cmidrule(lr){11-13} \cmidrule(lr){14-15} \cmidrule(lr){16-17} 
			
			&\rotatebox{30}{NrtvQA} & \rotatebox{30}{Qasper} & \rotatebox{30}{MF-en} & \rotatebox{30}{Hotpot}& \rotatebox{30}{2WikiQA} & \rotatebox{30}{Musique} & \rotatebox{30}{GovRep} & \rotatebox{30}{QMSum} & \rotatebox{30}{MultiNews} & \rotatebox{30}{TREC} & \rotatebox{30}{TriviaQA} & \rotatebox{30}{SAMSum} & \rotatebox{30}{PCount} & \rotatebox{30}{PR-en} & \rotatebox{30}{Lcc} & \rotatebox{30}{RB-P} & \\
			
			\midrule
			\multicolumn{18}{c}{Llama-3.1-8B-Instruct, $60\%$ Cache Size}\\
			\midrule
			Full Cache & 29.55 & 44.68 & 55.82 & 57.59 & 48.89 & 32.61 & 34.40 & 25.51 & 26.83 & 73.00 & 92.36 & 43.27 & 7.38 & 99.50 & 63.44 & 52.36 & 49.20 \\
			\arrayrulecolor{lightgray} 
			\midrule      
			\arrayrulecolor{black} 
			DuoAttention & 28.77 & 43.00 & 54.41 & 55.90 & 46.18 & 28.61 & 33.86 & 25.10 & \underline{26.80} & 72.00 & 91.16 & 43.37 & \underline{10.50} & \textbf{99.50} & \underline{66.23} & \underline{56.03} & 48.84 \\
			StreamingLLM & 25.21 & 39.92 & 33.53 & 49.72 & 39.98 & 22.62 & 31.51 & 23.12 & 25.91 & 69.50 & \underline{92.27} & 42.75 & 3.08 & 57.50 & \textbf{66.27} & \textbf{61.75} & 42.79 \\
			SnapKV & 28.92 & 40.35 & 48.00 & 56.79 & 48.60 & 30.12 & 32.54 & 24.20 & 25.58 & 64.50 & 91.64 & \textbf{44.53} & 8.85 &  \underline{99.00} & 65.03 & 53.81 & 47.65 \\
			CAKE & 29.99 & 41.87 & 53.09 & 55.39 & 42.83 & \underline{32.17} & 32.12 & 24.87 & 25.61 & 66.50 & \textbf{92.50} & 43.37 & 7.96 &\textbf{99.50} & 64.11 & 52.62 & 47.78 \\
			AdaKV & 30.10 & \underline{43.61} & 51.20 & 56.37 & \underline{49.70} & 30.18 & 32.37 & 24.38 & 25.54 & 66.50 & 91.48 & 43.87 & 8.02 & \textbf{99.50} & 63.91 & 54.73 & 48.22 \\
			CriticalKV & \underline{30.31} & 43.54 & 52.82 & \textbf{57.30} & 49.09 & 31.78 & 33.48 & \underline{25.18} & 26.00 & \underline{72.50} & 91.80 & \underline{43.95} & 7.47 &\textbf{99.50} & 64.05 & 54.92 & 48.98 \\
			DefensiveKV & \textbf{30.88} & 43.20 & \underline{55.17} & 55.85 & \textbf{50.17} & 31.84 & \underline{34.79} & \textbf{25.29} & \textbf{26.84} & \textbf{73.00} & 92.14 & 43.28 & 9.05 & \textbf{99.50} & 63.88 & 56.02 & \underline{49.43} \\
			Layer-DefensiveKV & 29.95 & \textbf{44.11} & \textbf{56.78} & \underline{57.11} & 47.47 & \textbf{32.58} & \textbf{34.81} & 25.11 & \underline{26.80} & 72.00 & 91.83 & 43.14 & \textbf{11.10} & \textbf{99.50} & 64.21 & 55.81 & \textbf{49.52} \\
			
			\midrule
			\multicolumn{18}{c}{Mistral-7B-Instruct-v0.3, $60\%$ Cache Size}\\
			\midrule
			Full Cache & 27.02 & 38.19 & 50.22 & 50.75 & 37.41 & 27.92 & 34.45 & 25.76 & 26.37 & 76.00 & 89.01 & 46.89 & 6.50 & 97.00 & 66.04 & 60.47 & 47.50 \\
			\arrayrulecolor{lightgray} 
			\midrule      
			\arrayrulecolor{black} 
			DuoAttention & \textbf{28.86} & 36.56 & 50.54 & \textbf{53.32} & \textbf{39.19} & \textbf{29.22} & 33.91 & 25.16 & 26.77 & \underline{76.00} & 87.57 & 45.40 & 5.00 & 95.00 & 64.75 & 58.91 & 47.26 \\
			StreamingLLM & 21.48 & 30.55 & 35.40 & 46.84 & 36.62 & 22.89 & 32.20 & 22.70 & 25.06 & 44.50 & 82.26 & 41.87 & 3.00 & 57.00 & 34.42 & 47.38 & 36.51 \\
			SnapKV & 25.44 & 33.77 & 45.63 & \underline{52.52} & 34.30 & 25.94 & 32.61 & 24.59 & 25.37 & 68.00 & \textbf{89.41} & \underline{47.12} & 5.00 & 95.50 & 66.22 & 59.47 & 45.68 \\
			CAKE & 27.17 & 36.69 & 48.77 & 50.91 & 38.21 & 23.16 & 33.53 & 23.73 & 26.06 & 75.00 & 88.46 & 46.93 & \underline{5.56} & 95.00 & 66.23 & 60.76 & 46.64 \\
			AdaKV & 25.47 & 34.87 & 47.21 & 48.61 & 35.64 & 25.97 & 32.38 & 24.24 & 25.54 & 70.50 & 88.91 & 47.00 & 5.06 & 96.00 & 65.85 & 60.20 & 45.84 \\
			CriticalKV & 26.06 & 37.46 & 50.28 & 50.41 & 37.11 & 26.80 & 33.28 & \underline{25.56} & 25.90 & 75.50 & 88.81 & \textbf{47.54} & \textbf{6.35} & \underline{97.00} & 65.30 & 59.61 & 47.06 \\
			DefensiveKV & \underline{27.82} & \textbf{39.13} & \underline{51.50} & 50.78 & \underline{38.39} & \underline{27.37} & \underline{34.29} & 25.41 & \textbf{26.78} & \underline{76.00} & \underline{89.21} & 46.89 & 4.60 & \textbf{98.00} & \underline{66.27} & \underline{61.39} & \textbf{47.74} \\
			Layer-DefensiveKV & 25.59 & \underline{38.95} & \textbf{51.99} & 50.81 & 37.09 & 25.35 & \textbf{34.50} & \textbf{25.68} & \underline{26.78} & \textbf{77.50} & 89.04 & 46.96 & 4.00 & \underline{97.00} & \textbf{67.00} & \textbf{61.50} & \underline{47.48} \\

			\midrule
			\multicolumn{18}{c}{Qwen2.5-32B-Instruct, $60\%$  Cache Size}\\
			\midrule
			Full Cache & 30.88 & 46.13 & 52.87 & 63.59 & 59.75 & 38.78 & 32.59 & 24.35 & 24.95 & 72.00 & 88.26 & 47.05 & 12.50 & 100.00 & 49.64 & 34.24 & 48.60 \\
			\arrayrulecolor{lightgray} 
			\midrule      
			\arrayrulecolor{black} 
			DuoAttention & N/A & N/A & N/A & N/A & N/A & N/A & N/A & N/A & N/A & N/A & N/A & N/A & N/A & N/A & N/A & N/A & N/A \\
			StreamingLLM & 26.06 & 34.49 & 33.06 & 57.33 & 46.75 & 30.57 & 30.39 & 21.93 & 24.83 & 72.50 & 87.01 & 46.97 & 10.62 & 57.50 & \textbf{58.09} & 29.86 & 41.75 \\
			SnapKV & 30.42 & 38.55 & 45.71 & 62.38 & 59.78 & 38.36 & 31.77 & 23.19 & 24.37 & 69.00 & 88.64 & \textbf{47.32} & \underline{12.00} & \textbf{100.00} & 52.09 & \underline{34.65} & 47.39 \\
			CAKE & 30.07 & 41.66 & 45.96 & \underline{64.45} & 59.80 & 36.18 & 31.64 & 22.70 & 24.80 & 68.00 & \textbf{89.11} & 47.06 & 10.50 & \textbf{100.00} & 52.56 & \textbf{35.24} & 47.48 \\
			AdaKV & 29.58 & 38.02 & 45.63 & 61.46 & 57.33 & 37.91 & 31.43 & 22.40 & 24.13 & 71.00 & \underline{88.86} & 46.64 & 11.50 & \textbf{100.00} & 52.54 & 33.95 & 47.02 \\
			CriticalKV & \textbf{33.01} & 41.97 & 48.07 & 62.35 & 62.58 & \underline{39.79} & 32.45 & 23.69 & 24.59 & 72.00 & 88.83 & 46.95 & \underline{12.00} &  \underline{99.88} & 52.03 & 34.02 & 48.39 \\
			DefensiveKV & \underline{32.10} & \textbf{46.38} & \underline{51.50} & \textbf{64.50} & \underline{63.07} & 39.13 & \textbf{32.77} & \textbf{24.41} & \textbf{24.96} & \textbf{73.50} & 88.69 & \underline{47.07} & 11.00 &  \underline{99.88} &  \underline{53.35} & 33.72 & \underline{49.13} \\
			Layer-DefensiveKV & 31.28 & \underline{46.18} & \textbf{51.97} & 63.24 & \textbf{63.20} & \textbf{39.80} & \underline{32.56} & \underline{24.37} & \underline{24.96} & \underline{73.00} & 88.65 & 46.73 & \textbf{13.50} & \textbf{100.00} & 52.74 & 34.06 & \textbf{49.14} \\

			\bottomrule
		\end{tabular}%
	}
	
\end{table*}

\begin{table*}[t!]
	\vspace{-0.2cm}
	\centering
	\caption{Detailed scores of 16 datasets on LongBench (80\% cache size).}
	\vspace{-0.2cm}
	\label{longbench_detail_80}
	\resizebox{\textwidth}{!}{%
		\setlength{\tabcolsep}{1.5pt}
		\begin{tabular}{@{}lllllllllllllllllllllllll@{}}
			\toprule
			
			\multirow{2}{*}[-20pt]{\makecell[l]{\hspace{10pt}\raisebox{0pt}{Method}}} & \multicolumn{3}{c}{Single-Document QA} & \multicolumn{3}{c}{Multi-Document QA} & \multicolumn{3}{c}{Summarization} & \multicolumn{3}{c}{Few-shot Learning} & \multicolumn{2}{c}{Synthetic} & \multicolumn{2}{c}{Code} & \multirow{2}{*}[-20pt]{Avg.} \\ \cmidrule(lr){2-4} \cmidrule(lr){5-7} \cmidrule(lr){8-10} \cmidrule(lr){11-13} \cmidrule(lr){14-15} \cmidrule(lr){16-17} 
			
			&\rotatebox{30}{NrtvQA} & \rotatebox{30}{Qasper} & \rotatebox{30}{MF-en} & \rotatebox{30}{Hotpot}& \rotatebox{30}{2WikiQA} & \rotatebox{30}{Musique} & \rotatebox{30}{GovRep} & \rotatebox{30}{QMSum} & \rotatebox{30}{MultiNews} & \rotatebox{30}{TREC} & \rotatebox{30}{TriviaQA} & \rotatebox{30}{SAMSum} & \rotatebox{30}{PCount} & \rotatebox{30}{PR-en} & \rotatebox{30}{Lcc} & \rotatebox{30}{RB-P} & \\
			
			\midrule
			\multicolumn{18}{c}{Llama-3.1-8B-Instruct, $80\%$ Cache Size}\\
			\midrule
			Full Cache & 29.55 & 44.68 & 55.82 & 57.59 & 48.89 & 32.61 & 34.40 & 25.51 & 26.83 & 73.00 & 92.36 & 43.27 & 7.38 & 99.50 & 63.44 & 52.36 & 49.20 \\
			\arrayrulecolor{lightgray} 
			\midrule      
			\arrayrulecolor{black} 
			DuoAttention & \textbf{30.04} & 44.49 & 55.77 & 57.51 & \underline{48.96} & 31.78 & 34.51 & \textbf{25.29} & 26.93 & \underline{73.00} & 91.35 & 43.28 & 8.08 & \textbf{100.00} & 63.19 & \underline{55.85} & 49.38 \\
			StreamingLLM & 28.57 & 43.96 & 37.87 & 52.57 & 44.06 & 26.50 & 32.88 & 24.23 & 26.58 & 70.50 & 90.52 & 43.33 & 4.03 & 83.50 & \textbf{65.34} & \textbf{60.57} & 45.94 \\
			SnapKV & 29.87 & 44.58 & 52.36 & 57.31 & 48.33 & 30.85 & 33.79 & 24.39 & 26.44 & 68.50 & 91.47 & \underline{43.71} & 8.33 & \underline{99.50} & \underline{64.66} & 53.47 & 48.60 \\
			CAKE & 29.53 & 43.76 & 56.26 & 57.28 & 47.81 & 30.71 & 33.26 & 25.14 & 26.59 & 72.50 & \textbf{92.75} & 42.82 & \textbf{9.60} & \underline{99.50} & 63.75 & 51.87 & 48.95 \\
			AdaKV & \underline{29.93} & \textbf{44.89} & \textbf{57.17} & 56.55 & 48.34 & \underline{32.59} & 34.13 & 25.14 & 26.36 & \underline{73.00} & 91.80 & 43.54 & \underline{8.66} & \underline{99.50} & 64.12 & 53.32 & 49.31 \\
			CriticalKV & 29.73 & 44.66 & 55.66 & \textbf{58.19} & 48.52 & 32.24 & 34.70 & \underline{25.27} & 26.56 & \textbf{73.50} & \underline{92.30} & \textbf{43.75} & 8.09 & \underline{99.50} & 63.90 & 54.04 & \textbf{49.41} \\
			DefensiveKV & 29.63 & 44.49 & \underline{56.70} & 57.41 & \textbf{49.49} & 31.08 & \textbf{34.97} & 25.23 & \textbf{27.25} & \underline{73.00} & 92.03 & 43.06 & 8.07 & \underline{99.50} & 63.90 & 54.53 & \underline{49.40} \\
			Layer-DefensiveKV & 29.63 & \underline{44.88} & 56.52 & \underline{58.18} & 48.10 & \textbf{32.85} & \underline{34.76} & 24.98 & \underline{27.20} & 72.50 & 91.78 & 42.98 & 8.62 & \underline{99.50} & 63.27 & 52.44 & 49.26 \\

			\midrule
			\multicolumn{18}{c}{Mistral-7B-Instruct-v0.3, $80\%$ Cache Size}\\
			\midrule
			Full Cache & 27.02 & 38.19 & 50.22 & 50.75 & 37.41 & 27.92 & 34.45 & 25.76 & 26.37 & 76.00 & 89.01 & 46.89 & 6.50 & 97.00 & 66.04 & 60.47 & 47.50 \\
			\arrayrulecolor{lightgray} 
			\midrule      
			\arrayrulecolor{black} 
			DuoAttention & 26.07 & 36.33 & 50.03 & \textbf{51.37} & 36.30 & 26.79 & 33.95 & \textbf{25.90} & \textbf{26.61} & 76.00 & 88.91 & 47.11 & 4.50 & \underline{97.50} & 65.53 & 60.53 & 47.09 \\
			StreamingLLM & 23.78 & 35.71 & 38.09 & 50.73 & \underline{37.79} & 24.82 & 33.51 & 24.31 & 25.85 & 50.00 & 83.38 & 42.98 & 2.65 & 82.00 & 34.22 & 45.51 & 39.71 \\
			SnapKV & 26.42 & 36.01 & 49.00 & 50.06 & 36.40 & \underline{28.56} & 34.08 & 24.54 & 25.89 & 73.50 & 88.91 & 46.86 & 6.00 & 96.00 & \underline{66.28} & 60.67 & 46.82 \\
			CAKE & 26.62 & 38.22 & 50.20 & 50.29 & 36.40 & 24.78 & \underline{34.28} & 25.63 & 26.02 & 76.00 & 88.91 & 46.41 & 4.56 & 95.00 & \textbf{66.70} & 60.29 & 46.89 \\
			AdaKV & 26.77 & 34.52 & 48.73 & 50.25 & 36.59 & \textbf{28.57} & 33.45 & 24.90 & 26.20 & \textbf{76.50} & 88.91 & 47.26 & \textbf{6.50} & 96.50 & 66.06 & \textbf{61.37} & 47.07 \\
			CriticalKV & \underline{27.34} & 36.72 & 49.04 & \underline{51.26} & 36.94 & 27.13 & 33.85 & 25.32 & 25.88 & \textbf{76.50} & 88.91 & \underline{47.28} & \underline{6.05} & \underline{97.50} & 65.78 & 59.56 & 47.19 \\
			DefensiveKV & \textbf{27.79} & \textbf{38.29} & \underline{50.34} & 50.86 & \textbf{37.84} & 27.63 & 34.24 & \underline{25.87} & 26.25 & 75.50 & \textbf{89.21} & \textbf{47.29} & 6.00 & 97.00 & 66.14 & 60.95 & \underline{47.58} \\
			Layer-DefensiveKV & 27.06 & \underline{38.23} & \textbf{51.56} & 50.76 & 36.48 & 28.32 & \textbf{34.54} & 25.37 & \underline{26.44} & 76.00 & \underline{89.04} & 47.21 & 5.50 & \textbf{98.00} & 66.10 & \underline{61.13} & \textbf{47.61} \\

			\midrule
			\multicolumn{18}{c}{Qwen2.5-32B-Instruct, $80\%$  Cache Size}\\
			\midrule
			Full Cache & 30.88 & 46.13 & 52.87 & 63.59 & 59.75 & 38.78 & 32.59 & 24.35 & 24.95 & 72.00 & 88.26 & 47.05 & 12.50 & 100.00 & 49.64 & 34.24 & 48.60 \\
			\arrayrulecolor{lightgray} 
			\midrule      
			\arrayrulecolor{black} 
			DuoAttention & N/A & N/A & N/A & N/A & N/A & N/A & N/A & N/A & N/A & N/A & N/A & N/A & N/A & N/A & N/A & N/A & N/A \\
			StreamingLLM & 27.44 & 43.24 & 35.78 & 59.73 & 51.76 & 33.23 & 31.32 & 22.84 & \textbf{25.40} & \textbf{74.50} & 86.83 & \textbf{48.18} & 11.00 & 82.00 & \textbf{55.92} & 32.59 & 45.11 \\
			SnapKV & 30.97 & 42.95 & 49.08 & \underline{63.92} & 59.10 & 39.73 & 32.00 & 23.33 & 24.75 & 71.00 & 88.48 & \underline{47.48} & \underline{12.00} & \textbf{100.00} & 51.02 & 33.55 & 48.09 \\
			CAKE & 30.32 & 45.68 & \underline{52.23} & \textbf{63.96} & 61.16 & \underline{39.91} & 32.17 & 23.67 & 25.06 & 70.50 & \textbf{88.77} & 47.39 & 11.00 & \textbf{100.00} & 51.02 & \textbf{34.84} & 48.60 \\
			AdaKV & 30.42 & 42.62 & 48.92 & 62.98 & 60.45 & 39.19 & 31.77 & 23.22 & 24.82 & 71.00 & 88.69 & 46.87 & 10.50 & \textbf{100.00} & 50.18 & 33.60 & 47.83 \\
			CriticalKV & 31.09 & 45.47 & 50.40 & 63.29 & \textbf{61.80} & \textbf{39.95} & 32.35 & 23.98 & 24.86 & \underline{72.00} & 88.50 & 46.92 & 11.50 & \textbf{100.00} & 50.56 & \underline{34.53} & 48.58 \\
			DefensiveKV & \textbf{31.34} & \textbf{46.43} & 51.92 & 63.33 & \underline{61.45} & 39.12 & \underline{32.70} & \textbf{24.30} & \underline{25.22} & \underline{72.00} & 88.39 & 47.14 & 11.00 & \textbf{100.00} & \underline{53.48} & 33.81 & \textbf{48.85} \\
			Layer-DefensiveKV & \underline{31.13} & \underline{46.01} & \textbf{52.46} & 63.29 & 60.22 & 38.90 & \textbf{32.85} & \underline{24.18} & 25.03 & \underline{72.00} & \underline{88.73} & 46.56 & \textbf{12.50} & \textbf{100.00} & 52.55 & 33.53 & \underline{48.75} \\

			\bottomrule
		\end{tabular}%
	}
	\vspace{-0.4cm}
\end{table*}

\section{The effectiveness of Defensive aggregation strategy}
\label{apdx:more_main}
\begin{figure}[t]  
	\vspace{-0.4cm}  
	\centering  
	\begin{subfigure}[b]{\textwidth}  
		\centering  
		\includegraphics[width=0.8\textwidth]{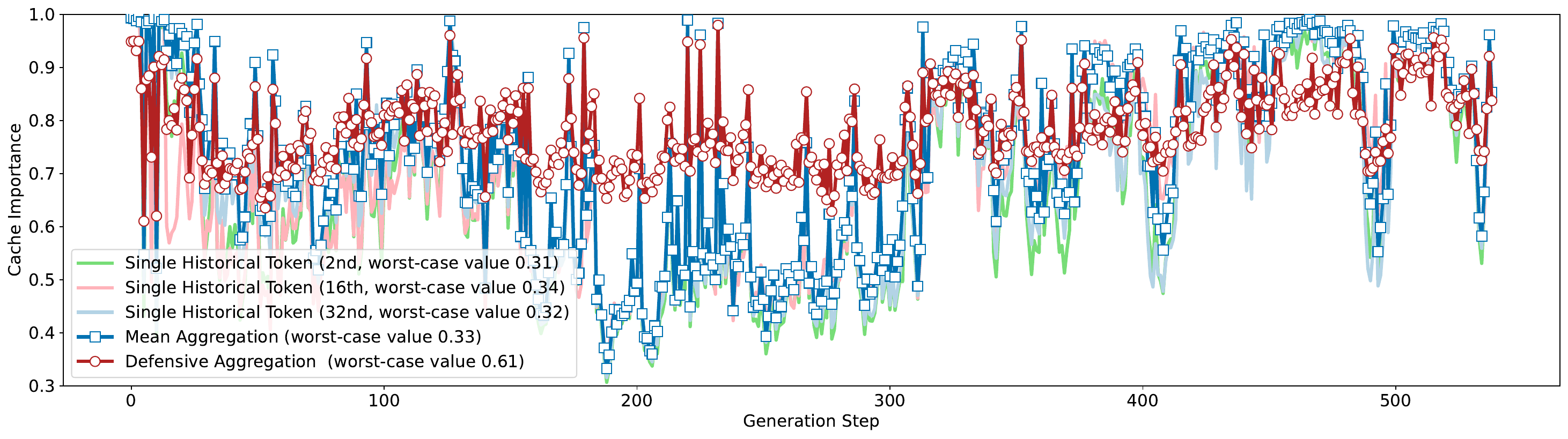}  
		\vspace{-0.3cm}  
		\caption{layer 13}  
		\label{fig:subfig-a}  
	\end{subfigure}  
	\begin{subfigure}[b]{\textwidth}  
		\centering  
		\includegraphics[width=0.8\textwidth]{./figures/intro_contrast_plots/fixbudgetFalse_frac0.5_poolingTrue_gov_report_contrast_plots/sample_19/layer_14_plot.pdf}  
		\vspace{-0.3cm}  
		\caption{layer 14}  
	\end{subfigure} 
	\begin{subfigure}[b]{\textwidth}  
		\centering  
		\includegraphics[width=0.8\textwidth]{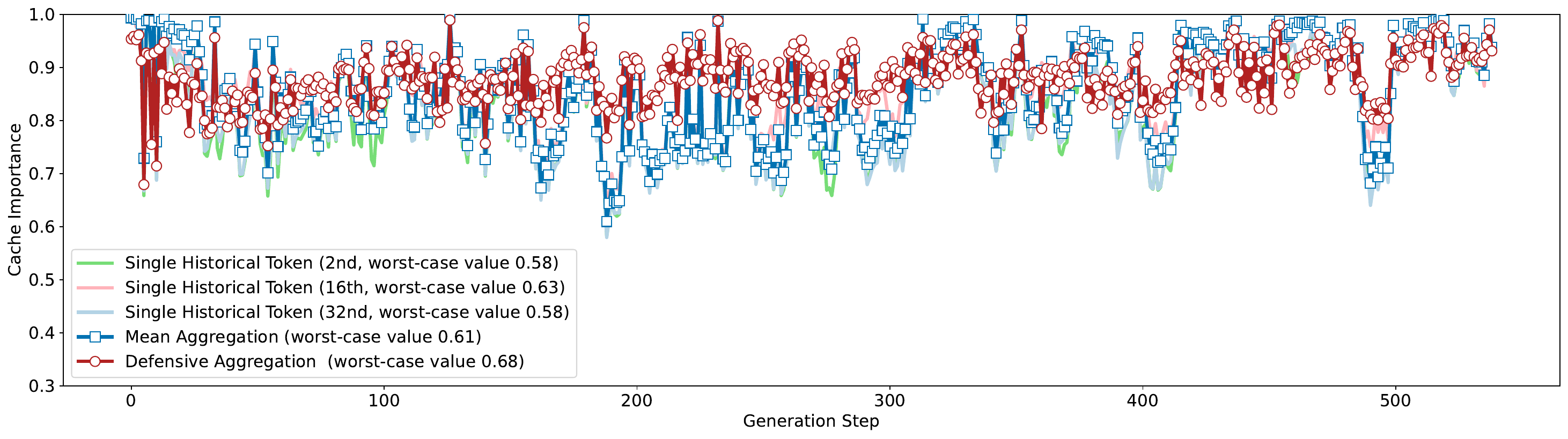}  
		\vspace{-0.3cm}  
		\caption{layer 15}  
		\label{fig:subfig-b}  
	\end{subfigure}  
	\begin{subfigure}[b]{\textwidth}  
		\centering  
		\includegraphics[width=0.8\textwidth]{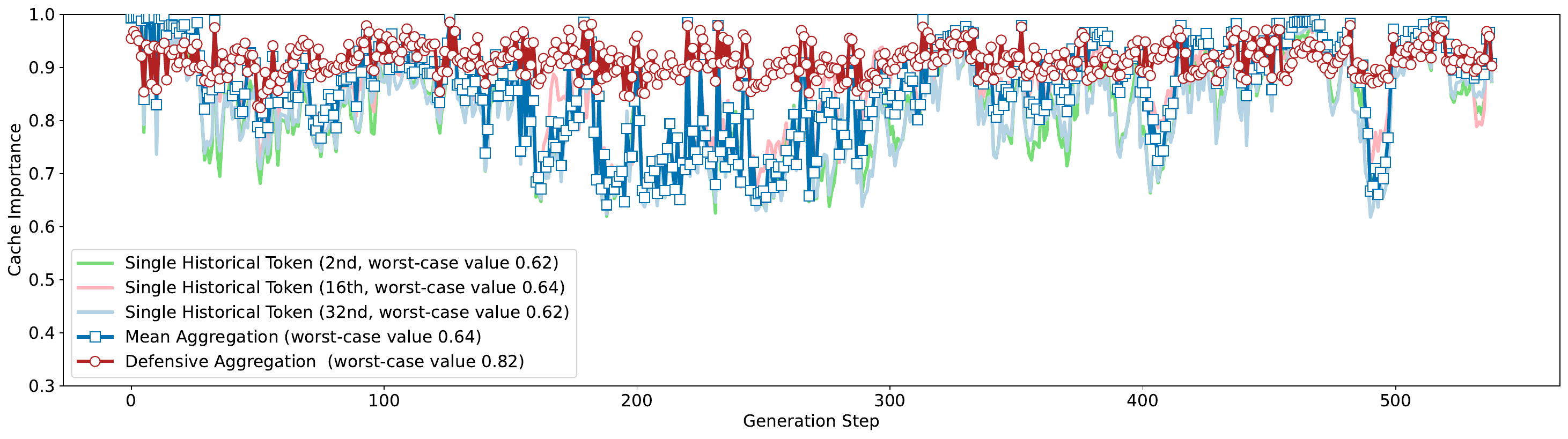}  
		\vspace{-0.3cm}  
		\caption{layer 16}  
		\label{fig:subfig-c}  
	\end{subfigure}  
	\begin{subfigure}[b]{\textwidth}  
		\centering  
		\includegraphics[width=0.8\textwidth]{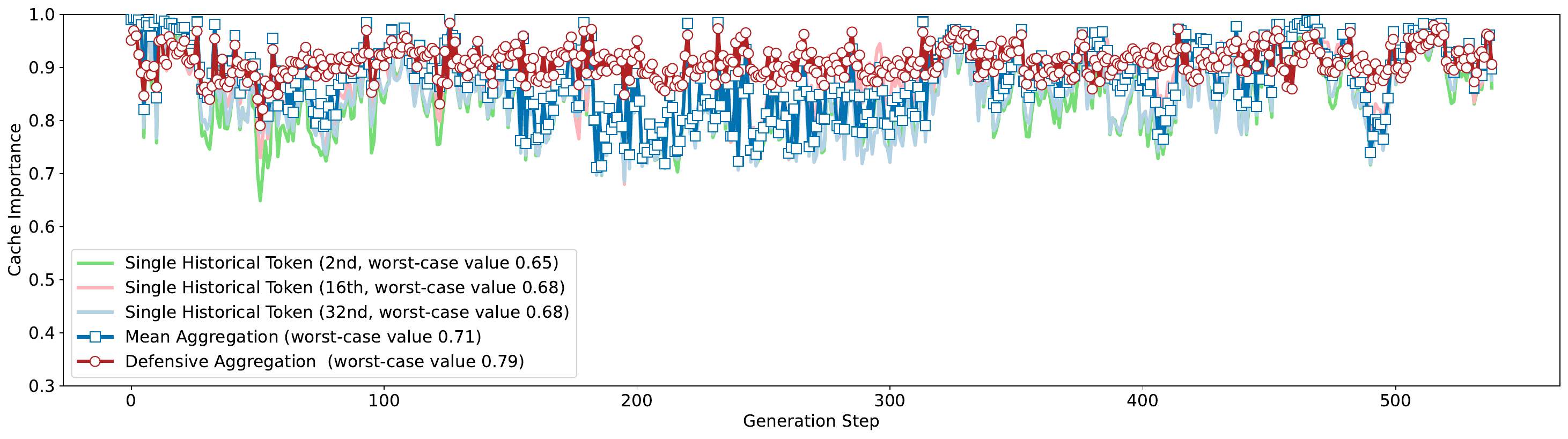}  
		\vspace{-0.3cm}  
		\caption{ layer 17}  
		\label{fig:subfig-d}  
	\end{subfigure} 
		\begin{subfigure}[b]{\textwidth}  
		\centering  
		\includegraphics[width=0.8\textwidth]{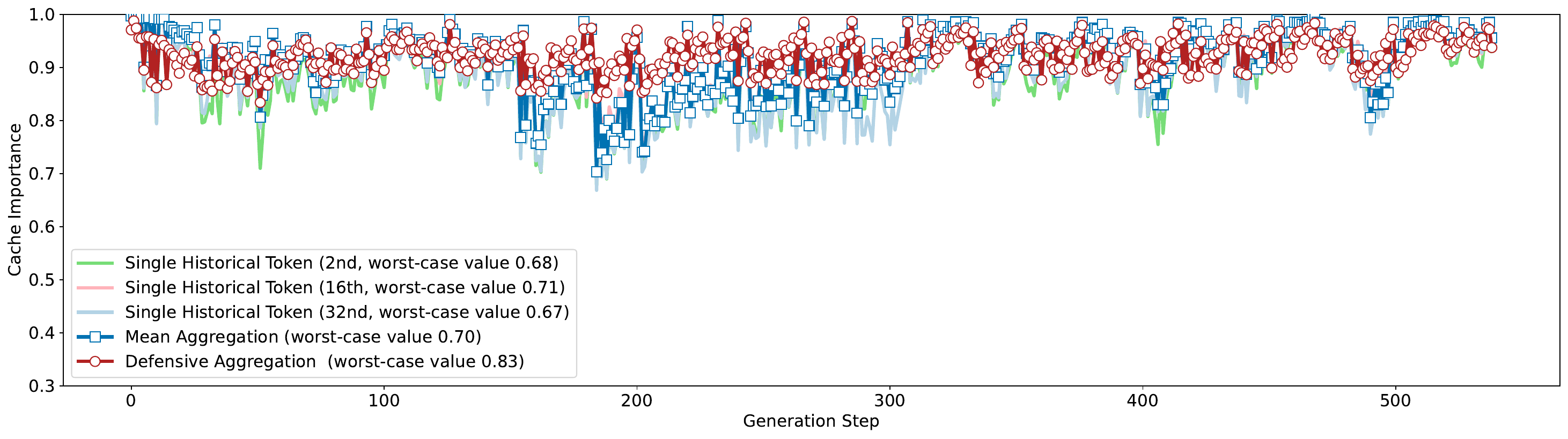}  
		\vspace{-0.3cm}  
		\caption{ layer 18}  
		\label{fig:subfig-d}  
	\end{subfigure} 
	\caption{Visualization across different layers using Llama-3.1-8B with 50\% cache size.}  
	\label{fig:more_main}  
\end{figure}

To complement Figure~\ref{fig:main} in the main text, Figure~\ref{fig:more_main} provides additional visualizations demonstrating that defensive aggregation offers greater robustness than mean aggregation under a 50\% cache size. The results reveal that this fragility is prevalent across numerous layers. Both "Single Historical token" and "Mean aggregation" methods exhibit high sensitivity to this fragility, leading to poor worst-case performance. In contrast, defensive aggregation effectively mitigates this issue, consistently maintaining higher worst-case values.

\section{Further Elaboration of the Fragile  Stability Assumption}
\label{apdx:more_assumption}

\begin{figure}  
	\centering
	\begin{minipage}[c]{0.8\textwidth}  
		\centering    
				\includegraphics[width=\textwidth]{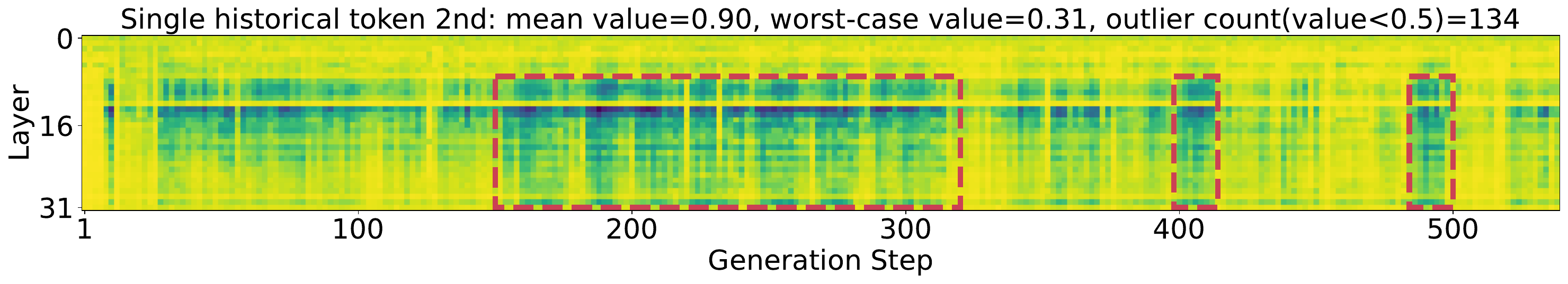}
				  
				\includegraphics[width=\textwidth]{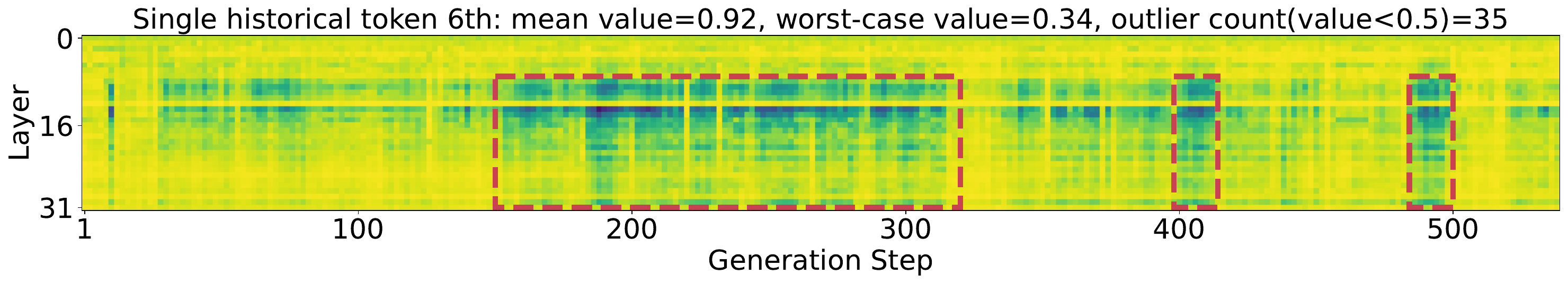} 
				  
				\includegraphics[width=\textwidth]{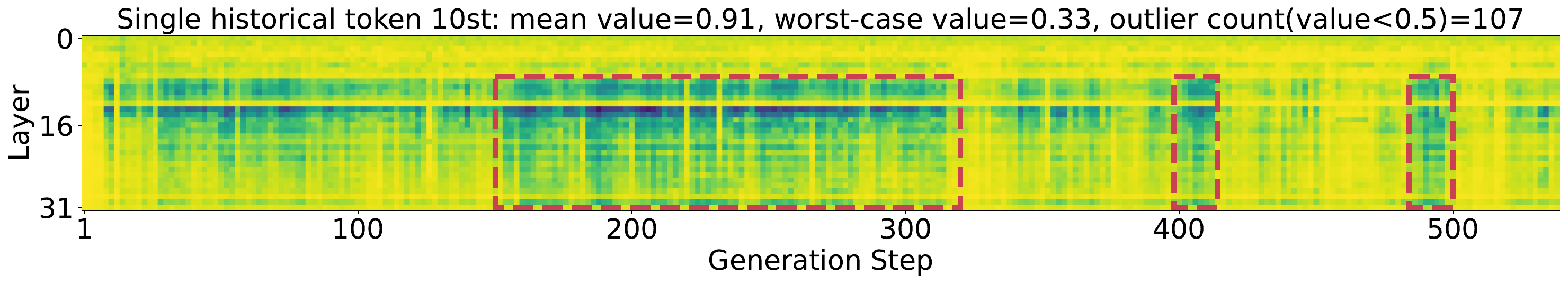} 
				  
				\includegraphics[width=\textwidth]{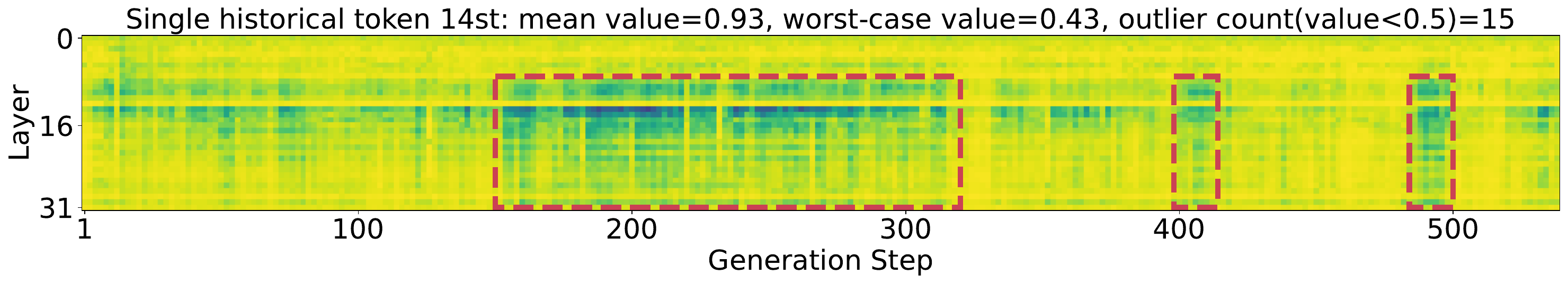} 
				  
				\includegraphics[width=\textwidth]{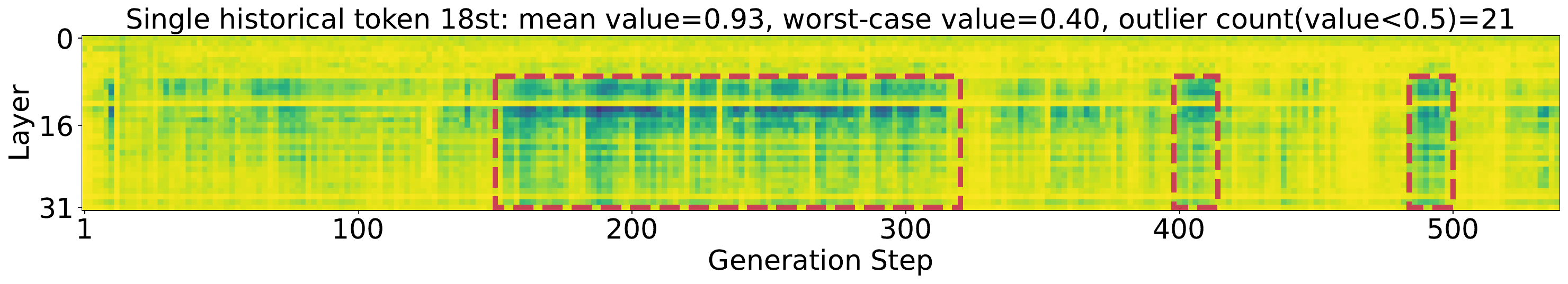} 
				  
				\includegraphics[width=\textwidth]{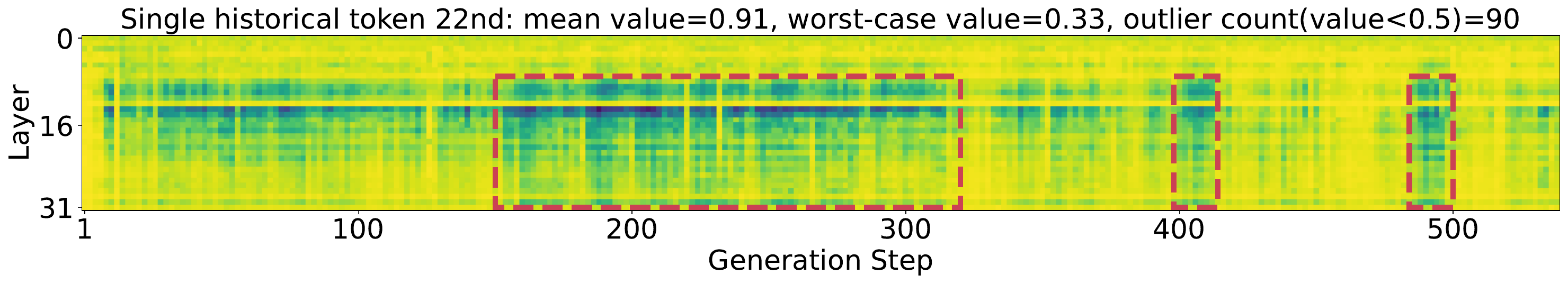}
				
				\includegraphics[width=\textwidth]{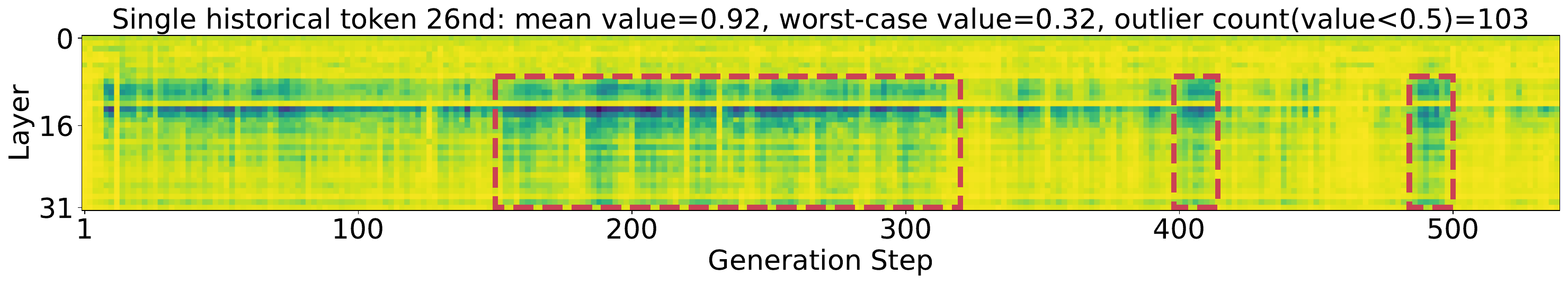} 
				\includegraphics[width=\textwidth]{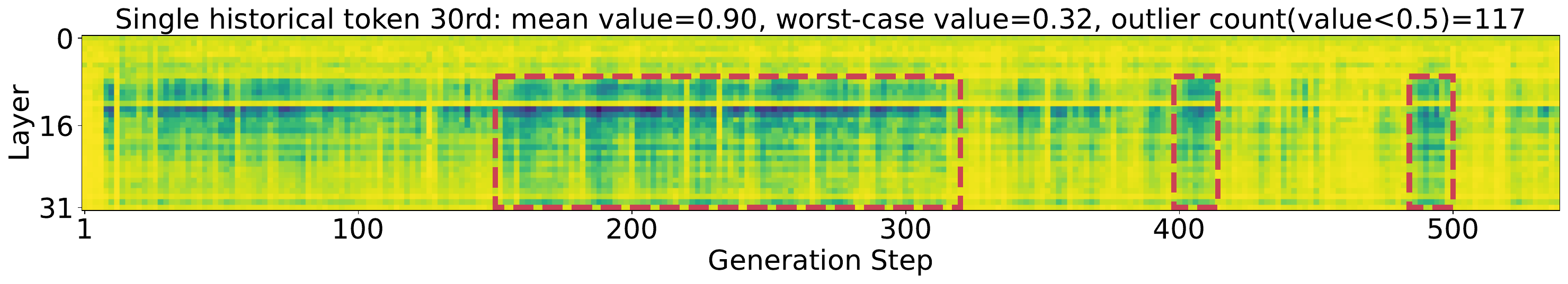}
		\end{minipage}%
		\begin{minipage}[c]{0.06\textwidth}  
			\includegraphics[width=\textwidth]{./figures/method_figure/legend.pdf}  
		\end{minipage}  
		\vspace{-0.2cm}
		\caption{ Breakdown of the stability assumption across different historical token measurements} 
		\label{fig:more_assumption}
		\vspace{-0.2cm}
\end{figure}

Complementing Figure \ref{fig:agg_assumption} in the main text, Figure~\ref{fig:more_assumption} provides a more detailed illustration. It demonstrates how measurements from single historical tokens, which guide cache eviction, experience significant degradation at certain generation steps.  The outlier cases occurs regardless of which specific historical token is used. Consequently, the failure of such averaging approaches is an expected outcome.

\section{Details of 16 Datasets in Longbench}
\label{apdx:details_datasets}
	
As a widely used long-context benchmark~\citep{ada,SnapKV,pyramidkv}, LongBench consists of 16 datasets across six task domains: single-document question answering (QA) \citep{kovcisky2018narrativeqa,dasigi2021dataset}, multi-document QA \citep{multi_hop1,ho-etal-2020-constructing,trivedi2022musique}, summarization \citep{huang2021efficient,zhong2021qmsum,fabbri2019multi}, few-shot learning \citep{joshi2017triviaqalargescaledistantly,gliwa2019samsum,li2002learning}, synthetic tasks \citep{bai2023longbench}, and code generation \citep{guo2023longcoderlongrangepretrainedlanguage,liu2023repobenchbenchmarkingrepositorylevelcode}. The average token length across all 16 datasets is 6,711. Table \ref{tab:detail_datasets} provides detailed information on the 16 datasets in LongBench.

\begin{table*}[thb!]
	\centering
	\small
	\caption{Details of 16 datasets in LongBench.}
	\label{tab:detail_datasets}
	\begin{tabular}{@{}lllllr@{}}
		\toprule
		Task                & Task Type     & Eval metric & Avg len & Language & Sample Num        \\ \midrule
		NarrativeQA         & Single-Doc. QA & F1          & 18,409   & EN       & 200            \\
		Qasper              & Single-Doc. QA & F1          & 3,619    & EN       & 200            \\
		MultiFieldQA-en     & Single-Doc. QA & F1          & 4,559    & EN       & 150            \\
		HotpotQA            & Multi-Doc. QA  & F1          & 9,151    & EN       & 200            \\
		2WikiMultihopQA     & Multi-Doc. QA  & F1          & 4,887    & EN       & 200            \\
		MuSiQue             & Multi-Doc. QA  & F1          & 11,214   & EN       & 200            \\
		GovReport           & Summarization & Rouge-L     & 8,734    & EN       & 200            \\
		QMSum               & Summarization & Rouge-L     & 10,614   & EN       & 200            \\
		MultiNews           & Summarization & Rouge-L     & 2,113    & EN       & 200            \\
		TREC                & Few-shot Learning & Accuracy    & 5,177    & EN       & 200            \\
		TriviaQA            & Few-shot Learning & F1          & 8,209    & EN       & 200            \\
		SAMSum              & Few-shot Learning & Rouge-L     & 6,258    & EN       & 200            \\
		PassageCount        & Synthetic     & Accuracy    & 11,141   & EN       & 200            \\
		PassageRetrieval-en & Synthetic     & Accuracy    & 9,289    & EN       & 200            \\
		LCC                 & Code          & Edit Sim    & 1,235      & Python/C\#/Java & 500\\
		RepoBench-P         & Code          & Edit Sim    & 4,206      & Python/Java & 500    \\ \bottomrule
	\end{tabular}
\end{table*}

\begin{table*}[h]
	\small
	\centering
	\caption{Single retrieval and multi retrieval templates in Needle-in-A-Haystack tests.}
	\label{tab:ruler_task_template1}
	\resizebox{\linewidth}{!}{
		\begin{tabular}{cp{0.9\linewidth}}
			\toprule

			\begin{tabular}{@{}c@{}}Single retrieval\end{tabular} & 
			\begin{tabular}{@{}p{\linewidth}@{}} 
				\textbf{Task Template:} \\
				Some special magic numbers are hidden within the following text. Make sure to memorize it. I will quiz you about the numbers afterwards.\\
				\textcolor{lightgray}{Paul Graham Essays.} \\
				\textcolor{lightgray}{......} One of the special magic numbers for \textcolor{violet}{\{word\}} is: \textcolor{orange}{\{number\}}. \textcolor{lightgray}{......}\\
				\textcolor{green!50}{What is the special magic number for \{word\} mentioned in the provided text?} \\ \\
				\textcolor{green!50}{The special magic number for \{word\} mentioned in the provided text is}
			\end{tabular}\\
			
			\midrule
			
			\begin{tabular}{@{}c@{}}Multi retrieval\end{tabular} &
			\begin{tabular}{@{}p{\linewidth}@{}} 
				\textbf{Task Template:} \\
				Some special magic numbers are hidden within the following text. Make sure to memorize it. I will quiz you about the numbers afterwards.\\
				\textcolor{lightgray}{Paul Graham Essays.} \\
				\textcolor{lightgray}{......} One of the special magic numbers for \textcolor{violet}{\{word\}} is: \textcolor{orange}{\{number-1\}}. \textcolor{lightgray}{......}\\
				\textcolor{lightgray}{......} One of the special magic numbers for \textcolor{violet}{\{word\}} is: \textcolor{orange}{\{number-2\}}. \textcolor{lightgray}{......}\\
				\textcolor{lightgray}{......} One of the special magic numbers for \textcolor{violet}{\{word\}} is: \textcolor{orange}{\{number-3\}}. \textcolor{lightgray}{......}\\
				\textcolor{lightgray}{......} One of the special magic numbers for \textcolor{violet}{\{word\}} is: \textcolor{orange}{\{number-4\}}. \textcolor{lightgray}{......}\\
				\textcolor{green!50}{What are all the special magic numbers for \{word\} mentioned in the provided text?} \\ \\
				\textcolor{green!50}{The special magic numbers for \{word\} mentioned in the provided text are}
			\end{tabular}\\

			\bottomrule
	\end{tabular}}
	
\end{table*}

\section{Limitations}
\label{apdx:limitations}
In this paper, we reveal for the first time the fragility of KV cache eviction and propose a defensive aggregation strategy for robust optimization. However, our work serves as a starting point and does not provide an in-depth investigation of broader robust optimization techniques. Future research can explore these techniques to further improve cache eviction performance.

\FloatBarrier

\end{document}